%% file: main.tex
\documentclass[11pt]{report}

\usepackage[utf8]{inputenc}
\usepackage{textcomp} 
\usepackage[american]{babel}
\usepackage{xspace}
\usepackage{tikz}
\usetikzlibrary{positioning}
\usepackage[fleqn]{amsmath} % math environments and more by the AMS 
\newtheorem{definition}{Definition}
\newcounter{dummy} % necessary for correct hyperlinks (to index, bib, etc.)
\usepackage{tabularx} % better tables
\usepackage{booktabs}
\usepackage{caption}
\usepackage[square,numbers]{natbib}
\usepackage{tikz}
\usepackage{makecell}
\usepackage{url}
\usepackage{multirow}
\usepackage{subcaption}
\usepackage[stable,bottom]{footmisc}
\usepackage{framed}
\usepackage{color}
\usepackage{float}
\usepackage{graphicx}
\usepackage{parskip}
\usepackage{algorithm}
\usepackage{geometry}
\usepackage{algcompatible}
\usepackage[noend]{algpseudocode}

\newcommand{\etal}{\textit{et al}.}
\makeatletter
\newcommand*{\rom}[1]{\expandafter\@slowromancap\romannumeral #1@}
\makeatother

\title{Master Thesis: Federated Transfer Learning with Multimodal Data}
\author{Yulian Sun }

\begin{document}
\maketitle
%\selectlanguage{american} % american ngerman

%\begin{abstract}
%\end{abstract}
\input{content}

\refstepcounter{dummy}
\addcontentsline{toc}{chapter}{\bibname}
\bibliographystyle{alpha} % <--- layout of the bib
\bibliography{bibliography} % file name of your bib

\end{document}

%% file: content.tex
% !TeX spellcheck = en_US
\begin{abstract}
%\hint{This document is given as a guideline for students to the template and a general structure of a thesis.}
% What you have done so far in short, and what excellent result have you got.

Smart cars, smartphones, wearable devices and other devices in the Internet of Things (IoT), which usually have more than one sensors, produce multimodal data. Federated Learning supports collecting a wealth of multimodal data from different devices without sharing raw data. Transfer Learning methods help transfer knowledge from some devices to others. Federated Transfer Learning methods benefit both Federated Learning and Transfer Learning.

This newly proposed Federated Transfer Learning framework aims at connecting data islands and offering privacy guarantee. Our construction is based on Federated Learning and Transfer Learning. Compared with some previous Federated Transfer Learning, where each user should have data with identical modalities (either all unimodal or all multimodal), our new framework is more generic, because it allows a hybrid distribution of user data. 

The core strategy is to use two different but inherently connected training methods for our two types of users. Supervised Learning is adopted for users with only unimodal data (Type 1), while Self-Supervised Learning is applied to user with multimodal data (Type 2) for both the feature of each modality and the connection between them. This connection knowledge of Type 2 will help Type 1 in later stages of training.

Training in the new framework can be divided in three steps.
In the first step, users who have data with the identical modalities are grouped together. For example, user with only sound signals are in group one, and those with only images are in group two, and users with multimodal data are in group three, and so on. In the second step, Federated Learning is executed within the groups, where Supervised Learning and Self-Supervised Learning are used depending on the group's nature. Most of the Transfer Learning happens in the third step, where the related parts in the network obtained from the previous steps are aggregated (federated).

%The selected dataset is a 
To demonstrate the effectiveness and robustness of our framework, we choose a
multimodal dataset with image (visual) and audio (auditory) modalities, and the goal of learning is scene classification. The experimental results demonstrate that the framework has better performance than the baselines. In addition, the framework shows high accuracy in the setting of Non-independent and identically distributed (non-IID) data. Compared with the baseline models, our framework has better performance. The accuracy of our framework can be up to 94.41\% for image modality and 92.82\% for audio modality, while the baseline models in Federated Learning without transferring can only achieve 93.68\% for image and 88.16\% for audio.

\end{abstract}

%*****************************************
\chapter{Introduction}
%*****************************************
The introduction is to motivate the use of Federated Transfer Learning \cite{yang2019federated} with multimodal data, state the problems and list the main contributions.
\section{Motivation}
In the real world, information often comes in different modalities. For example, we can describe one object with different modalities of data, e.g. images, videos, text, and voice. These multimodal data are obtained from different sensors and characterized by different statistical properties. Transfer Learning \cite{bozinovski2020reminder} techniques can help represent information jointly, making the machine learning model capture the relevant knowledge between different modalities. However, the scope of sharing between different devices is limited.

Moreover, it is not easy to obtain multimodal data from different sensors, and some sensors only provide unimodal data. On the other hand, the rapid growth in the processing power of mobile devices inspires more and more data heavy applications, which usually face privacy risks. The strictest privacy and security law is the General Data Protection Regulation (GDPR) \cite{EUdataregulations2018}, that is enforced by the European Union on May 25, 2018. GDPR intends to protect users’ privacy and data security. As a result, there is an increasing need to store and process data locally. Federated Learning \cite{konevcny2016federated1} can be used to aggregate the data from different participants with data privacy protection.

Thus, we set up a new framework with unimodal and multimodal data. This framework combines Federated Learning and Transfer Learning methods, which enables participants with multimodal data to help participants with unimodal data.

%What is the motivation for doing research in this area?
\section{Problem Statement and Contribution}
%What is the problem that should be solvs have different amounts of data with different modalities. 
The upper part of the following figure (Figure \ref{fig:figure_1_1}) shows the problem statement. All the participants cannot communicate with each other directly due to privacy protection. Each participant holds only a small amount of data. Thus, the problem is how the participants with multimodal data can transfer knowledge to others with only unimodal data. The lower part of the figure shows the core idea of solution, which will be elaborated in design part of this thesis.

\begin{figure}[htb]
	\centering
	\includegraphics[width=1.0\textwidth]{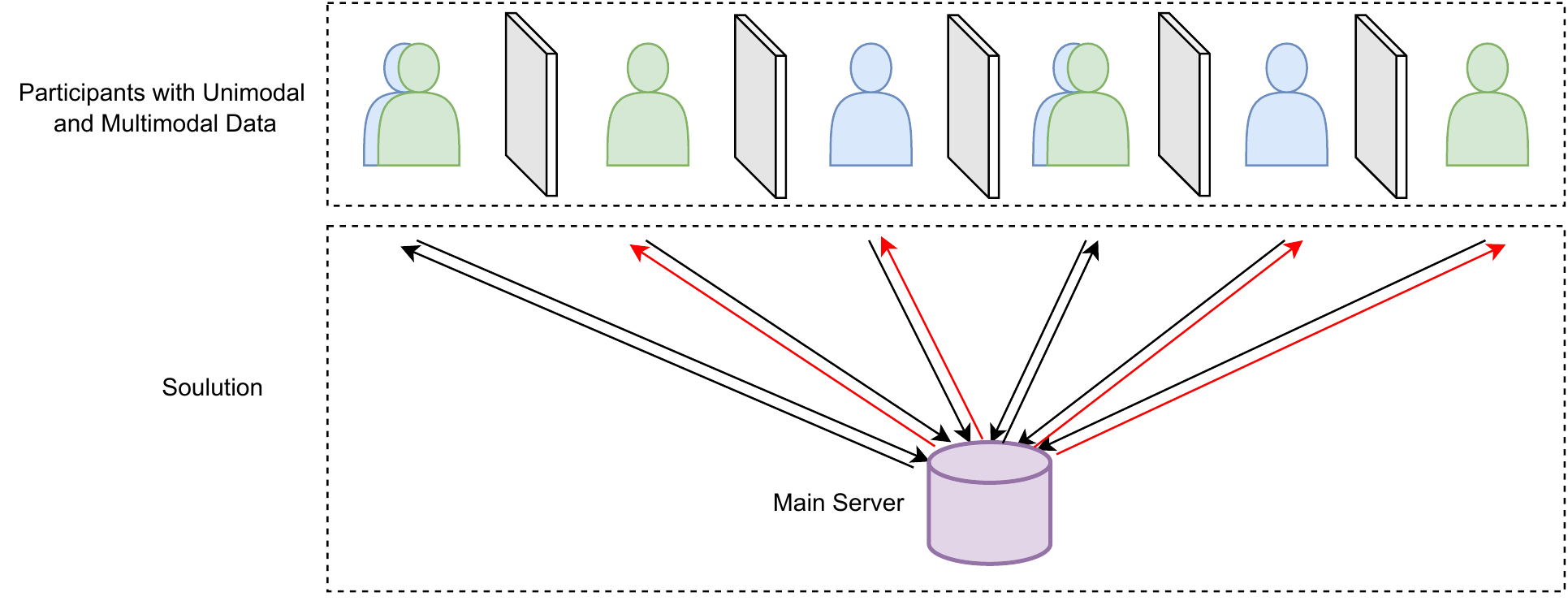}  	  	 	
	\caption{Problem statement for Federated Transfer Learning with multimodal data. The participants in blue or green have data with only one modality, while the participants in both blue and green have data with multi-modalities. The arrows denote sending and receiving models from participants or main server, and the red arrows represent knowledge transferring to participants with unimodal data.}
	\label{fig:figure_1_1}
\end{figure}

The contributions are as follows. 
\begin{enumerate}
	\item A new Federated Transfer Learning framework is presented, whose inputs are from sources with unimodal or multimodal data.
	\item Our core Transfer Learning technique analyzes the alignment in the exact modalities and uses self-supervision in pairs of data with different modalities but corresponding to (almost) the same object.
	\item Experiments over the scene classification dataset (i.e. audio-visual dataset) \cite{bird2020look} show that our method achieves effectiveness and robustness in the sense of transferring knowledge from multi-modality to uni-modality. 
\end{enumerate}

\section{Outline}
%How is the rest of this thesis structured?
%We organize the thesis as follows. 
First, we introduce background of sensors and modalities, different fusion approach with multimodal data, Supervised Learning \cite{cunningham2008supervised} and Self-Supervised Learning \cite{doersch2017multi}, Federated Learning \cite{konevcny2016federated1}, and Transfer Learning \cite{bozinovski2020reminder}.
Next, we analyze related work of Federated Transfer Learning, Transfer Learning with multimodal data, and Self-Supervised Learning with multimodal data.
Then, in design chapter, we design a new Federated Transfer Learning framework for multimodal data.
After that, we implement the new designed Federated Transfer Learning framework, which use Contrastive Learning \cite{chen2020simple} as a Transfer Learning strategy to learn transferable features. Besides, we implement the centralized approach with late fusion for unimodal and multimodal data as baselines, the effectiveness of which is viewed as a comparable group.
Finally, we evaluate the new designed Federated Transfer Learning in multimodal scene classification task \cite{bird2020look}. Besides, we compare the results of new designed Federated Transfer Learning framework with baseline models.

%*****************************************
\chapter{Background}
\label{ch:background}
%*****************************************
% Explains all concepts the reader needs to understand the present paper. This typically includes references to existing work that introduced the concepts, but usually a limited number thereof (<= 5, unless your paper builds on exceptionally diverse foundations).
%\hint{This chapter should give a comprehensive overview on the background necessary to understand the thesis.}
%Bib\TeX-Test: \cite{Steinmetz2005} \citeauthor{Steinmetz2005} \citep{Steinmetz2005}
This chapter is to introduce the background about our main task, the selected dataset, and the related methods. We first introduce the sensors and modalities of the selected dataset. Then, we analyze different multimodal data fusion methods. Next, we introduce Supervised Learning and Self-Supervised Learning. Finally, we present Federated Learning and Transfer Learning.
%The most related methods are multimodal data fusion, self-supervision, Federated Learning and Transfer Learning.

%*****************************************
\section{Sensors and Modalities}
In this thesis, we use scene classification to validate the feasibility of our proposal. Intuitively, scene classification is a task to answer the question: where am I? A typical example is to classify scenes in a video. The image-audio dataset \cite{bird2020look}, which is composed of 2-dimensional images (digital images) and 1-dimensional audio (acoustic signals), is a dataset with image-audio of multi-modality. The definitions of modality and multi-modality will be explained. The modalities of a video, i.e. image and audio, in scene classification will be elaborated, too. %To judge their environment, whether it happens to be indoors or outdoors, one step further, for example, in a restaurant or a beach. Humans can fully use their senses of sight, hearing, touch, etc. Further, the distinguishment depends on what they have experienced. Each video is composed of two modalities, i.e. 2-dimensional digital images and 1-dimensional acoustic signals.\\

\textbf{Multi-modality.} We humans can interact with the environment through touching, listening, seeing, etc. Similar to us humans, a machine interacts with an environment through different sensors, and each sensor can extract data like a human's sense. The information achieved from one type of sensor is defined one modality \cite{baltruvsaitis2018multimodal}. 
%The information achieved from more than one type of sensors, and the sensors are installed on the same machine.
Note that one machine can have several types of sensors. 
Then, the information collected by multiple sensors on the same machine, is defined multi-modality \cite{baltruvsaitis2018multimodal}. For example, a video is collected from only one single machine, which has two types of sensors, visual and acoustics sensor. Visual sensors produce visual modal data, image. Acoustics sensors produce acoustic data, audio.

\textbf{2D Digital Image.} \textit{Image} is a media that illustrates visual perception, such as a 2-dimensional digital picture that resembles a subject (usually a physical object) and provides a refined description \cite{latham2009federal}. A digital image is formed by discrete picture elements called pixels. So each digital image is 2-dimensional array of pixels. Pixels, the minor units of the digital images, contain fixed values that describe any particular point's color.

\textbf{Visual Sensor.} The two primary visual sensors are the charge-coupled device (CCD) \cite{howes1979charge} and complementary metal-oxide semiconductor (CMOS) \cite{fossum1997cmos}. Both of them are used in cameras. 
%but we widely use CMOS.\cite{fossum1997cmos}
However, CMOS has better performance than CCD, and offers advantages in lower system power, higher noise immunity, lower cost, and a smaller system size \cite{fossum1997cmos}. CMOS is widely used in smart phones and digital cameras \cite{el2005cmos}.
We mainly model the CMOS sensor as follows. The CMOS sensor converts photons into voltages at the pixel site, which causes less sensitivity. The transistors near the pixels are used to measure and amplify the signal from the pixels. In these two processes, CMOS sensors have higher speed to produce pixels. In addition, a CMOS sensor consists of two-dimensional color filter arrays of blue (B), red (R),and green (G) pixels, i.e., the Bayer filter pattern, in which the active area of the G color filters is two times larger than that of the B and R color filters \cite{ohta2017smart}.
Perceiving the colors of the environment affects the reliability and accuracy, which in turn affects the related scene classification tasks. Particularly, RGB color-based cameras can lead to a significant performance improvement.
% actually, the popular color format is RGB.

\textbf{1D Acoustic Signal and Sound Sensor.} Audio refers to sound as it can be percepted by sound sensors. Sound sensors work just like human ears, and they also have a diaphragm that detects sound waves by their intensity and converts vibrations into electrical signals. Each sound sensor contains an integrated condenser microphone, a peak detector, and an amplifier that is exceptionally attentive to sound \cite{soundsensor}. The specific measurement is usually carried out by calculating the amplitude of the sound in a fixed time interval. These converted electrical signals are then extracted by Mel-Frequency Cepstral Coefficients(MFCC) \cite{muda2010voice}.
\section{Multimodal Data Fusion}
% What is Multimodal Data Fusion?
Parcalabescu \etal\ \cite{parcalabescu2021multimodality} presented a survey about multi-modality in our environment. This survey highlights that a machine processes the input and acts as a (multimodal) agent that decides how to perceive the input. In principle, a machine uses multiple sensors in a combination way to perceive the environment. This method is formalized as Data Fusion.
%Information may undergo constant changes in representation, does not immediately leapfrog a new modality \cite{parcalabescu2021multimodality}.
%Multimodal data fusion is in order to improve performance neither of them.
%The process of integration of data and knowledge from several sources of modalities is known as data fusion \cite{klein2004sensor}. Following is well-known definition of data fusion for multi-modality.
%But Multimodal Data Fusion has problem which 
% mathematic description or definition
\begin{definition}[Data Fusion \cite{hall1997introduction}]\label{def:lahat2015multimodal}
	"Data fusion techniques combine data from multiple sensors and related information from associated databases to achieve improved accuracy and more specific inferences than could be achieved by the use of a single sensor alone."
\end{definition}
Different modalities represent variants of data. The diversity, which the particular natural processes and phenomena can describe themselves under totally various physical characters, is the motivation for multimodal data fusion \cite{castanedo2013review}. However, very little is understood about the potential association among the modalities. The main task of any multimodal analysis is to identify the connections among the modalities, their mutual properties, their complementarity, and shared modality-specific information \cite{castanedo2013review}. 

Data fusion relates to many fields. Although it is difficult to set up an explicit, generic and rigorous classification of the techniques, the authors of \cite{castanedo2013review} \cite{zhang2020multimodal} suggested that we use for classification the following four criteria.

\begin{table}[htb]
	\centering
	\begin{tabular}{|l|l|l|}
		\hline  
		\textbf{Criterion} & \textbf{Details} \\
		\hline 
		\rom{1} & (1) complementary, (2) redundant, (3) cooperative data\\
		\hline
		\rom{2} & (1) raw measurement/signal, (2) pixel, (3) characteristic or decision\\
		\hline
		\rom{3} & (1) early, (2) late\\
		\hline 
		\rom{4} & (1) centralized, (2) decentralized, (3) distributed, (4) hierarchical\\
		\hline 
	\end{tabular}
	\label{tab:table_2_1}
	\caption{Criteria for classification of data fusion techniques \cite{castanedo2013review} \cite{zhang2020multimodal}.}
\end{table}

\textbf{Criterion \rom{1}.} The first criterion was proposed by Durrant-Whyte \cite{durrant1990sensor}, where the links within source datasets should be considered. The links can be defined as (1) \textit{complementary}, (2) \textit{redundant}, (3) \textit{cooperative data}. The links within the source datasets are complementary, if the input source data represent various scenes and can be used to acquire global information, e.g. the information obtained by two cameras observing the same object from different fields of view is considered complementary. The relations in the source datasets are redundant, if large equal than two input source data supply details about the identical object and can be merged to increase the confidence, e.g. data from overlapping regions are conclude as redundant. The relationship between the source datasets is cooperative, if the supplied features are merged as a new feature which is often more complicated than the original features \cite{castanedo2013review}. 

%Other possible abstract fusion level can be described as (1) early fusion, (2) medium level fusion, (3) high level fusion and (4) multiple level fusion.
\textbf{Criterion \rom{2}.} The second criterion is to consider the abstraction level of source data:(1) \textit{raw measurement/signal}, (2) \textit{pixel}, (3) \textit{characteristic or decision}. When the signals obtained from the sensor can be processed directly, the abstraction level is raw measurement/signal. When the fusion happen at image and can be utilized to increase image clarity performance, the abstraction level is pixel level. When the fusion uses features extracted from images or signals, the abstract fusion level is \textit{characteristic}. At \textit{characteristic} level, fused information is represented as symbols, which is also called decision level \cite{castanedo2013review}. 

\textbf{Criterion \rom{3}.} The third criterion is consider when to perform fusion during the associated procedures: (1) \textit{early fusion}, (2) \textit{late fusion} \cite{zhang2020multimodal}. Early fusion performs fusion at early training stage, while late fusion performs fusion at almost the end of training stage.

% 重新整理
\textbf{Criterion \rom{4}.} The fourth criterion is considered as different architecture variants:(1) \textit{centralized}, (2) \textit{decentralized}, (3) \textit{distributed}, (4) \textit{hierarchical}. In the \textit{centralized} architecture, the fusion nodes locate in the central processor where the information from all of the inputs are received, measured and transmitted. A \textit{centralized} approach is theoretically optimal, if it is assumed that data alignment and data association are performed correctly, and that the required transfer time is not significant. However, there is drawback that a large of bandwidth is required to send raw data over the architecture. This drawback can have a greater impact on the results than other architectures \cite{castanedo2013review}. 
A \textit{decentralized} architecture consists of a network of nodes, where each node has its own processing power, there exists no individual point of data fusion. Thus, the information that each node receives from its peers is fused autonomously with its local information. However, this fusion schema has a disadvantage, which is $O(n^2)$ for cost at each communication step ($n$ is the number of nodes). Furthermore, this schema may suffer from scale expansion issues when the number of nodes increases \cite{castanedo2013review}. 
In a \textit{distributed} architecture, each source node performs its data association and state estimation individually before the raw information conveys to the fusion node. This means that each source node contributes an estimation of the object state from its local perspective, and then the fusion node fuses estimations based on the global perspective. Therefore, this schema provides various range of options, from just one fusion node to many intermediate fusion nodes. In a \textit{hierarchical} architecture, \textit{decentralized} and \textit{distributed} architectures are combined to generate a hierarchical schema, where the data fusion can be achieved at different levels \cite{castanedo2013review}.

% models 
\textbf{Classification of Data Fusion of Our Framework.} Our framework is \textit{complementary}, because our framework applies for the multimodal data with multi modalities (views). Our framework is \textit{characteristic} level, because the features are fused after they are extracted from images and audio. This framework is also a \textit{late fusion} approach, because the fusion operation occurs at the end of training process. Besides, this approach can be seen as a centralized model and is used in Federated Learning, because it can be used as. 
%In addition, our framework is \textit{model-driven}, because it fuse different models (sub-networks). 
% multimodal Transfer Learning approach introduction hateful memes detection through Transfer Learning using multimodal imaging data.\cite{}

\section{Supervised Learning and Self-Supervised Learning}
\textbf{Supervised Learning.} Supervised Learning is defined by its use of labeled data \cite{cunningham2008supervised}. One typical task of Supervised Learning is classification.
In Supervised Learning, models are trained using labeled dataset, where the model learns about each type of data. Once the training process is completed, the model is tested on the test data (a subset of the dataset), and then it predicts the output.\\
There is a dataset with inputs $X$ and labels/classes $Y$, a model of Supervised Learning $M$. $M$ is considered as a mapping from one given sample $x\in X$ to its label $y\in Y$. The definition is following:
\begin{align}
	M: x\mapsto y,
\end{align}
The output of $M$ is $\widetilde{y}$, if one given sample is $x$, namely $\widetilde{y}=M(x)$. where $\widetilde{y}$ is a vector that represents the probability distribution of sample $x$ in a certain label. $\widetilde{y}$ is considered as prediction posteriors. When we train a supervised model, a loss function is needed $L(M(x), \widetilde{y})$, which is a metric to measure the distance between the prediction posterior of a sample and its label. In training process, the loss function is to minimize over a training dataset $D_{tr}$. The formula is as follow,
\begin{align}
%\sum\limits_{i=0}^n {x_i}
\operatorname*{arg min}_M \frac{1}{|D_{tr}|} \sum \limits_{(x,y)\in D_{tr}} L(M(x), \widetilde{y}).
\end{align}
% how to train Supervised Learning
Cross entropy loss is one of the most used loss functions for classification tasks. The definition of cross entropy is as the following.
\begin{align}
L_{CE} = - \sum\limits_{i=1}^l {y^i} \text{\xspace log \xspace} \widetilde{y}^i,
\end{align}

where $l$ is the total number of labels. If the sample is predicted as label $i$, $y^i$ equals to 1 (otherwise 0). Meanwhile, $\widetilde{y}^i$ is the $i-$th value in the prediction posteriors $\widetilde{y}$. In our framework, we use cross entropy loss \cite{bishop2006pattern} as the loss function in all the supervised training.

\textbf{Self-Supervised Learning.} Self-Supervised Learning is Unsupervised Learning, which learns the representation of unlabeled data by just observations of how different parts of the data interact with one another \cite{doersch2017multi}. Self-Supervised Learning reduces the requirements for a large amount of data of labeled data. In addition, the method can be used to explore the association in multi-modalities within a single data sample \cite{chen2020simple}.

\begin{figure}[H]
	\centering
	\includegraphics[width=0.8\textwidth]{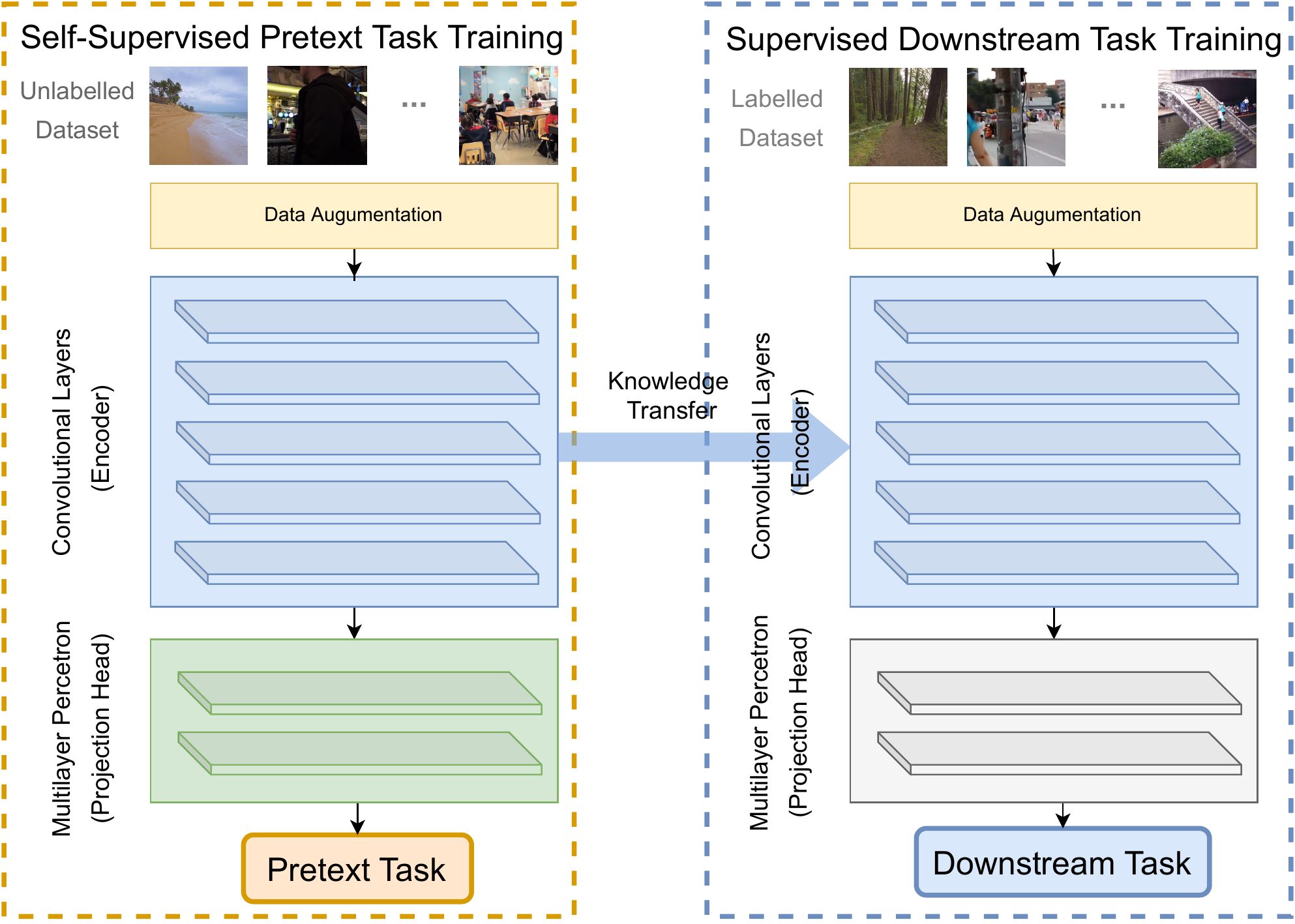}
	\caption{The pipeline of Self-Supervised Learning \cite{jaiswal2020survey}. Self-Supervised Pretext Task Training (left): unlabeled data is processed by data augmentation. Then, the augmented data applies to a neural network, which consists of an encoder and a projection head. Supervised Downstream Task Training (right): labeled data is processed by data augmentation. Then, the augmented data applies to the other neural network, which also consists of an encoder and a projection head. The knowledge transferring exists two encoders.}
	\label{fig:figure_2_1}
\end{figure}

Self-Supervision tasks include two phases: \textit{pretext task} and \textit{downstream task}. Following is the general pipeline of Self-Supervised Learning (see Figure \ref{fig:figure_2_1}). The feature of visual data is extracted from convolutional layers of networks to support a pre-trained pretext task. Then, the parameters from the pre-trained model are transferred to the downstream visual tasks (e.g. image classification) using fine-tuning. The downstream tasks are used as an evaluation tool to assess the performance of the pretext tasks. 
%The downstream tasks is 
%In the Transfer Learning, the features are transferred only from the previous convolutional layers of the pretext task to the downstream task \cite{jaiswal2020survey}.

\textbf{\textit{Pretext Task.}} Pretext tasks aim to learn the representations of input data. Specifically, pretext tasks use convolutional layers based models to extract features from input data. The learned models are treated as pre-trained models. Moreover, these models are usually be used for the downstream tasks, for example, image classification, image segmentation, object detection, etc. \cite{jaiswal2020survey}. In addition, these tasks can be used for almost any type of data \cite{jaiswal2020survey}. The frequently used pretexts based on different applied scenarios can be divided into four variants: color transformation, geometric transformation, context-based tasks, and cross-modal based tasks \cite{jaiswal2020survey}.

\begin{figure}[htbp]
	\centering
	\begin{minipage}[H]{0.49\textwidth}
		\includegraphics[width=0.90\textwidth]{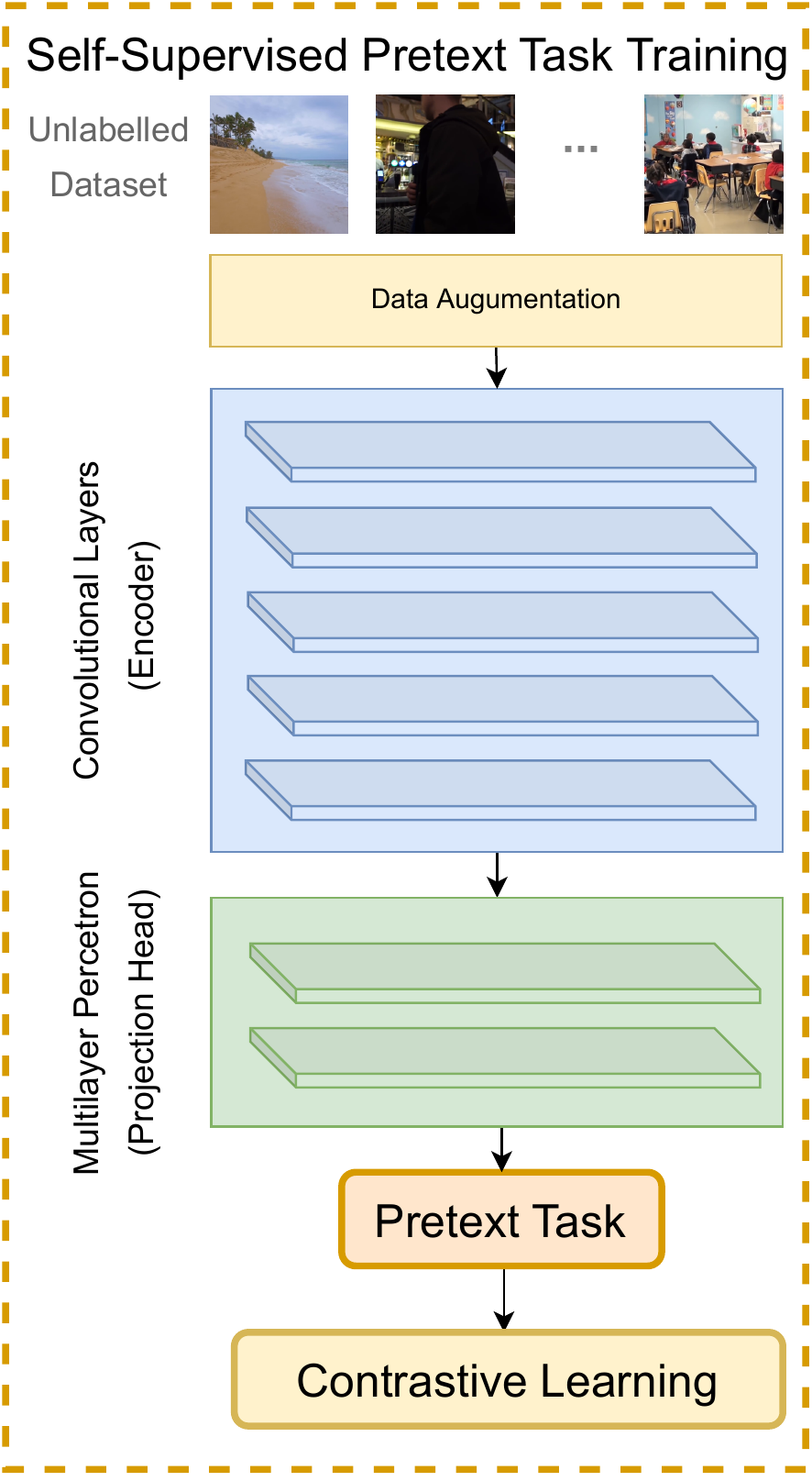}
		\caption{Contrastive Learning pipeline \cite{jaiswal2020survey}. Contrastive Learning is similar to Self-Supervised pretext task training. (see Figure \ref{fig:figure_2_1})}
		\label{fig:figure_2_2}
	\end{minipage}
	\begin{minipage}[H]{0.49\textwidth}
	\includegraphics[width=1.0\textwidth]{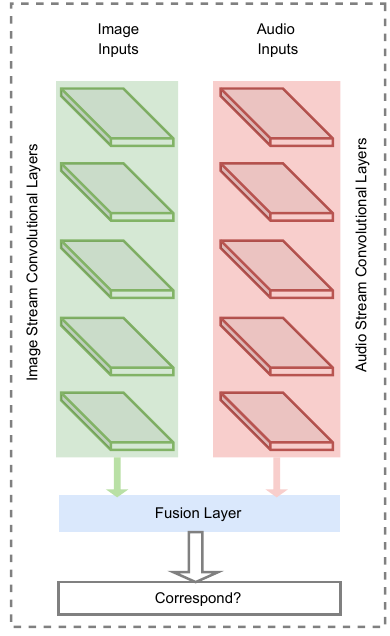}   
	\caption{The architecture of Video and Audio Correspondence Verification Task \cite{jing2020self}. A pair of image and audio inputs are fed into two sub-networks consisting of stacked convolutional layers.}
	\label{fig:figure_2_3}	
	\end{minipage}
\end{figure}

Most pretext tasks focus on learning the invariance of colors in images, when color converting techniques are used in color transformation \cite{jaiswal2020survey}. In geometric transformation, pretext tasks mainly focus on learning similarity under spatial transformations of the images, which include flipping, scaling, cropping, etc. \cite{jaiswal2020survey}. Context-based tasks include jigsaw puzzle, frame order based tasks and future prediction. Jigsaw puzzle tasks aim to rearrange the scrambled patches of an image with an encoder. Frame order based tasks deal with time series data, for example a video. Future prediction is to do the future prediction from a past sequential data. In cross-modal based tasks (view prediction), encoder models from different modalities simultaneously learn the similarity of feature representations \cite{jaiswal2020survey}. For example, video based data features of the corresponding data streams like RGB image frame and audio sequence (see Figures \ref{fig:figure_2_3}). Besides, the constraint of image and audio modalities provide auxiliary information about the contentment of videos. Overall, the critical purpose is to force the model to maintain the invariance to transformations and keep distinctions to other data points.

\textbf{\textit{Downstream Task.}} Downstream tasks are goal-oriented tasks include almost all tasks in computer vision field \cite{jaiswal2020survey}, e.g. image classification, image or video segmentation, object detection, etc. The parameters from the pre-trained models are transferred to any downstream tasks, in which fine-tuning method is used \cite{jaiswal2020survey}.

% 考虑什么是similar,什是dissimilar
\begin{figure}[htbp]
	\centering
	\includegraphics[width=1.0\textwidth]{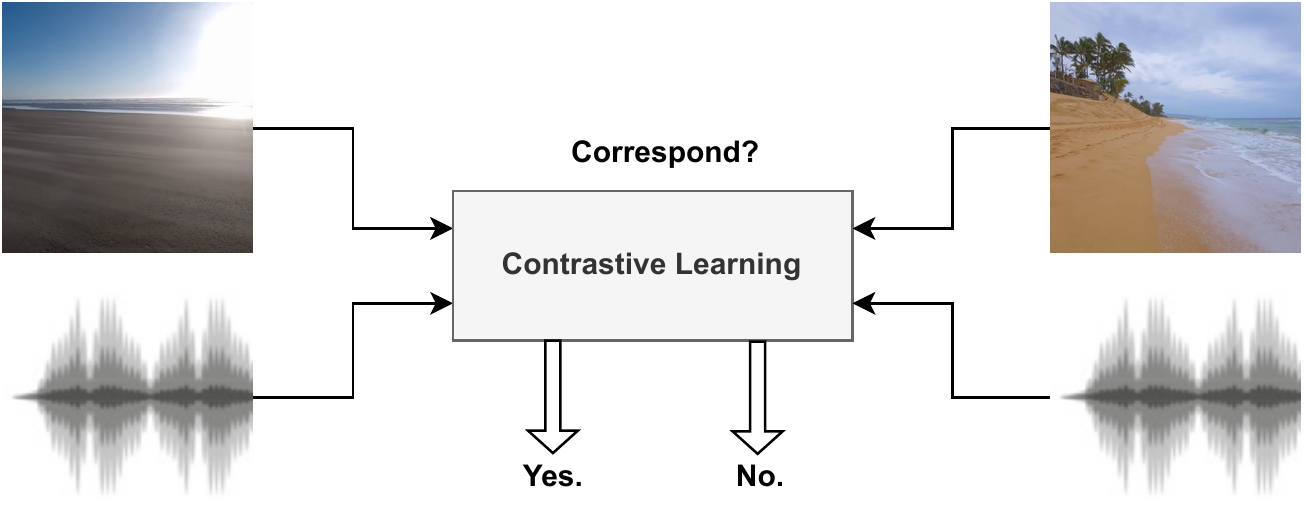}
	\caption{Contrastive Learning for videos \cite{jaiswal2020survey}. This is an example of Contrastive Learning used to determine whether an audio and an image are consistent. }
	\label{fig:figure_2_4}
\end{figure}

Liu \etal\ \cite{jaiswal2020survey} presented a survey about Self-Supervision. In this survey, Self-Supervised Learning can be divided into three categories: \textit{generative} , \textit{contrastive}, and \textit{generative-contrastive (adversarial)}. Generative means that a pair of encoder and decoder models are trained, and the encoder model encodes the input data into an vector and the decoder model reconstructs input data from vector (e.g. graph generation ). Moreover, contrastive describes that an encoder model encodes input data into a vector to calculate the similarity (e.g. instance estimation). In addition, Generative-Contrastive (Adversarial) indicates that an encoder-decoder model generates a discriminator as well as fake instances with view to judge the fake ones from real ones (e.g. GAN).

\textbf{\textit{Contrastive.}} Contrastive Learning (see Figure \ref{fig:figure_2_2}, \ref{fig:figure_2_4}) is the most frequently used in these three categories \cite{chen2020simple} \cite{arandjelovic2018objects} \cite{patrick2020multi} \cite{morgado2021audio}
 \cite{afouras2020self} \cite{khurana2020cstnet} \cite{tian2019contrastive}, with the most representative method SimCLR \cite{chen2020simple}. Besides, a contrastive loss, Noise Contrastive Estimation (NCE) \cite{gutmann2010noise}, which is similar to the loss function in Supervised Learning, is also commonly used in Contrastive Learning. Following is the formulas for NCE:
\begin{align}
\text{L}_{NCE} = -\text{log}\frac{\text{exp}(\frac{\text{sim}(q,p_+)}{\tau})}{\text{exp}(\frac{\text{sim}(q,p_+)}{\tau}) + \text{exp}(\frac{\text{sim}(q,p_-)}{\tau})}
\end{align}
%\begin{align}
%\text{L}_{NCE} = -\text{log}\frac{\text{exp}(\frac{\text{sim}(q,p_+)}{\tau})}{\text{exp}(\frac{\text{sim}(q,p_+)}{\tau}) + \sum^K_{i=0}\text{exp}(\frac{\text{sim}(q,p_i)}{\tau})}
%\end{align}
where $q,p_+, p_-$ represent the original sample, a positive sample and a negative sample, respectively. $\tau$ is a hyperparameter, which is called temperature coefficient. The function $sim()$ can be a cosine similarity \cite{bishop2006pattern}. \\

Contrastive Learning has following components (workflow see Figure \ref{fig:figure_2_2}).\\

\textbf{Data Preprocessing.} In SimCLR, Chen \etal\ \cite{chen2020simple} used data augmentation to transform one image sample $x$ into two augmented views, denoted by $\widetilde x_{left}$ and $\widetilde x_{right}$, which can be seen as a positive pair for $x$.
%In our framework, we consider a video sample is $x$, a RGB image frame and a audio sequence from this video are two different modalities: a RGB image frame is transformed by flipping, denoted as $x_i$, a audio sequence is extract by MFCC, denoted as $x_a$. \\

% need to describe encoders.
\textbf{Encoder $f$.} Encoder (stacked convolutional layers) is the main components in pretext tasks. Encoders provide useful feature representations, which makes classification model easier to distinguish different classes \cite{jaiswal2020survey}. Encoders can be various deep neural networks. The representation are $h_{left}=f(\widetilde x_{left})$ for $\widetilde x_{left}$ and $h_{right}=f(\widetilde x_{right})$ for $\widetilde x_{right}$
%In a video sample with multi-modality, we use VGG16 \cite{simonyan2014very} as a encoder for image inputs, and 3 stacked fully connected layers for audio inputs. We obtain the representations from image modality $h_i = f(x_i)$ and audio modality $$. \\

\textbf{Project Head $p$.} Projection head $p$ is a shallow neural network, which can be considered as a projection function with several fully connected layers, which projects the representations from encoded data to a hidden space. The aim of project head is to enhance the performance of encoders, and align the representations with an identical hidden space shape. The outputs of project head are $z_{left}=p(h_{left})$ for $h_{left}$ and $z_{right}=p(h_{right})$ for $h_{right}$.
% need to describe the training process.

\textbf{Contrastive Loss Function.} The contrastive loss function \cite{gutmann2010noise} is formally the same as cross entropy loss \cite{bishop2006pattern}. The contrastive loss function \cite{chen2020simple} is what makes the model to learn the feature representations by itself. There is dataset with augmented samples $\widetilde X=\{\widetilde x_1, \widetilde x_2, ..., \widetilde x_k\}$. Each sample $x$ has a positive pair $\widetilde x_{left}$ and $\widetilde x_{right}$. Contrastive loss is used to maximize the similarity between $\widetilde x_{left}$ and $\widetilde x_{right}$, and minimize the positive pair ($\widetilde x_{left}$, $\widetilde x_{right}$) and other samples. With a batch of $N$ samples, we have $2N$ augmented views. The contrastive loss function is as follows: 
\begin{align}
\text{L}_{(left,right)} = -\text{log}\frac{\text{exp}(\frac{\text{cos\_sim}(z_{left},z_{right})}{\tau})}{\sum \limits_{k=1,k\neq i}^{2N} \text{exp}(\frac{\text{cos\_sim}(z_{left},z_{right})}{\tau})}.
\end{align}
The similarity function is a cosine similarity \cite{bishop2006pattern}. The term $\tau$ is a hyperparameter, which is called \textit{temperature coefficient}. The function cos\_sim() is to calculate the similarity between $\widetilde x_{left}$ and $\widetilde x_{right}$. It is defined as:
\begin{align}
\text{cos\_sim}(\widetilde x_{left}, \widetilde x_{right}) = \frac{\widetilde x_{left} \cdot \widetilde x_{right}}{\lVert \widetilde x_{left}\lVert \cdot \lVert \widetilde x_{left}\lVert}.
\end{align}
The final loss $\text{L}_{CL}$ is calculated over all the positive pairs. In a batch with $B$ samples, $\text{L}_{CL}$ is computed as:
\begin{align}
\text{L}_{CL} =\frac{1}{2B}\sum \limits_{k=1}^{2B}[L(2k-1, 2k)+L(2k, 2k-1)],
\end{align}
where $k$ is the index of samples, $(2k-1, 2k)$ and $(2k, 2k-1)$ are the indices of each positive pair.

%that utilize the knowledge that was learned during the pretext task. They can be anything such as classification, detection, segmentation, future prediction, etc. in computer vision. Once example of downstream task can be hand gesture classification [55] that involves both object detection and classification. Figure 17 represents the overview of how knowledge is transferred to a downstream task. The learned parameters serve as a pretrained model and are transferred to other downstream computer vision tasks by fine-tuning. 

% evaluate the results
\textbf{Evaluation without Downstream Task.} The general evaluation of self-supervision focus in the second phase of downstream tasks. The self-learned pretext models serve as pre-trained models. Then, pre-trained models are fine-tuned by any visual downstream tasks include image classification, image segmentation, pose estimation, .etc. In other words, the evaluation of Self-Supervised Learning is to assess its performance on transferring perspective, which is in a high-level to prove the generalization power of the pre-trained models. Thus, the evaluation performs over the implementation of general Supervised Learning with fine-tuning. Our framework needs only the first phase of self-supervision to achieve the representations.

%总结目前的情况,已了解到的
%In conclusion, Self-Supervised Learning methods achieve great success in representing features, which reduces the requirements of labeled data. We apply contra
\textbf{Contrastive Learning in Our Framework.} We apply Contrastive Learning to our framework. The workflow of contrastive learning in our framework is similar as described above. However, there is small difference. The modalities of each sample are considered as \textit{views}. For example, if we use a video with RGB frame images and audio sequence, each video sample $x$ is preprocessed by data augmentation, denoted as a flipped RGB image $\widetilde x_{i}$ and a MFCC extracted feature $\widetilde x_{a}$.

\section{Federated Learning}
% the short history of Federated Learning
The term of Federated Learning was first proposed by Google in 2016 \cite{konevcny2016federated1}. Federated Learning is a collaborative learning paradigm that allows many participants (e.g. mobile devices) to learn a model in parallel without sharing raw data, while a central server controls the aggregation of various participant models \cite{mcmahan2016federated}. 
%definition of Federated Learning, I use mathematics to describe it

Federated Learning allows data diversity with communication efficiency \cite{konevcny2016federated2}. Difficulty like network unavailability in edge computing equipment may prevent industries from merging datasets from different sources, which makes each data source like an island. Even when the data source can only communicate at a specific time, Federated Learning helps to access heterogeneous data \cite{konevcny2016federated2}. Federated Learning supplies real time continual learning. Federated Learning uses participants' data to continuously improve the model, without the need to aggregate data for continuous learning \cite{zhang2021real}. Federated Learning offers hardware efficiency. Federated Learning uses less complex hardware, because Federated Learning has no need of central server to analyze data \cite{zhang2021real}. Last but not least, Federated Learning protect the data privacy without sharing raw data \cite{mcmahan2017federated}.

The following specific definition of Federated Learning was proposed by Yang \etal\ \cite{yang2019federated}.

They define a set of $N$ data holders $P = \left\{P_1, ..., P_N\right\}$, all of them intend to train a machine learning model by connecting their local data $D=\left\{D_1, ... D_N\right\}$. A common method is to collect all data together and use $D = D_1\cup ... \cup D_N$ to train a modal $M_{SUM}$. A Federated Learning framework is a learning procedure in which the data holders collaboratively train a model $M_{FED}$, during the training of which any data owner $P_i$ does not share its data $D_i$ to others. Additionally, the accuracy of $M_{FED}$, denoted as $A_{FED}$, should be very close to the performance of $M_{SUM}$, namely, $A_{SUM}$. Formally,
\begin{align}
	|A_{FED}-A_{SUM}|< \delta,
\end{align}
the term $\delta$ in Federated Learning algorithm is called $\delta$-accuracy loss, $\delta$ is a small positive real number \cite{yang2019federated}.

% variants of Federated Learning by which uility
According to the distribution of data feature space and sample space, researchers divide Federated Learning into three categories, i.e. horizontal Federated Learning, vertical Federated Learning and Federated Transfer Learning \cite{yang2019federated}.
%(see Figure \ref{fig:figure_2_5})
\textbf{Horizontal Federated Learning.} Horizontal Federated Learning or sample-based Federated Learning utilizes datasets with the identical feature space but different sample space on all devices, which means that participant A and participant B have the identical features.
%(see Figure \ref{fig:figure_2_6})
\textbf{Vertical Federated Learning.} Vertical Federated Learning or feature-based Federated Learning utilizes different datasets with different feature space but identical sample space in order to train a global model. For example, participant A has information about the customer's jewellery and clothing purchases, and participant B has information about the customer's comments and reviews of jewellery and clothing, utilizing these two datasets from two different fields, participant B is enable to serve the customers better using comments and reviews of clothing to provide better clothing recommendation to the customers searching clothes in participant A.
%(see Figure \ref{fig:figure_3_1})
\textbf{Federated Transfer Learning.} Federated Transfer Learning utilizes different datasets which are not only different in sample space but also in feature space. For example, Federated Transfer Learning aims to train a customized model, which does clothing recommendation based on the single customer's past searching behaviors.

% how it works, process of Federated Learning
There are one main server for model aggregation and multiple client devices, and the considered procedure of model updates is synchronized. The entire Federated Learning workflow can be described with four steps, an iterative in learning process is assumed as one global epoch \cite{bonawitz2019towards}. 
\begin{itemize}
\item \textbf{Step 1:} Participant holds a portion of the data. The main server chooses a machine learning model, and initializes the model. Then, the main server send the initialized model to each participant.
\item \textbf{Step 2:} In parallel, each participant receives the model from the main server, and trains locally. Then, each participant updates the current model based on its own data using an optimization algorithm and sends the resulting model to the server.
\item \textbf{Step 3:} The server receives the models from participating nodes. Then, the server updates the global model as the average aggregation of these received models, and sends the updated model to each participating nodes. 
\item \textbf{Step 4:} The iteration is repeated for many global epochs until the end of the maximum global epoch or when the expected performance (e.g. accuracy) is achieved.
\end{itemize}
% benifits,why federated fearning ?

% problems or challenges
\textbf{Heterogeneous Data.} Federated Learning still has challenges with heterogeneous data or unbalanced distributions of data, i.e. (None Independent and Identically Distributed) non-IID. In theoretical setting in federated learning, nodes of local data samples are often Independent and Identically Distributed (IID). However, in the real world, the data samples can not be IID. In the paper \cite{sattler2019robust}, robust and communication-efficient Federated Learning from non-IID data was proposed.

%need a picture for federated fearning process
%\begin{figure}[htb]
%	\centering
% 	\includegraphics[width=0.8\textwidth]{images/HFL.pdf}  	  	 	
%	\caption{Horizontal Federated Learning \cite{yang2019federated}.}
%	\label{fig:figure_2_5}
%%	\label{fig:example}
%\end{figure}
% need a picture for three variants of federated fearning
%\begin{figure}[htb]
%	\centering
%	\includegraphics[width=0.8\textwidth]{images/VFL.pdf}  	  	 	
%	\caption{Vertical Federated Learning \cite{yang2019federated}.}
%	\label{fig:figure_2_6}
%%	\label{fig:example1}
%\end{figure}

{Federated Learning in Our framework.} Our framework is for Federated Transfer Learning. However, our framework works as normal horizontal Federated Learning, when the participant has only a single modality. The participants with only one single modality perform Federated Learning in a group. The workflow is similar as described above. Besides, our framework focus on non-IID problems.

\section{Transfer Learning}
%what is TL?
Transfer Learning is a machine learning technique which aims at enhancing the performance of target models developed on target domains by reusing the knowledge contained in diverse but related models developed on source domains \cite{bozinovski2020reminder}.

The definition of Transfer Learning is given by Pan \etal\ in the survey \cite{pan2009survey}. In this survey, they use binary document classification as a description example. The definitions of a \textit{domain} and a \textit{task} should be explained.
A domain $D$ has two parts, i.e. a feature space $X$ and a marginal probability distribution $P(X)$ with $X=\{x_1,x_2,...,x_n\}\in X$. In the example of binary document classification, $X$ consists of vectors and is the space of all representations, $x_i$ is the $i$th term vector according to some document, and $X$ is the sample for training. 
A task consists two parts corresponding to a give domain, $D=\{X, P(X)\}$, i.e. a label space $Y$ and a conditional probability distribution $P(Y|X)$ which is obtained from training data, which is in the form of pairs $\{x_i, y_i\}$. For the set of labels $Y$,  a element in the binary classification is either True or False \cite{pan2009survey}.
Considering a source domain $D_S$ and a source task $T_S$, and a pair of corresponding target domain $D_T$ and target task $T_T$, the goal of Transfer Learning is to learn a conditional probability distribution $P(Y_T|X_T)$ in $D_T$ with the samples obtained from $D_S$ and $T_S$ where $D_S\neq D_T$ and $T_S \neq T_T$. It is generally assumed that the number of available labeled target examples is limited, which is exponentially smaller than the number of labeled source examples. In addition, there are explicit or implicit between the feature spaces of target and source domains, when it exists some relationship \cite{pan2009survey}.

%variants of TL 
Moreover, Pan \etal\ give the specific variants of Transfer Learning \cite{pan2009survey}. According to the questions: which part of the knowledge can be transferred across domains or tasks (What to transfer); which machine learning should be chosen to Transfer Learning, after pointing out which knowledge can be transferred(how to transfer); in which scenario can transferred be applied (when to transfer), it is divided into three categories:inductive Transfer Learning, transductive Transfer Learning, and unsupervised Transfer Learning, respectively \cite{pan2009survey}. 

\begin{table}[H]
	\centering
	\begin{tabular}{|l|l|l|}
		\hline  
		\textbf{Classes of Transfer Learning} & \textbf{Details} \\
		\hline 
		Inductive Transfer Learning \cite{pan2009survey} & \makecell[l]{Which part of the knowledge can be \\ transferred across domains or tasks \\ (What to transfer)?}\\
		\hline
		Transductive Transfer Learning \cite{pan2009survey} & \makecell[l]{ Which machine learning should be chosen \\to Transfer Learning, after pointing out \\ which knowledge can be transferred \\(how to transfer)?}\\
		\hline
		Unsupervised Transfer Learning \cite{pan2009survey} & \makecell[l]{Which scenario can transferred be applied \\ (when to transfer)?}\\
		\hline 
	\end{tabular}
	\caption{Classes of Transfer Learning \cite{pan2009survey}.}
	\label{tab:table_2_2}
\end{table}

The specific explanations of three categories are following: 

%\begin{enumerate}
\textbf{Inductive Transfer Learning.} In the Inductive Transfer Learning, no matter if the target and source domains are the same or not, the target task differs from the source task. This inductive Transfer Learning tries to utilize the inductive bias of the source domain to help improve the performance of target tasks. Corresponding to whether the data from the source domain is labeled, this type of Transfer Learning can be further classified into two subdivisions, i.e. multitask learning and self-supervised (self-taught) learning \cite{pan2009survey}. 

\textbf{Transductive Transfer Learning.} In the Transductive Transfer Learning, there are similarities between the source and target tasks, but the differences exist in corresponding domains. Besides, there is no labeled data in the target domain, while there are a large amount of labeled data available in the source domain. This can be further divided into two categories, corresponding to scenario with different feature space or different marginal probabilities \cite{pan2009survey}.

\textbf{Unsupervised Transfer Learning.} In the Unsupervised Transfer Learning, the source and target domains are similar, but the source and target tasks are different from each other. This unsupervised Transfer Learning is similar to inductive Transfer Learning, with the focal point on solving Unsupervised Learning tasks in the target domain. In addition, no labeled data exists in both source and target domains \cite{pan2009survey}. 

%\end{enumerate}
%The summarization of different types or scenarios for Transfer Learning is following.
There are four different ways to use Transfer Learning, i.e. instance transfer \cite{10.1145/1273496.1273521}, feature representation transfer \cite{raina2007self}, parameter transfer \cite{lawrence2004learning} and relational knowledge transfer \cite{mihalkova2007mapping} \cite{pan2009survey}. 

\begin{table}[H]
	\centering
	\begin{tabular}{|l|l|l|}
		\hline  
		\textbf{Approach Index} & \textbf{Approaches to Use Transfer Learning} \\
		\hline 
		\rom{1} & Instance transfer \cite{10.1145/1273496.1273521}\\
		\hline
		\rom{2} & Feature-representation transfer \cite{raina2007self}\\
		\hline
		\rom{3} & Parameter transfer \cite{lawrence2004learning}\\
		\hline 
		\rom{4} & Relational knowledge transfer \cite{mihalkova2007mapping}\\
		\hline 
	\end{tabular}
	
	\caption{Criterion for classification of data fusion techniques \cite{castanedo2013review}\cite{zhang2020multimodal}.}
	\label{tab:table_2_3}
\end{table}

%\begin{enumerate}
\textbf{Approach \rom{1}.} For instance transfer, the main purpose is to reuse knowledge from the source domain to the target domain. Although the source domain cannot be reused directly, particular parts of the data can be extended along with target data in the target domain for training by re-weighting \cite{pan2009survey}.

\textbf{Approach \rom{2}.}  For feature-representation transfer, the main idea is to scale up the convergence and performance through recognizing useful feature representations, which are able to be used from source to target domains. According to whether data is labeled, feature-representation transfer can be used based on supervised or Unsupervised Learning \cite{pan2009survey}.

\textbf{Approach \rom{3}.} The main intuition is to share prior probability distribution of parameters or even some parameters, which is based on the assumption of related tasks \cite{pan2009survey}. 

\textbf{Approach \rom{4}.} For relational-knowledge-transfer, the main goal is to deal with data that is not independent and identically distributed. That is to say, there is a relationship between data nodes. For example, the web information of social network can use relational-knowledge-transfer methods \cite{pan2009survey}.

Transfer Learning has a big success in the machine learning area. In image classification problems, the paper from Wu \etal\ has shown that additional data extracted from a different distribution can help the main classification learners greatly improve the performance \cite{wu2004improving}. In natural language processing, Raina \etal\ present that an approach for learning a mapping covariance matrix from additional labeled text can help original classifier improve the classification accuracy \cite{raina2006constructing}. Zhuo \etal\ present that problems not using the existing domain for transfer are worse than the action model learned by Transfer Learning \cite{zhuo2008transferring}.

% maybe we can add more specific steps into the algorithms.
Especially, Transfer Learning is beneficial to image field. Digital medical image is a helpful technique for computer-aided diagnosis. Medical images are difficult to collect and their quantity is limited because medical data are generated by special techniques (e.g. X-ray radiography). Therefore, Transfer Learning can be used as an auxiliary diagnostic tool. There are many successful applications. Maqsood \etal\ use Transfer Learning technique to detect Alzheimer’s disease by fine-tuning AlexNet \cite{krizhevsky2012imagenet}. The proposed approach gets the highest accuracy rate in the experiments of Alzheimer’s stage detection. For this Alzheimer’s stage, the medical images (MRI) is preprocessed by using a contrast stretching operation in the target domain. Next, the AlexNet architecture is pre-trained (source domain) as the start of learning new tasks. Then, the last three layers (one softmax layer, one fully connected layer, and one output layer) of AlexNet is replaced, but the previous convolutional layers are reserved. At last, the fine-tuned training on Alzheimer’s dataset \cite{maqsood2019transfer} in the target domain is carried out using the modified AlexNet.

According to the same physical natures between Electromyographic (EMG) signals from the muscles and Electroencephalographic (EEG) brainwaves, Bird \etal\ \cite{bird2020cross} utilize Transfer Learning from the gesture recognition domain to the mental state recognition domain. It also shows that EEG brainwaves can be transferred to classify EMG signals. The experimental results show that Transfer Learning is helpful to improve the performance of neural network classifiers \cite{bird2020cross}. 

% need to rewrite the paper
\textbf{Negative Transfer.} However, negative transfer may happen. By negative transfer we mean that the learned knowledge contributes to the decreased performance of learning in the new knowledge. That is, the performance of learning from target domain could decrease due to the source domain data and task \cite{pan2009survey}. The experimental results have shown that if two tasks are extremely different, naively applying transfer technique may cause the accuracy loss of target task \cite{smith2001transfer}.
The paper \cite{ben2003exploiting} can provide us direction to avoid negative transfer, and the most important tool is task clustering.
The similar tasks should be gathered using task clustering techniques, when data is clustered regarding to task models. Moreover, the learning tasks can be split into different groups. Each group of tasks is relevant to a low-dimensional feature, and different groups hold different low-dimensional features. Finally, efficient knowledge transfer is done within each group \cite{pan2009survey}.

%%%%%Rewrite this graph
\textbf{Transfer Learning in Our framework.} The Transfer Learning in our framework belongs to inductive Transfer Learning, because Contrastive Learning is one of the most representative approach of Self-Supervised Learning. Besides, Contrastive Learning also tries to minimize the contrastive loss of the source domain (multi modalities) to help improve the performance of target domain (one single modality). The transfer approach in our framework applies for feature-representation transferring, because the Contrastive Learning tasks aim to learn the feature representation from unlabeled data.

\section{Summary}
% Federated Learning, Transfer Learning and Self-Supervised Learning
%In general, the selected multimodal dataset with image-audio multi-modality... 
To evaluate the effectiveness of our framework, we apply our framework to a scene classification dataset with visual-auditory modalities. Because our framework uses multi-modality help uni-modality, multimodal data fusion can be considered. Supervised Learning is applied to learning single modalities. Self-supervision, i.e. Contrastive Learning gives support for learning feature representations from multimodal data. Federated Learning is a important component in our framework, which is to solve data islands problem.

%*****************************************
\chapter{Related Work}
\label{ch:relatedwork}
%*****************************************
%\hint{This chapter should give a comprehensive overview on the related work done by other authors followed by an analysis why the existing related work is not capable of solving the problem described in the introduction.}
% Discusses other related work. This should not only summarize the existing work, but also discuss how your present paper differs from it and why this is a 
We survey the related work covering Federated Transfer Learning, multimodal Transfer Learning and self-supervision in multimodal data. Different methods for enhancing privacy protection or data security are proposed in the Federated Transfer Learning framework. Existing Transfer Learning methods in the Federated Learning framework are introduced, for example, using a pre-trained model. We also introduce multimodal Transfer Learning is a fusion way between different machine learning models and transfers knowledge between different parts of the network for different modalities of inputs. 

\section{Federated Transfer Learning} % (see Figure 3.1)
%Use which approach to do, which difference is ?
Federated Transfer Learning is a variant of Federated Learning \cite{yang2019federated}. Federated Transfer Learning utilizes different datasets which are neither identical in sample space nor in feature space. 

%\begin{figure}[H]
%	\centering
%	\includegraphics[width=0.8\textwidth]{images/FTL.pdf}  	  	 
%	\caption{Federated Transfer Learning Framework \cite{yang2019federated}.}
%	\label{fig:figure_3_1}	
%\end{figure}

The initial goal of Federated Learning is to carry out useful machine learning methods in multiple devices, and build an aggregated model based on dataset across multiple devices, while ensuring users' privacy and security \cite{konevcny2016federated1}. Thus, we should obey the original goal of federated (transfer) learning. Although they are not chosen by us, we briefly review some other alternative for security and privacy in machine learning (see Table 3.1) for completeness.

\begin{table}[H]
	\centering
	\begin{tabular}{|l|l|l|}
		\hline  
		\textbf{Security Solution Index} & \textbf{Methods} \\
		\hline 
		\rom{1} & MPC \cite{sharma2019secure}, SPDZ \cite{damgaard2013practical} \\
		\hline
		\rom{2} & DDPG (S-TD3) \cite{maurya2021federated}\\
		\hline
		\rom{3} & HFTL \cite{gao2019privacy}\\
		\hline 
	\end{tabular}
	\caption{Security solutions for Federated Transfer Learning.}
	\label{tab:table_3_1}
\end{table}

\textbf{Security Solution \rom{1}.} The paper \cite{liu2020secure} is a start in federated machine learning, and this paper is focused on data security in a multi-party privacy-preserving setting. The similar focus is also in the paper \cite{sharma2019secure}, which emphasizes the use of multi-party computation (MPC) can improve efficiency by an order of magnitude in semi-honest security settings. In the paper \cite{sharma2019secure}, the authors use the SPDZ \cite{damgaard2013practical} protocol, which is an implementation of MPC, and experimental results proved that the protocol outperforms homomorphic encryption (HE) in terms of communication and time under the same Federated Transfer Learning framework. 

\textbf{Security Solution \rom{2}.} Further, the authors in \cite{maurya2021federated} proposed a new method of authentication and key exchange protocol to provide an efficient authentication mechanism for Federated Transfer Learning blockchain (FTL-Block), which uses the Novel Supportive Twin Delayed DDPG (S-TD3) algorithm \cite{maurya2021federated}. The participants implement the authentication in each part, which completely relies on the credit of the participants. This authentication mechanism can integrate the users' credit with both local credit and cross-region credit. With the help of S-TD3 algorithm, the training of the local authentication model achieves the highest accuracy \cite{maurya2021federated}. Transfer Learning method is applied to reduce the extra time cost in the authentication model, and domain the authentication model can be accurately and successfully migrated from local to foreign users \cite{maurya2021federated}. Moreover, Majeed \etal\ \cite{majeed2021cross} proposed the cross-silo secure aggregation technique based on MPC for secure Federated Transfer Learning.

\textbf{Security Solution \rom{3}.} In addition, the existing approaches mainly focus on homogeneous feature spaces, which will leak privacy when dealing with the problems of covariate shift and feature heterogeneity. Thus, Gao \etal\ \cite{gao2019privacy} provide a privacy preserving Federated Transfer Learning framework for homogeneous feature spaces called heterogeneous Federated Transfer Learning (HFTL). Specifically, Gao \etal\ designed privacy-preserving Transfer Learning method to remove covariate shifts in homogeneous feature spaces, and connect heterogeneous feature spaces from different participants \cite{gao2019privacy}. Besides, two variants based on HE and secret sharing techniques are applied in the HFTL. Experimental results have demonstrated that the framework performs general feature-based Federated Learning methods and self-learning methods under the same challenge constraints, and HFTL also is shown to have practical efficiency and scalability \cite{gao2019privacy}.
% other perspective of different tranfser learning methods settings. Reorder the different applications based on different Transfer Learning methods. make conclusion.

\textbf{Our Framework without Additional Security Solutions.} To conclusion, all the these security solutions \cite{sharma2019secure} \cite{damgaard2013practical} \cite{maurya2021federated} \cite{gao2019privacy} added to add additional security protection mechanisms. Since our primary goal is to investigate Federated Learning itself, we do not incorporate any of the mechanisms above, although we believe our solution and framework can benefit from them in security, too.

There are some successful applications based on different Transfer Learning methods. For edge devices, tiny devices like mobile phones, medical instruments, smart manufacturing, etc. \cite{chen2020fedhealth} \cite{yang2020fedsteg} \cite{ju2020federated}. There are two strategies are used in Federated Transfer Learning: \textit{Pre-trained models} can be reused in related tasks, while \textit{domain adaptation} can be used from a source domain to a related target domain. A \textit{pre-trained} model can be used directly in some Federated Transfer Learning frameworks. Domain adaptation is also known as the knowledge transferring from the source domain to the target domain \cite{bozinovski2020reminder}. 

\begin{table}[H]
	\centering
	\begin{tabular}{|c|c|c|}
		\hline  
		\textbf{Index of Applications for Pre-trained Models} & \textbf{Applications} \\
		\hline 
		\rom{1} & FedHealth2 \cite{chen2020fedhealth} \\
		\hline
		\rom{2} & FedURR \cite{gao2021fedurr} \\
		\hline
		\rom{3} & FedDCSCN \cite{zhang2022federated}\\
		\hline 
	\end{tabular}
	\caption{Applications with pre-trained models in Federated Transfer Learning.}
	\label{tab:table_3_2}
\end{table}

\textbf{Application \rom{1} with Pre-trained Models.} After FedHealth \cite{chen2020fedhealth}, Chen \etal\ \cite{chen2021fedhealth} proposed FedHealth2, a weighted Federated Transfer Learning framework through batch normalization . In FedHealth2, all participants aggregate the features without compromising privacy security, and obtain local models for participants via weighting and protecting local batch normalization \cite{chen2021fedhealth}. More specifically, FedHealth2 achieves the similarities in participants with the support of a pre-trained model. The similarities are confirmed by the metrics of the data distributions, and the metrics can be determined by outputs values of the pre-trained model. Then, with the achieved similarities, the server can do the averaging of the weighted models' parameters in a localized approach and produce a unique model for each participant \cite{chen2021fedhealth}. Experimental results have shown that FedHealth2 enables the local participants' models to do the recognition with higher accuracy. Besides, FedHealth2 can achieve similarity in several epochs even if no pre-trained model exists \cite{chen2021fedhealth}.

\textbf{Application \rom{2} with Pre-trained Models.} Moreover, FedURR \cite{gao2021fedurr} also benefits from the pre-trained model. In Urban risk recognition (URR) task, the urban management usually has multiple departments, each of which stores a large amount of data locally. When data is uploaded to a central database, it means huge cost and a lot of time consumption, and there exists a risk of data leakage \cite{gao2021fedurr}. Thus, the proposed framework FedURR integrates two types of Transfer Learning into the Federated Learning framework, i.e. fine-tuning based and parameter sharing based Transfer Learning methods. With the help of fine-tuning and parameter sharing, they are connected into different stages of Federated Learning with an precisely design. The experimental results are shown that FedURR can improve multi-department collaborative URR accuracy \cite{gao2021fedurr}.

%Compared to the FedSteg, we both use convolutional neural network architecture as a basic model. There are also differences. In this thesis, transfer knowledge is performed one individual user, while FedSteg performs knowledge transfer between the cloud server and the users (clients).\\

%Federated Transfer Learning for disaster classification in social computing networks \cite{zhang2022federated}
\textbf{Application \rom{3} with Pre-trained Models.} Furthermore, Zhang \etal\ \cite{zhang2022federated} proposed the first Federated Transfer Learning framework, to solve problems in Disaster Classification in Social Computing Networks (FedDCSCN). The authors want to eliminate shortcomings of the local models of the participants, which are deep learning models. The local models need a large number of high-quality samples, and fast computation speed is required to accelerate the training process \cite{zhang2022federated}. In addition, the data labeling process is time consuming in the field of social computing, which hinders the use of deep learning networks \cite{zhang2022federated}. Thus, Federated Learning and Transfer Learning are combined to address the problems. Pre-trained model based Transfer Learning is used as to reduce communication and computation costs \cite{zhang2022federated}. Besides, homomorphic encryption approach is applied as a additional to preserve the local data privacy of social computing participants \cite{zhang2022federated}. Experimental results are shown that a feasible but not ideal performance is obtained by the framework in the social computing field \cite{zhang2022federated}.

% add one more applications

\begin{table}[H]
	\centering
	\begin{tabular}{|l|l|l|}
		\hline  
		\textbf{Index of Applications with Domain Adaptation} & \textbf{Applications} \\
		\hline 
		\rom{1} & FedSteg \cite{yang2020fedsteg} \\ 
		\hline
		\rom{2} & Fedhealth \cite{chen2020fedhealth} \\
		\hline
		\rom{3} & EEG signal classification \cite{ju2020federated}\\
		\hline 
	\end{tabular}
	\caption{Applications with domain adaptation in Federated Transfer Learning.}
	\label{tab:table_3_3}
\end{table}

%FedSteg: A Federated Transfer Learning Framework for Secure Image Steganalysis \cite{yang2020fedsteg} which specific Transfer Learning method? knowledge 
\textbf{Application \rom{1} with Domain Adaptation.} FedSteg \cite{yang2020fedsteg} provides an example for using domain adaptation.
Image steganography is the method of concealing secret information within images. Conversely, image steganalysis is a counter method to image steganography. This method intends to detect the secret information within images. Through this detection technique, the steganographic features which are generated by image steganographic methods can be extracted. However, there are still problems that exist in image steganalysis. Image steganalysis algorithms train on machine learning models which rely on a large amount of data. However, it is hard to aggregate all the steganographic images to a global cloud server.
Moreover, the users do not want unrelated people to snoop on confidential information. To solve the problems, Yang \etal\ propose the framework called FedSteg. FedSteg trains a machine learning model with a privacy-protecting technique through domain adaptation. Domain adaptation is used to train the local model by decreasing the domain discrepancy between the global server and local data. Compared with traditional non-federated steganalysis techniques, the experiment results show that FedSteg achieves certain improvements \cite{yang2020fedsteg}.

% FedHealth: A Federated Transfer Learning Framework for Wearable Healthcare \cite{chen2020fedhealth}
\textbf{Application \rom{2} with Domain Adaptation.} Fedhealth \cite{chen2020fedhealth} benefits from domain adaptation. Wearable devices allow people to get access to and record healthcare information. Additionally, smart wearable devices use a large amount of personal data to train machine learning models. Different wearable devices have diverse characteristics and domains. However, the healthcare data from different people with diverse monitoring patterns are difficult to aggregate together to generate robust results. Each personal data is an island. Besides, the machine models using personal data are hard to train on cloud servers. To solve data isolation and locally training problem, Chen \etal\ proposed a Federated Learning framework called FedHealth \cite{chen2020fedhealth}. In this paper, the authors used a neural network (NN), which has two convolutional layers, two pooling layers, and three fully-connected layers \cite{chen2020fedhealth}. NN aims at extracting low-level features. Domain adaptation is applied to transfer the extracted features from server to clients by minimizing the feature distance between server and clients. Compared to the approaches without Federated Learning and traditional methods (KNN, SVM, and RF), FedHealth achieves better performance \cite{chen2020fedhealth}.

% Federated Transfer Learning for EEG Signal Classification \cite{ju2020federated}
\textbf{Application \rom{3} with Domain Adaptation.} One more example that uses domain adaptation is the electroencephalographic (EEG) signal classification \cite{ju2020federated}. Brain-Computer Interface (BCI) systems are mainly to identify the users' consciousness from the brain states. Deep learning methods achieve success in the BCI field for classification of EEG signals. However, the success is restricted to the lack of a large amount of data. Besides, according to the privacy of personal EEG data, it is constrained to build a collection of big BCI dataset. In order to solve the lack of data and the private privacy problems, Ju \etal\ proposed a Federated Transfer Learning method for EEG Signal classification. They propose an method which use Transfer Learning technique with domain adaptation to extract the common discriminative information, and map the common discriminative information into a spatial covariance matrix, then subsequently fed the spatial covariance matrix to a deep learning based Federated Transfer Learning architecture \cite{ju2020federated}. The proposed architecture based on deep learning has 4 layers, namely Manifold reduction layer (M), Common embedded space (C), Tangent projection layer (T) and Federated layer (F), the middle two layers (M and T) provide the functionality of Transfer Learning \cite{ju2020federated}. The experimental result shows that this method using domain adaption in Federated Learning architecture has robust generation ability.

There are two special cases of the problems to be solved in the heterogeneous Federated Transfer Learning setting, and one case for quantifying the performance of Federated Transfer Learning.

% FedMD: Heterogeneous Federated Learning via Model Distillation \cite{li2019fedmd}
\textbf{Model Distillation.} FedMD \cite{li2019fedmd} provides a way to solve statistical heterogeneity (the non-IID problem) in Federated Transfer Learning. Concretely, the authors in FedMD focus on the differences of local models \cite{li2019fedmd}. The authors in FedMD identify that communication is the key to fix model heterogeneity. Devices should have the ability to learn the communication protocol to leverage Transfer Learning and model distillation. The communication protocol aims to reuse the models, which are trained from a public dataset. Each client achieves a well-trained model,  and applies the well-trained model on local data which is considered as Transfer Learning with model distillation. Thus, the proposed FedMD, which combines Federated Learning and Transfer Learning with knowledge distillation, allows participants to create their models locally, and a communication protocol that utilizes the power of Transfer Learning with model distillation \cite{li2019fedmd}. FedMD is demonstrated its efficiency to work on different tasks and datasets \cite{li2019fedmd}.

% Heterogeneous Defect Prediction Based on Federated Transfer Learning via Knowledge Distillation \cite{wang2021heterogeneous}
\textbf{Knowledge Distillation.} Wang \etal\ \cite{wang2021heterogeneous} propose Federated Transfer Learning via Knowledge Distillation (FTLKD), which is a robust centralized prediction framework, and is used to solve data islands and data privacy. This framework helps participants to do heterogeneous defect prediction (HDP), predict the defect tendency regarding private models. Concretely, a pre-trained model of public datasets is transferred to the private model, and the model on the private data to converge by fine-tuning, and then the final output in each participant's private model is conveyed through knowledge distillation \cite{wang2021heterogeneous}. Besides, HE is used to encrypt data without disturbing the processing results. Experimental results on 9 projects in 3 public databases (NASA, AEEEM and SOFTLAB) show that FTLKD outperforms the related competing methods \cite{wang2021heterogeneous}.

% 'Quantifying the Performance of Federated Transfer Learning'\cite{jing2019quantifying}
\textbf{Quantifying Performance.} In addition, the authors in \cite{jing2019quantifying} analyze \textbf{three} major bottlenecks in Federated Transfer Learning and their potential solutions. The main bottleneck is inter-process communication. Data exchange and memory copy in a device can cause extremely high latency. JVM native memory heap and UNIX domain sockets give us the opportunity to alleviate the type of bottlenecks \cite{jing2019quantifying}. The second bottleneck is in the additional encryption tool that increases computational cost. The last is the traditional congestion control problem. Intensive data exchange causes heavy network traffic \cite{jing2019quantifying}.

%一句带过 domain adaptation, 重新总结self-supervised Federated Learning
%Federated Self-Supervised Learning of Multi-Sensor Representations for Embedded Intelligence\cite{saeed2020federated}
\textbf{Transfer Learning Strategies for Our Framework.} The secure methods from papers \cite{sharma2019secure} \cite{damgaard2013practical} \cite{maurya2021federated} \cite{gao2019privacy}, show that these are additionally add to Federated Transfer Learning meanwhile keep its original structure. The paper \cite{jing2019quantifying} shows that additional secure methods bring a bottleneck to Federated Transfer Learning. Thus, we need no additional security methods but keep the original structure of Federated Transfer Learning. The methods with Pre-trained models \cite{chen2020fedhealth} \cite{gao2021fedurr} \cite{zhang2022federated} can be considered in our framework. The applications \cite{yang2020fedsteg} \cite{chen2020fedhealth}\cite{ju2020federated} show that domain adaptation can be used to transfer features from server to participants. Methods \cite{li2019fedmd}\cite{jing2019quantifying} can be considered to solve problems where the data distributions of participants are different. However, these two methods require an additional public dataset. In conclusion, the methods of transferring from pre-trained models and strategies with domain adaptation can be considered in our framework.

\section{Multimodal Transfer Learning}
Multimodal Transfer Learning has a wide range of application in multi-modality applications \cite{qi2018unified}. 
The main purpose of multimodal Transfer Learning is to use the diversity of modalities to improve performance, and to take advantage of multi-modality with more feature space under the condition of limited data.

The recent methods of multimodal Transfer Learning are focused on the transferring between two partial deep neural networks. There are two main strategies about multimodal Transfer Learning: \textit{multimodal transfer module} and \textit{multimodal domain adaptation}. Multimodal transfer module works as a component, which is added between layers of two multimodal sub-networks \cite{joze2020mmtm}. The goal of multimodal domain adaptation focus on learning transferable representations \cite{qi2018unified}. This goal is consistent with the motivation of our framework. Following gives us applications of multimodal Transfer Learning.

% table 描述在多模态之间是怎么转换的
\begin{table}[H]
	\centering
	\begin{tabular}{|l|l|l|l|}
		\hline  
		\textbf{Strategies} & \textbf{Index of Methods} & \textbf{Methods} & \textbf{Scenarios}\\
		\hline 
		\multirow{1}{*}{Multimodal Transfer Module} 
		& \rom {1} & MMTM  \cite{joze2020mmtm} & \makecell[l]{Sign Language Recognition\\ Action Recognition\\
			Speech Enhancement} \\
		\hline 
		\multirow{6}{*}{\makecell[l]{Multimodal\\ \\ Domain\\ \\ Adaptation}} 
		& \rom {1}.\rom{1} & MDANN \cite{qi2018unified} & \makecell[l]{Emotion Recognition\\ Cross-Media Retrieval}\\ \cline{2-4}
		 & \rom {1}.\rom{2} & ADA \cite{li2018multimodal} & Vigilance Estimation \\ \cline{2-4}
%		\hline 
		 & \rom {1}.\rom{3} & MM-SADA \cite{munro2020multi} & Vigilance Estimation \\ \cline{2-4}
%		\hline 
		 & \rom {1}.\rom{4} & DLMM \cite{lv2021differentiated} & \makecell[l]{Event Recognition\\ 
		 Action Recognition} \\ \cline{2-4}
%		\hline 
		 & \rom {1}.\rom{5} & PMC \cite{zhang2021progressive} & Visual Recognition\\ \cline{2-4}
%		\hline 
		 & \rom {1}.\rom{6} & JADA \cite{li2019joint} & Image Classification\\ \cline{2-4}
		\hline
	\end{tabular}
\end{table}

% MMTM
\textbf{MMTM.} Joze \etal\ propose a multimodal fusion component for cross-modal fusion of convolutional neural network modules \cite{joze2020mmtm}. This component is called \textit{Multimodal Transfer Module} (MMTM). This module has two important operations squeeze and excitation, mainly for the channel level, and it is not sensitive to spatial information. There is the size of feature map here, because the channel information is aimed at the features of a channel, which will be condensed into a little information. Besides, this component MMTM can be flexibly added in two modal information layers. So it does not affect the initialization of each modal. This module can be applied in, for example, sign language recognition, action recognition, speech enhancement (one of which is to use mouth shape to suppress the influence of surrounding noise to enhance the speaker's voice), etc.

%A unified framework for multimodal domain adaptation \cite{qi2018unified}
%MDANN
\textbf{MDANN.} Traditional domain adaptation methods are to mitigate the domain gap by assuming both the source and target domains, which have the same single modality. Qi \etal\ proposed a framework to deal with the domain shift problems which always involve multiple modalities \cite{qi2018unified}. It is the framework to answer the question: "Is it possible to design a domain adaptation algorithm that can transfer multimodal knowledge from one multimodal dataset to another one?" \cite{qi2018unified} The proposed framework is called \textit{Multimodal Domain Adaptation Neural Networks} (MDANN). It includes three modules:(1)A covariant multimodal attention aims to learn common feature representations for multiple modalities. (2)A fusion module adaptively combines joined features of diverse modalities. (3)Mixed domain limits are proposed to completely understand domain-invariant features by limiting individual modal features, fused features, and attention scores \cite{qi2018unified}. By co-engaging and fusing under adversarial objectives, the most distinctive and domain-adaptive parts of the features are fused together. Experimental results on two real-world cross-domain applications (emotion recognition and cross-media retrieval) confirm the effectiveness of MDANN.

%ADA Multimodal Vigilance Estimation with Adversarial Domain Adaptation Networks; 
\textbf{ADA.} Multimodal vigilance estimation is another instance of the success in multimodal domain adaptation. For multimodal vigilance estimation, the popular used approaches require collecting sufficient subject-specific labeled data for calibration. However, it is expensive for practical applications. In order to solve the problem, Li \etal\ \cite{li2018multimodal} employed two recently proposed \textit{Adversarial Domain Adaptation Networks} (ADA), and compared their performance with some traditional domain adaptation methods and a baseline with no domain adaptation method. In comparison with these methods, experimental results show that the two ADA networks applied on multimodal vigilance estimation achieves better performance.

%MM-SADA Multi-Modal Domain Adaptation for Fine-Grained Action Recognition;
\textbf{MM-SADA.} Similar to MDANN, fine-grained action recognition is also an application of multimodal domain adaptation. Munro \etal\ \cite{munro2020multi} propose \textit{Multi-Modal Self-Supervised Adversarial Domain Adaptation} (MM-SADA) to address the inevitable domain shift problem, which training a model in one environment and then deploying it in another leads to performance degradation. In addition to domain adaptation, i.e. adversarial alignment, Munro \etal\ also leverage the correspondence of modalities as a self-supervised alignment method for Unsupervised Learning \cite{munro2020multi}. More specifically, they use self-supervision technique between different modalities in one sample, and domain adaptation (adversarial alignment) is used in the same modality \cite{munro2020multi}. Experiment results show MM-SADA outperforms other unsupervised domain adaptation methods.

%Differentiated learning for multi-modal domain adaptation \cite{lv2021differentiated}
\textbf{DLMM.} The above multimodal domain adaptation methods deal with each modality in the same way and simultaneously do the sub-networks' optimization for diverse modalities. However, in the real world, the measurement of domain shift in diverse modalities are different \cite{lv2021differentiated}. Lv \etal\ \cite{lv2021differentiated} proposed a \textit{Differentiated Learning Multimodal domain adaptation} framework called DLMM that takes advantage of difference between modalities to make domain adaptation more effective. In order to measure the reliability of each transferred sub-modality on the samples in the target domain, the authors propose a Prototype based Reliability Measurement and a Reliability-aware Fusion schema can be used to help make decision. Experimental results show that DLMM can achieve better performance than the state-of-the-art multi-modal domain adaptation models.

%Progressive modality cooperation for multi-modality domain adaptation \cite{zhang2021progressive}
\textbf{MMDA-PI.} Since it is difficult to determine whether each modality data in the source domain has in the target domain, Zhang \etal\ \cite{zhang2021progressive} propose a new generic multimodal domain adaptation framework called Progressive Modality Cooperation (PMC), which is general with two different settings, namely multi-modality domain adaptation (MMDA) and \textit{Multi-Modality Domain Adaptation using Privileged Information} (MMDA-PI). In MMDA, all the samples from both the source and target domains have all the modalities. However, in MMDA-PI, the target domain may not exist with some modalities corresponding to the source domain. Besides, the authors propose a multi-modality data generation module (MMG) to fix the missing modality. In MMG, domain distribution mismatching and semantic information keeping problems are considered. Experimental results on different multimodal datasets show that the framework holds the effectiveness, generalization and robustness in MMDA and MMDA-PI settings.
%Cross-Domain First Person Audio-Visual Action Recognition through Relative Norm Alignment\cite{}

\textbf{JADA.} %Shuffle and Attend: Video Domain Adaptation \cite{choi2020shuffle}
The existing research of multimodal domain adaptations aim to solve domain shift problem in domain-level, which focus on extracting transferable feature representations through matching or adversarial adaptation networks. However, these methods ignore label mismatching across domains, which causes inaccurate distribution alignment and affects transferring. Li \etal\ \cite{li2019joint} proposed \textit{Joint Adversarial Domain Adaptation} (JADA) aims to solve label mismatching across domains. The authors apply multimodal domain adaptation based on both domain-wise and label-wise matchings, which learns jointly minimize two types of losses. Experimental results show that JADA can achieve better performance than other state-of-the-art deep domain adaptation approaches \cite{li2019joint}.

%Besides, Zhao \etal\ propose an aggregation approach between unimodality and multimodality, which can also help unimodal data improve performance \cite{zhao2021multimodal}.
\textbf{Multimodal Domain Adaptation for Our Framework.} Multimodal domain adaptation gives us a hint. The core ideas is to learn transferable features representations by matching source and target domain \cite{qi2018unified} \cite{munro2020multi} \cite{lv2021differentiated} \cite{li2019joint}. The idea of \textit{matching} can be used in our framework to extract transferable features.

\section{Self-Supervision with Multimodal Data}
Contrastive Learning is one of the most widely used Self-Supervised Learning methods \cite{chen2020simple} \cite{arandjelovic2018objects} \cite{patrick2020multi} \cite{morgado2021audio} \cite{afouras2020self} \cite{khurana2020cstnet} \cite{tian2019contrastive}. 
Contrastive loss is utilized as a metric to co-train specific modality based sub-networks in Contrastive Learning \cite{chen2020simple}. Contrastive Learning learns features representations from unlabeled multimodal data by itself \cite{arandjelovic2017look}. This motivates our framework, as we can use this method to achieve transferable features. Existing methods are summarized as follows.

\begin{table}[H]
	\centering
	\begin{tabular}{|l|l|l|l|l|}
		\hline  
		\textbf{Index for Contrastive Learning} & \textbf{Methods} & \textbf{Scenarios} & \textbf{Matching Methods}\\
		\hline 
	 	\rom {1} & AVC \cite{arandjelovic2017look} & Audio-Visual & Late Fusion \\
		\hline
	 	\rom {2} & AVE-Net \cite{arandjelovic2018objects} & Audio-Visual & \makecell[l]{Late Fusion \\ Embedding} \\
		\hline
		\rom {3} & CrossCLR \cite{zolfaghari2021crossclr} & Audio-Visual & \makecell[l]{Late Fusion \\ Embedding \\ Inter-Similarity}\\
		\hline
		\rom {4} & AVID \cite{maurya2021federated}& Audio-Visual & Attention Map\\
		\hline
		\rom {5} & LWTNet \cite{afouras2020self} & Audio-Visual & \makecell[l]{Late Fusion\\ Embedding\\ Agreement}\\
		\hline
		\rom {6} & CBT \cite{sun2019learning} & Video-Text & \makecell[l]{Late Fusion with Pre-trained\\ models}\\
		\hline
		\rom{7} & CSTNet \cite{khurana2020cstnet} & Audio-Text& Complex Late Fusion\\
		\hline
	\end{tabular}
	\caption{Applications for multimodal Transfer Learning. The matching column represents different approaches of connecting different modalities.}
	\label{tab:table_3_5}
\end{table}

\textbf{AVC.} Arandjelovic \etal\ \cite{arandjelovic2017look}, which is the first paper using self-supervision technique for co-training visual-audio multimodal data. In \cite{arandjelovic2017look}, a novel Audio-Visual Correspondence (AVC) is proposed, which can be utilized to train visual and audio sub-networks simultaneously in two \textit{views}, and determine the correspondence between these two modalities. The results of experiments show that AVC can localize objects in both modalities in recognition tasks. Besides, AVC is as good as state-of-the-art self-supervised methods in classification tasks.

\textbf{AVE-Net.} Arandjelovic \etal\ proposed Audio-Visual Embedding Network (AVE-Net) \cite{arandjelovic2018objects}, which is another example over audio-visual multimodal data. Instead of using two sub-networks, two embedding level based encoders are used to extract features. More sepecific, Audio with a alternative of 1 second and centralized on the corresponding selected frame, is viewed as a positive pair modalities, otherwise a negative pair is drawn from different videos. This is different from previous method AVC, in which a concatenation based non-linear multi-layer perceptron (MLP) is used for the two embedding features and decides whether the signals are related. The decision distance is also measured through a contrastive loss, which is Euclidean distance based loss. Similarities between embedding level based representations are performed explicitly in \cite{arandjelovic2018objects}, rather than represented features are learned implicitly in a fused MLP in \cite{arandjelovic2017look}. Besides, the embedding representations by AVE-Net has well alignments and are more efficient for cross model retrieval tasks \cite{arandjelovic2018objects}.

\textbf{CrossCLR.} Zolfaghari \etal\ \cite{zolfaghari2021crossclr} proposed Cross-modal Contrastive Learning For Multi-modal Video Representations (CrossCLR). The authors found that the existing contrastive loss functions have not considered the inter-similarities of different modalities. Without inter-similarities, Contrastive Learning leads to inefficient feature representations, when the representations are mapped into embedding spaces. The proposed CrossCLR improves the mapped embeddings. Experimental results show that the generality of CrossCLR by learning joint embeddings for other pairs of modalities.
%Cross-modal Audio Visual Instance Discrimination (Cross-AVID) [72] jointly learn the general representation from video using corresponding image frames and audio segments. In addition to contrasting between audio and visual representations of the same instance, they introduced a Cross-modal Agreement (CMA), a mining method that extends the set of positive pairs beyond just from a single instance. CMA measured the agreement of two videos based on both their visual and acoustic characteristics and if two videos have high agreement in both modalities, they are considered positive pairs.\\

\textbf{AVID.} Morgado \etal\ \cite{morgado2021audio} proposed Audio-Visual Instance Discrimination (AVID) with Cross-model agreement (CMA) , which extract features from image frames together with audio segments. Compared to AVE-Net and AVC, which train just sample pairs contrastive distance loss, CMA is proposed to obtain extra information. CMA is a data mining method, and \textit{vote} for the agreement of two modalities of videos. Two modalities inputs are considered as positive pairs, when they have high \textit{voting} both in image and audio features. AVID creates better positive and negative pairs, and experimental results show that CMA achieves good performance  in downstream tasks using fine-tuning.

% Self-Supervised Learning of audio-visual objects from video \cite{afouras2020self} page-249
\textbf{LWTNet.} Afouras \etal\ \cite{afouras2020self} proposed The Look Who’s Talking Network (LWTNet), which uses synchronization cues to learn the attention map by choosing negative samples with misalignment from random video clips. This method is different from learning the contrastive visual-audio of diverse samples. LWTNet is to solve these three challenges: (i) there are many visually similar but sound different generating objects in a scene (multiple people speak together), and the model must correctly identify the sound to the actual sound source; (ii) these to-be-identified objects may change their position over time; and (iii) there can be multiple other objects in the scene (clutter) as well. More concretely, synchronization cues are measured for each frame by aggregating local scores on each pixel. A maximum spatial response is calculated for cues of each video frame. This spatial response creates the result of a attention map. Experimental results show that LWTNet performs better than other state-of-art Contrastive Learning in audio-visual object detection task.

% \cite{sun2019learning}\cite{devlin2018bert}\cite{tian2019contrastive}
\textbf{CBT.} Sun \etal\ \cite{sun2019learning} proposed a Contrastive Learning based method, which is called Contrastive Bidirectional Transformer (CBT). The existing Contrastive Learning based methods are not only used to train on cross-modal data from scratch, but also can be used to learn the correspondence between pre-trained sub-networks of different modalities. CBT uses a pre-trained Bidirectional Encoder Representations from Transformers (BERT) \cite{devlin2018bert} to deal with automatic speech recognition (ASR) discrete tokens, and split video BERT model to handle continuous features. CBT is verified by Contrastive Representation Distillation (CDR) \cite{tian2019contrastive}, in which a \textit{student} network achieves new knowledge from a pre-trained \textit{teacher} network with encoded positive keys. More specific, the student network queries to match feature representations of the teacher network, while the pre-trained teacher network outputs both positive and negative pairs. Experimental results show that CBT outperforms other Contrastive Learning methods, and achieve significant gains compared with supervised face detection task.

\textbf{CSTNet.} Khurana \etal\ \cite{khurana2020cstnet} proposed Contrastive Speech Translation Network (CSTNet) for learning linguistic representations from speech. CSTNet learns over speech-translation pairs (English speech-text translation in other languages) inputs data. CSTNet is trained with a mixture of two triplet loss terms, which achieves feature representations not only from different modalities but also from different languages. CSTNet is also a novel Contrastive Learning framework for learning speech representation. The experiments over Wall Street Journal (WSJ) dataset \cite{marcinkiewicz1994building} show that CSTNet can extract phonetic information and performs as good as existing methods on downstream phone classification task.
% CSTNet encodes phonetic information as evidenced by the good performance on the downstream phone classification task on the Wall Street Journal dataset.

\textbf{Effectiveness Analysis of Self-Supervision.} Ericsson \etal\ \cite{ericsson2021well} analyze the effectiveness of Self-Supervised Learning with comparison between Supervised Learning and Self-Supervised Learning. There is no best pre-trained Self-Supervised model is suitable for all downstream tasks, the best self-supervised methods can surpass supervised differs from Supervised Learning in terms of the information representations. If the downstream tasks perform on recognition tasks, the performance is seriously correlated. However, when the datasets are irrelevant to recognition, the correlation exists little.

\textbf{Contrastive Learning in Our Framework.} We apply Contrastive Learning in our framework. The core idea of Contrastive Learning is similar to multimodal domain adaptation, which learns transferable features by \textit{matching}. However, multimodal domain adaption focus on \textit{matching} between source and target domain, which Contrastive Learning aims to \textit{matching} between modalities. 

\section{Analysis of Related Work / Summary}

Federated Transfer Learning goes beyond the original purpose of providing better privacy protection. Different Transfer Learning methods (e.g. use pre-trained model) are used as components and transfer knowledge successfully in Federated Learning. However, in Federated Learning, we can not make sure that each the participant holds a pre-trained model, and domain adaptation has the risk of negative transfer \cite{smith2001transfer}. Multimodal Transfer Learning approaches provide the possibility of knowledge transferring between different modalities. However, early and late fusion can suppress either intra-or inter-modality interactions \cite{zhang2020multimodal}, which indicate directly combine two modalities in domain adaptation may cause problems for Transfer Learning. Self-Supervision methods make different modalities connect closely during training, meanwhile the trained multimodal feature can do better knowledge transferring. Therefore, we choose Self-Supervised Learning as one of the fundamental building blocks of our solution.

%*****************************************
\chapter{Design}

\label{ch:design}
%*****************************************
%\hint{This chapter should describe the design of the own approach on a conceptional level without mentioning the implementation details.}
The details of the designed Federated Transfer Learning framework will be explained in this chapter. More specifically, we answer the following questions. Under what requirements and assumptions Federated Transfer Learning framework are built? How many possible solutions for these requirements and assumptions? Which one do we choose? Why we choose the solution? What is the Federated Transfer Learning framework workflow? What are the baseline models associated with the Federated Transfer Learning framework we designed, and how do they work?

\section{Requirements and Assumptions}

% Requirements represent capabilities the solution must have. Assumptions and constraints are fuzzier: they impact the creative process. They are easily forgotten or overlooked.
In this thesis we design a new Federated Transfer Learning framework used for multimodal data with two or more modalities. We assume that some participants hold a part of data with multi-modality, and the others hold a part of data with uni-modality. The purpose of new designed framework is that participants have data with multi-modality help the others have data with uni-modality and protect their privacy.

\begin{figure}[H]
	\centering
	\begin{minipage}[H]{0.30\textwidth}
		\includegraphics[width=1.0\textwidth]{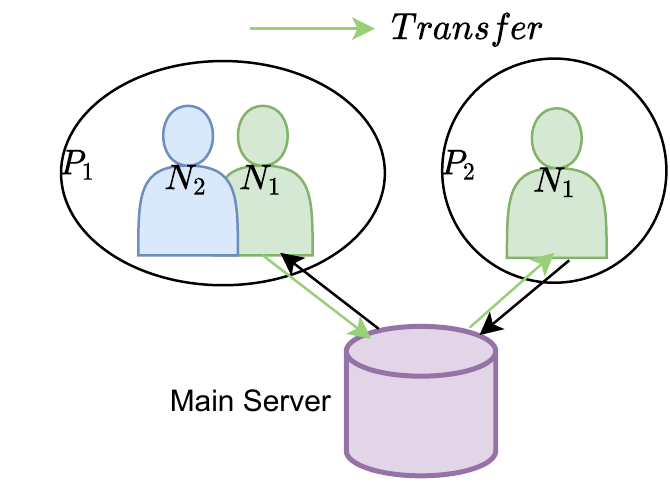}  
		\subcaption{Combination \rom{1}}	  
	\end{minipage}	
	\begin{minipage}[H]{0.3\textwidth}
		\includegraphics[width=1.0\textwidth]{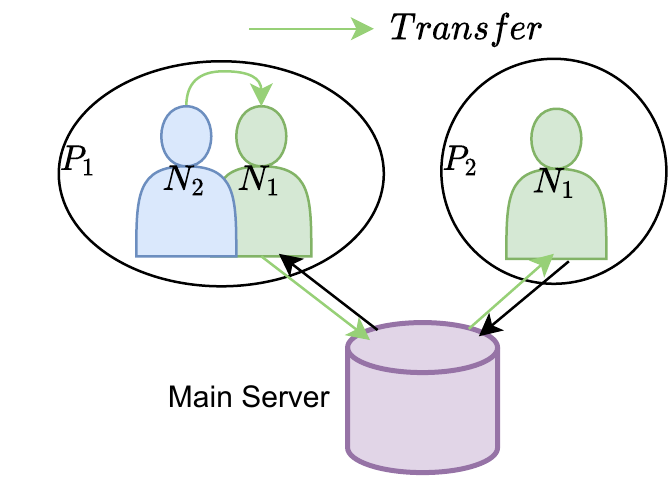}  
		\subcaption{Combination \rom{2}}
	\end{minipage}
	\begin{minipage}[H]{0.3\textwidth}
		\includegraphics[width=1.0\textwidth]{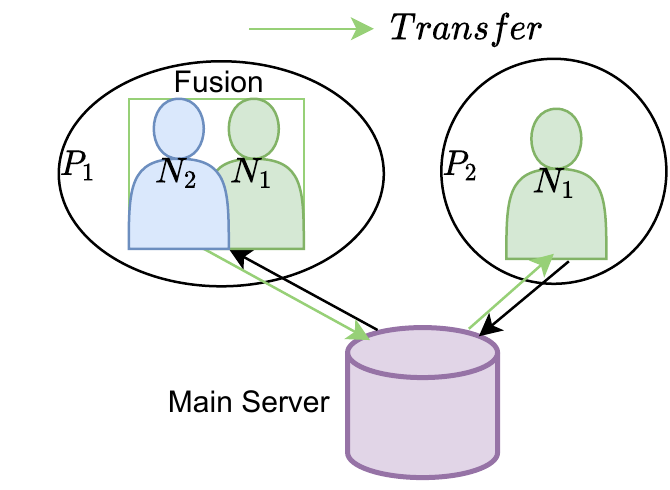}  
		\subcaption{Combination \rom{3}}
	\end{minipage}
	\caption{Three combination ways of Federated Learning and Transfer Learning. The participants grouped as $P_1$ with blue $N_1$ and green $N_2$ indicate that they hold multi-modality. In contrast, the participants grouped as $P_2$ with only green $N_2$ indicate that they has only uni-modality. The arrows represent the models are received and sent. In particular, the green arrows indicate the process of transferring. The main server in purple aggregates models from participants.}
	\label{fig:figure_4_1}
\end{figure}

We have three possible approaches to combine Federated Learning \cite{konevcny2016federated1} and Transfer Learning \cite{bozinovski2020reminder} methods (see Figures \ref{fig:figure_4_1}). We define a set of unimodal datasets $N = \{n_1, n_2, ..., n_m\}$, where $m$ is total number of modalities. The following gives us an overview of different combinations. To explain various combinations more concisely, we define $m$ to 2. Besides, we define two participants, $P_1$ and $P_2$, to clearly describe transferring between participants. 

\begin{table}[H]
	\centering
	\begin{tabular}{|l|l|l|}
		\hline  
		\textbf{Combination Index} & \textbf{Combinations after Grouped Federated Learning} \\
		\hline 
		\rom{1} & From ($N_1$ in $P_1$) transfer to ($N_1$ in $P_2$)\\
		\hline
		\rom{2} & From ($N_2$ in $P_1$) transfer to ($N_1$ in $P_1$), then transfer to ($N_1$ in $P_2$)\\
		\hline
		\rom{3} & Fuse ($N_2$ and $N_1$ in $P_1$), then transfer to ($N_1$ in $P_2$)\\
		\hline 
	\end{tabular}
	\caption{Different combinations of Transfer Learning for different modalities.}
	\label{tab:table_4_1}
\end{table}

The following three figures (Figure \ref{fig:figure_4_1}) illustrate the three combinations of Federated Learning and Transfer Learning. 

In all three combinations, models perform always grouped Federated Learning before the transferring process.\\

\textbf{Combination \rom{1}.} The first combination approach is that feature representations are directly transferred from modality $N_1$ in $P_1$ to modality $N_2$ in $P_2$ after grouped federation. The most representative approaches are to apply pre-trained models \cite{chen2020fedhealth} \cite{gao2021fedurr} \cite{zhang2022federated}.

\textbf{Combination \rom{2}.} The second combination approach is that feature representations are transferred $N_2$ to $N_1$ in $P_1$, then transfer $N_1$ in $P_2$. The representative approach of transferring from $N_2$ to $N_1$ in $P_1$ is MMTM \cite{joze2020mmtm}, in which some connections are added between modalities.

\textbf{Combination \rom{3}.} The third combination approach is that a fusion approach is used to connect all the modalities $N_1$ and $N_2$ in $P_1$, namely co-training sub-networks of their modalities, and then transfer $N_1$ in $P_2$. The representative approaches of multimodal domain adaptation, including MDANN \cite{qi2018unified}, MM-SADA \cite{munro2020multi}, ADA \cite{li2018multimodal}, DLMM \cite{lv2021differentiated}, PMC \cite{zhang2021progressive}, and JADA \cite{li2019joint}, give us a hint to transfer after co-training sub-networks. However, multimodal domain adaptation approaches apply transferring between two multimodal datasets to solve domain shift problem. Domain adaptation approach in Federated Transfer Learning \cite{ju2020federated} give us a direction that all participants can learn a common discriminative information to do better transferring. Besides domain adaptation, Contrastive Learning (Self-Supervised Learning) have the advantages of both co-training sub-networks and learning discriminative information. AVC \cite{arandjelovic2017look}, AVE-Net \cite{arandjelovic2018objects}, CrossCLR \cite{zolfaghari2021crossclr}, AVID \cite{maurya2021federated}, LWTNet \cite{afouras2020self}, CBT \cite{sun2019learning}, CSTNet \cite{khurana2020cstnet} give us support to learn fused feature representations and keep the \textit{matching} information at same time (see Table \ref{tab:table_3_5}). 

We choose \textbf{Combination \rom{3}}. The reasons are: If we use \textbf{Combination \rom{1}}, we can not guarantee that each participant holds a pre-trained model from other tasks. Moreover, FedMD \cite{li2019fedmd} and FTLKD \cite{wang2021heterogeneous} are pre-trained over a public dataset in main sever and then distribute the pre-trained to local participants to their local data. We also can not guarantee that the main server holds a public dataset. If we choose \textbf{Combination \rom{2}}, there are risks of negative transfer \cite{smith2001transfer}. Different modalities with different sub-networks lead to different feature representations and may lead to negative transferring \cite{smith2001transfer}. Besides, it is hard to transfer knowledge when the number of modalities is larger than two. If we choose \textbf{Combination \rom{3}}, we can select Contrastive Learning to learn the \textit{matching} feature representations of each modality from multimodal data. Thus, the best choice is \textbf{Combination \rom{3}} in all combinations.

\section{Important Components}
We choose Contrastive Learning as our Transfer Learning strategy. From the related work of Contrastive Learning, AVC \cite{arandjelovic2017look} is the first paper using Contrastive Learning with an image-audio dataset. AVC applies contrastive loss to learn \textit{matching} between different modalities. The contrastive loss of AVC is similar as it in SimCLR \cite{chen2020simple}, which is calculated after fusing different modalities. AVE-Net \cite{arandjelovic2018objects} goes one step further based on AVC. This step is embedding \cite{arandjelovic2018objects}, which is used to extract features after fusing. However, the authors in CrossCLR \cite{zolfaghari2021crossclr} analyzed that the embeddings ignore the inter-similarities between modalities. Thus, we choose the contrastive loss function in SimCLR to learn representative features from multimodal data. (refer to Table \ref{tab:table_3_5})\\

To describe our design, we first introduce the important component, Contrastive Learning. We define a multimodal dataset $X$ with modalities $N = \{n_1, n_2, ..., n_m\}$, where $m$ is the number of modalities. 
%To clear describe our design, we still define $m=2$. 
The Contrastive Learning in our design is as follows.

\begin{enumerate}
	\item\textbf{Data Preprocessing.} In our framework, we consider a multimodal sample is $x\in X$, a sample has different modalities (\textit{views}), we split a sample into different \textit{views}. Then, we use different data augmentations to preprocess different views of samples. The augmented sample is $\widetilde{x}$ and the samples of different modalities are $\widetilde{x}_{n_1}$, $\widetilde{x}_{n_2}$, ..., and $\widetilde{x}_{n_m}$.

% need to describe encoders.
	\item\textbf{Encoders $f_N$.} Encoder (stacked convolutional layers) is the main components in Contrastive Learning \cite{chen2020simple}. Encoders can be different neural networks for modalities $N = \{n_1, n_2, ..., n_m\}$. Then, each modality applies a neural network. The representations are $h_{n_1}=f_{n_1}(\widetilde{x}_{n_1})$ for $\widetilde x_{n_1}$, $h_{n_2}=f_{n_2}(\widetilde{x}_{n_2})$ for $\widetilde x_{n_2}$, ..., and $h_{n_m}=f_{n_m}(\widetilde{x}_{n_m})$ for $\widetilde x_{n_m}$.
%In a video sample with multi-modality, we use VGG16 \cite{simonyan2014very} as a encoder for image inputs, and 3 stacked fully connected layers for audio inputs. We obtain the representations from image modality $h_i = f(x_i)$ and audio modality $$. \\

	\item\textbf{Project Head $p$.} Projection head $p$ is a shallow neural network, which is has several fully connected layers, projects the representations of different modalities from encoders to hidden space. The aim of project head is to enhance the performance of encoders, and align the represents with a same hidden space shape. The outputs of project head are $z_{n_1}=p(h_{n_1})$ for $h_{n_1}$ , $z_{n_2}=p(h_{n_2})$ for $h_{n_2}$, ..., and $z_{n_m}=p(h_{n_m})$ for $h_{n_m}$.
% need to describe the training process.

	\item\textbf{Contrastive Loss Function.} The contrastive loss function \cite{chen2020simple} is to make the encoders to learn the feature representations by themselves. The contrastive loss is to maximize the similarity between modalities $\{\widetilde x_{n_1},\widetilde x_{n_2}, ..., \widetilde x_{n_m}\}$, and minimize the samples ($\widetilde x_{n_1},\widetilde x_{n_2}, .., \widetilde x_{n_m}$) and other samples. 
\end{enumerate}

For the participants with only unimodal data, we train traditional Supervised Learning over unimodal data. We define this uni-modality is $n_1$.

\begin{enumerate}
	\item\textbf{Data Preprocessing.} We consider a sample is $x$. Then we apply data augmentation over the sample, denoted as $\widetilde x_{n_1}$. 
	
	\item\textbf{Feature Representations $M_{n_1}$.} Neural networks are applied for feature representations. A neural network is applied for uni-modality $n_1$. A neural network consists of several convolutional layers and fully connect layers. The neural network is to learn feature representations. The outputs of neural network are $r_{n_1}=M(\widetilde x_{n_1})$ for $\widetilde x_{n_1}$.
	
	% need to describe the training process.
	\item\textbf{Output Layer $f_f$.} We apply a fully connected layer $f_f$ as the final output layer. The predicted output is formed as $\widetilde y_{n_1} = f_f(r_{n_1})$.
	
	\item\textbf{Loss Function for Supervision.} Finally, a loss function for Supervised Learning is to measure the distance the prediction and its labels. 
\end{enumerate}

The \textit{new} things in our framework are that we are the first to use Contrastive Learning to extract transferable representations in Federated Transfer Learning, also the first to use Unsupervised Learning to help Supervised Learning in Federated Transfer Learning. In our Federated Transfer Learning framework, the participants with multimodal data can help the participants with unimodal data to improve their performance. Besides, this framework is general, and it also protects participants' privacy.
 
% how many possible combinations? Federated Learning must be used-solve data island problems.
% dose each possible combination is useful? why? and why choose it, why give the answer of something new in the thsis
% here need picture to describe combinations, and explain with related work.

\section{System Overview}
\label{sec:system_overview}
The whole system has two types of groups, i.e. Federated Learning for unimodal participants, Federated Learning for multimodal participants. The Transfer Learning method exists between these two types of groups. %Moreover, the users of our framework can set up different settings of a configuration file to extend the effectiveness. \\

\begin{figure}[H]
	\centering
	\includegraphics[width=0.85\textwidth]{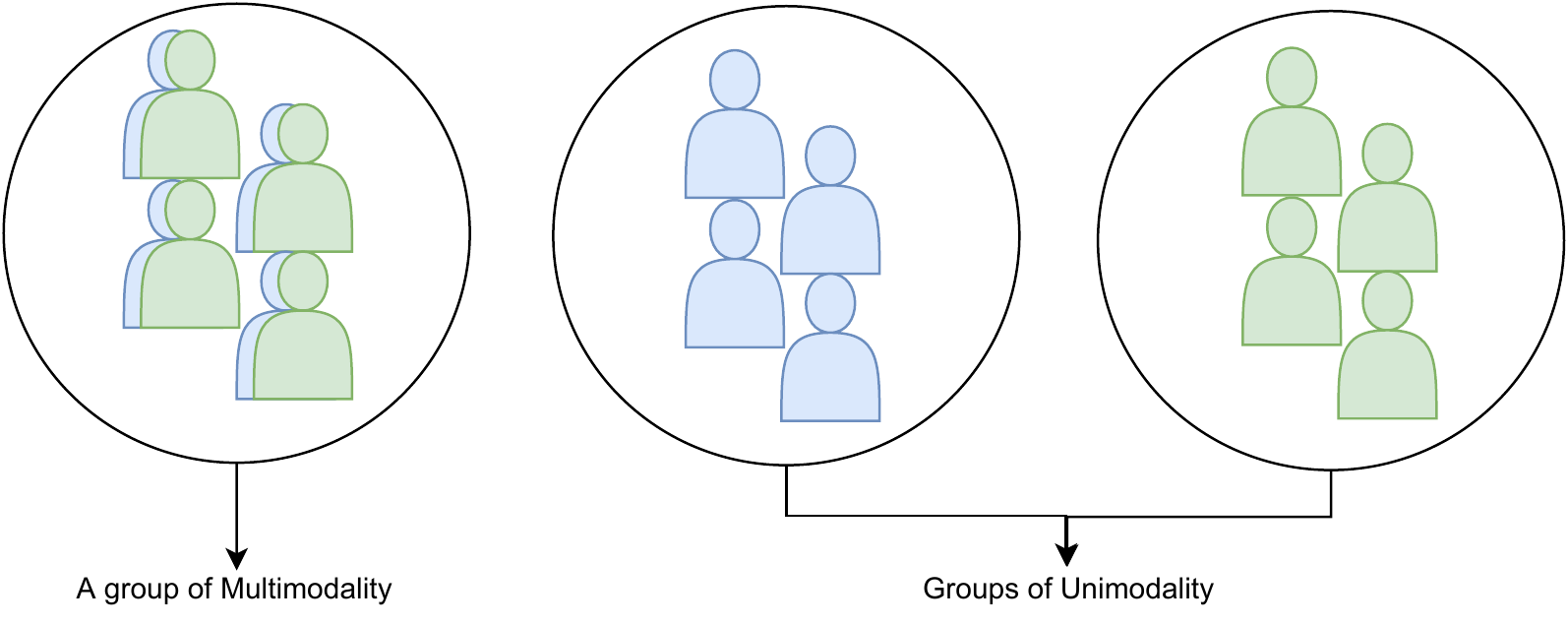}  	  	 	
	\caption{Federated Learning in groups for multi-modality (left) and uni-modality (right). The participants in both blue and green indicate that they hold multi-modality. In contrast, the participants in only green or blue indicate that they has only uni-modality. }
	\label{fig:figure_4_2}
\end{figure}

Figure \ref{fig:figure_4_2}) shows the group form in Federated Learning participants with unimodal or multimodal data. The participants in a round represent that they are in the same group. The participants only in blue or in green represent that they have only one modality, while the others with both colors represent that they have two modalities.

The whole framework consists of three components: Federated Learning for groups with only one modality, Federated Learning for a group with multi-modality and Federated Transfer Learning of two types of groups.
% In order to clearly explain the workflows, some notations, which are used in the framework, are as follows.

\subsection{Federated Learning with Unimodal Data}
% unimodal
The workflow of unimodal participants in the training process of Federated Learning is as follows.
\begin{itemize}
	\item \textbf{Step 1:} One modality data of the dataset is randomly divided, and each participant holds a portion of the data. The main server chooses a main model for unimodal data, and initializes the model. Then, the main server sends the initialized model to each participant.
	\item \textbf{Step 2:} In parallel, each participant receives the model from the main server, and trains locally. Then, each participant updates the current model based on its own loss using an optimization algorithm and sends the resulted model to the server.
	\item \textbf{Step 3:} The server receives all the models from participating nodes. Then, the server updates its global model as the average aggregation of these received models, and sends the aggregated model to each participating nodes. 
	\item \textbf{Step 4:} The iteration is repeated for many global epochs until maximum global number of epochs is reached or it has achieved the expected performance (e.g. accuracy).
\end{itemize}

\begin{figure}[htp]
	\centering
	\includegraphics[width=0.75\textwidth]{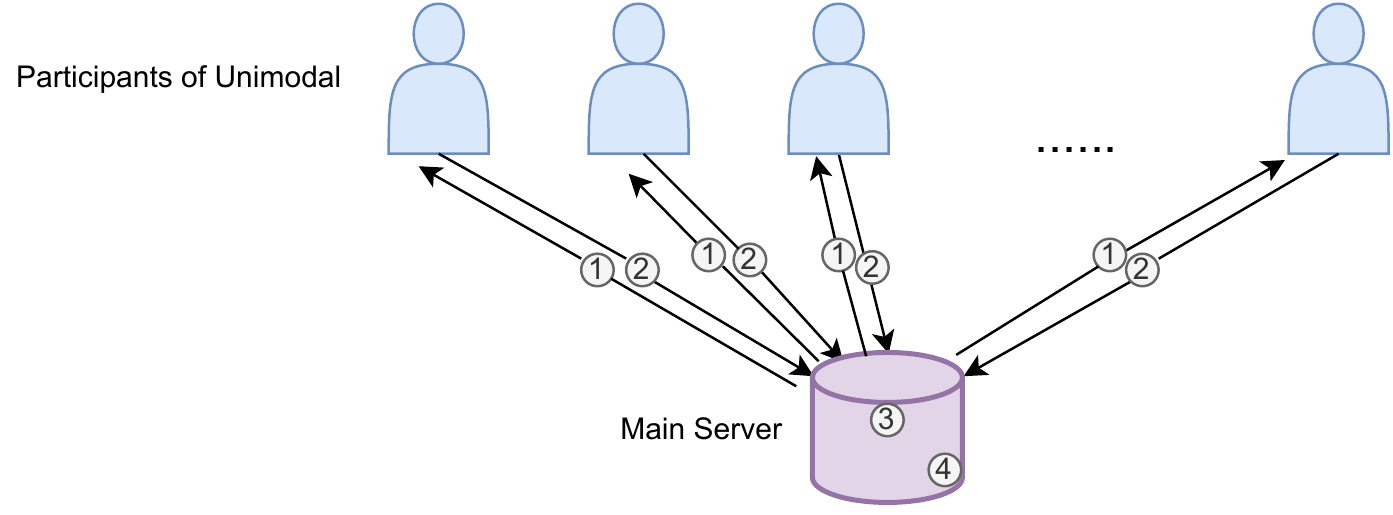}  	  	 	
	\caption{Federated Learning framework with unimodal data nodes. The participants only in blue represent that they have only one modality. Steps are marked on the diagram. The arrows between participants and the main server denote that the models are sent and received. }
	\label{fig:figure_4_3}
%	\label{fig:example1}
\end{figure}

Figure \ref{fig:figure_4_3} shows the workflow of participants with unimodal data. Each participant is blue, denoting the ownership of the same data modality. The arrows between participants and the main server denote that the models are sent and received.

\subsection{Federated Learning with Multimodal Data}
% multimodal
Participants in Federated Learning with multimodal data hold multi-modality. Nevertheless, the essential Federated Learning workflow remains the same.

\begin{itemize}
	\item \textbf{Step 1:} Two modalities (or more) of the dataset are randomly and divided into pairs or multiples, and each participant holds a portion of the data. The main server chooses a Contrastive Learning model, initializes the model, and sends the initialized model to each participant.
	\item \textbf{Step 2:} In parallel, each participant receives the model from the main server, and trains locally. Then, each participant updates the current model based on its own loss using an optimization algorithm, and sends the locally trained model to the server.
	\item \textbf{Step 3:} The server receives the models from participating nodes. Then, the server updates its global model as the average aggregation of these received models, and sends the aggregated model to each participating nodes. 
	\item \textbf{Step 4:} The iteration is repeated for many global epochs until maximum global number of epochs is reached or it has achieved the expected performance (e.g. accuracy).
\end{itemize}

\begin{figure}[htp]
	\centering
	\includegraphics[width=0.85\textwidth]{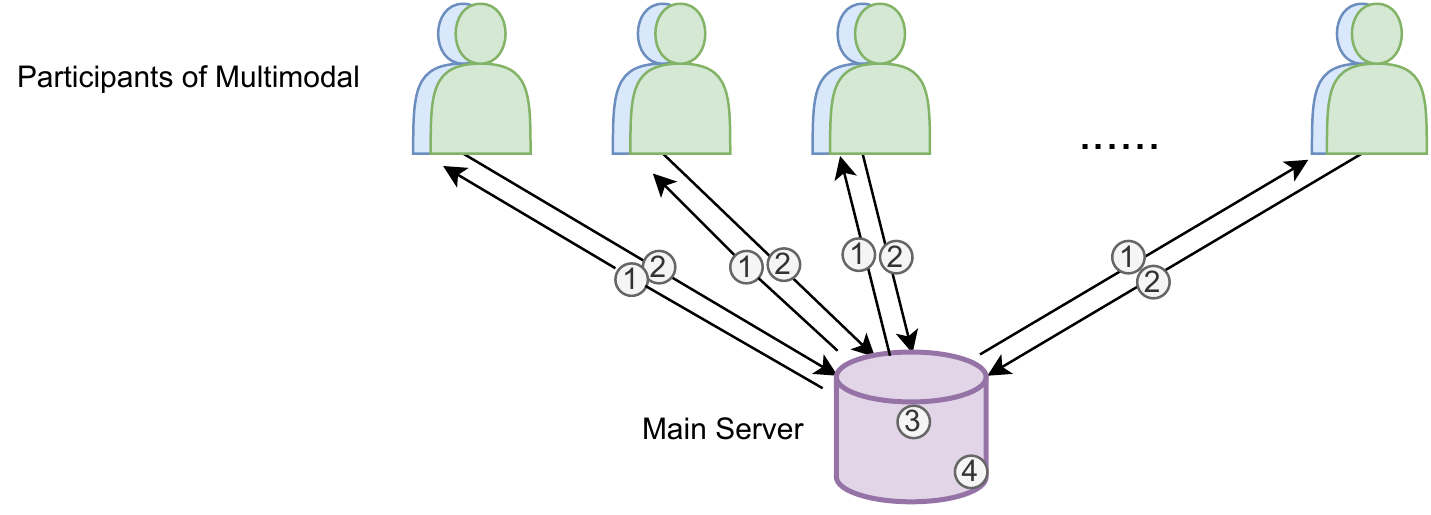}  	  	 	
	\caption{Federated Learning framework with multimodal data nodes. The participants in both blue and green represent that they have multiple modalities. Steps are marked on the diagram. The arrows between participants and the main server denote that the models are sent and received.}
	\label{fig:figure_4_4}
\end{figure}

Figure \ref{fig:figure_4_4} shows the workflow of participants with multimodal data. 

\subsection{Federated Transfer Learning with Unimodal and Multimodal Data}
% FTL
Participants in the Federated Transfer Learning framework with multimodal data either hold one modality or multi-modality. It works as follows.
\begin{itemize}
	\item \textbf{Step 1:} When the participants have the same modality, and they are split into the same group. 
	\item \textbf{Step 2:} Participants in the same group perform Federated Learning. Inside each group, each participant holds a portion of the data. First, the main server chooses related models for groups, and a pretext for a group with image and audio), and initializes the models. Then, the main server sends the initialized model to each participant.
	\item \textbf{Step 3:} In parallel, each participant of each group receives the model from the main server, and trains locally. Then, each participant updates the current model based on its own loss using an optimization algorithm like and sends the locally trained model to the server.
	\item \textbf{Step 4:} The server receives the models from participating nodes. Then, the sever aggregate twice. The first aggregation is to average the models within each group. The second aggregation is to average the models between different modalities, if they have the same part of models (the same sub-models). After aggregation, the server sends the aggregated same part of models to each related participating nodes. 
	\item \textbf{Step 5:} The iteration is repeated for many global epochs until the last global epoch or it has achieved the expected performance (e.g. accuracy). Then, the final model which is the expected pretext model, which is usually saved as a file.
\end{itemize}

\begin{figure}[htp]
	\centering
	\includegraphics[width=1.0\textwidth]{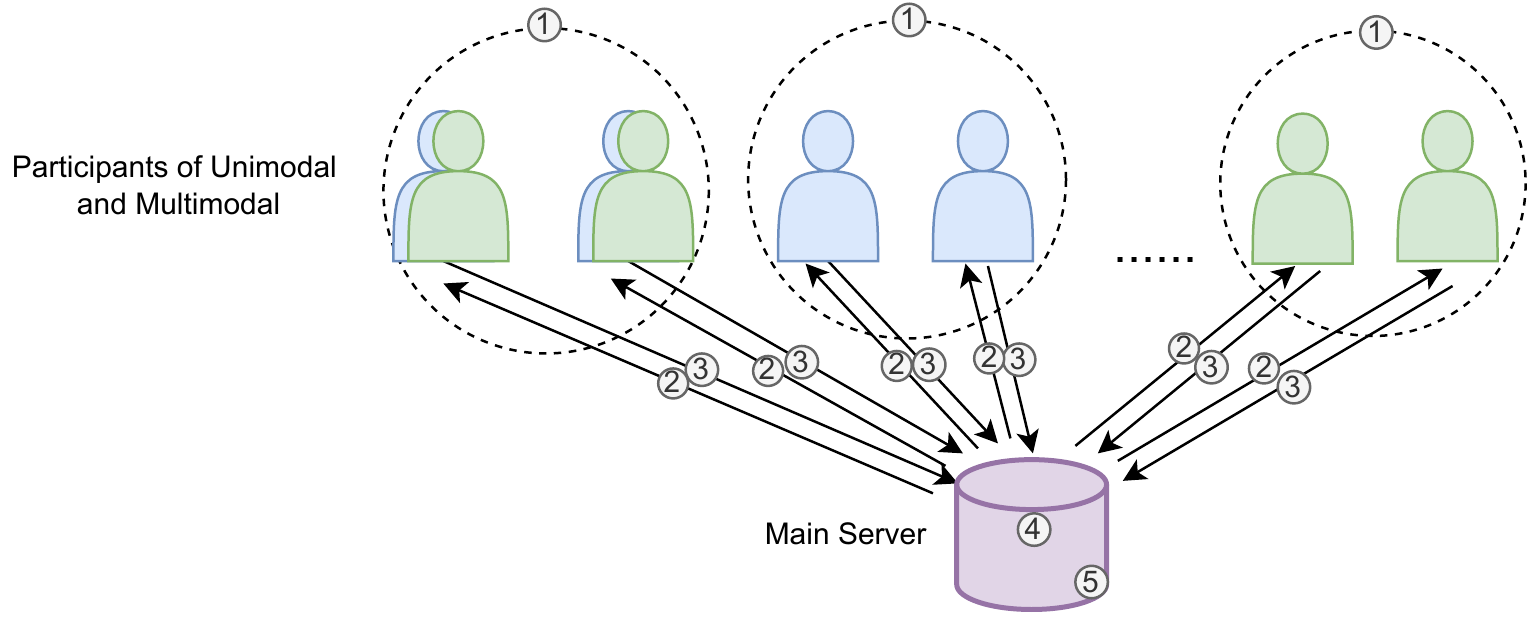}  	  	 	
	\caption{Federated Transfer Learning framework with unimodal and multimodal data. Steps are marked on the diagram. The dashed round represents that they are in the same modality type of group. The arrows between participants and the main server denote that the models are sent and received. }
	\label{fig:figure_4_5}
\end{figure}

Figure \ref{fig:figure_4_5} shows participants' workflow with multimodal data. Different colors represent different modalities. The participants are only in blue or green, indicating that they hold one data modality. In contrast, the participants are both in green and blue colors, indicating that they hold the two(or more) data modalities.

Figure \ref{fig:figure_4_6} below shows the centralized model of Federated Learning with unimodal and multimodal inputs. We fuse the last layer from sub-networks and the last layer from MLPs. More concretely, we concatenate the final neurons from two sub-networks. 
%For $N_1$, the final neurons from VGG16+MLP are 41, and the DMLP has also 41 neurons, the concatenated neurons are 82.
The final concatenated layer, which is then used to calculate the similarity \cite{jaiswal2020survey}.
%The similarity in multi-modality can be seen as transferable knowledge The transferable knowledge is then used to uni-modalities by aggregation. %The Transfer Learning can be seen as a fine-tuning process from pretext tasks to downstream tasks. After being trained for pretext tasks, the fine-tuned linear classifier can be stored, then the fine-tuned linear classifier is used for prediction.

\begin{figure}[htbp]
	\centering
	\includegraphics[width=0.85\textwidth]{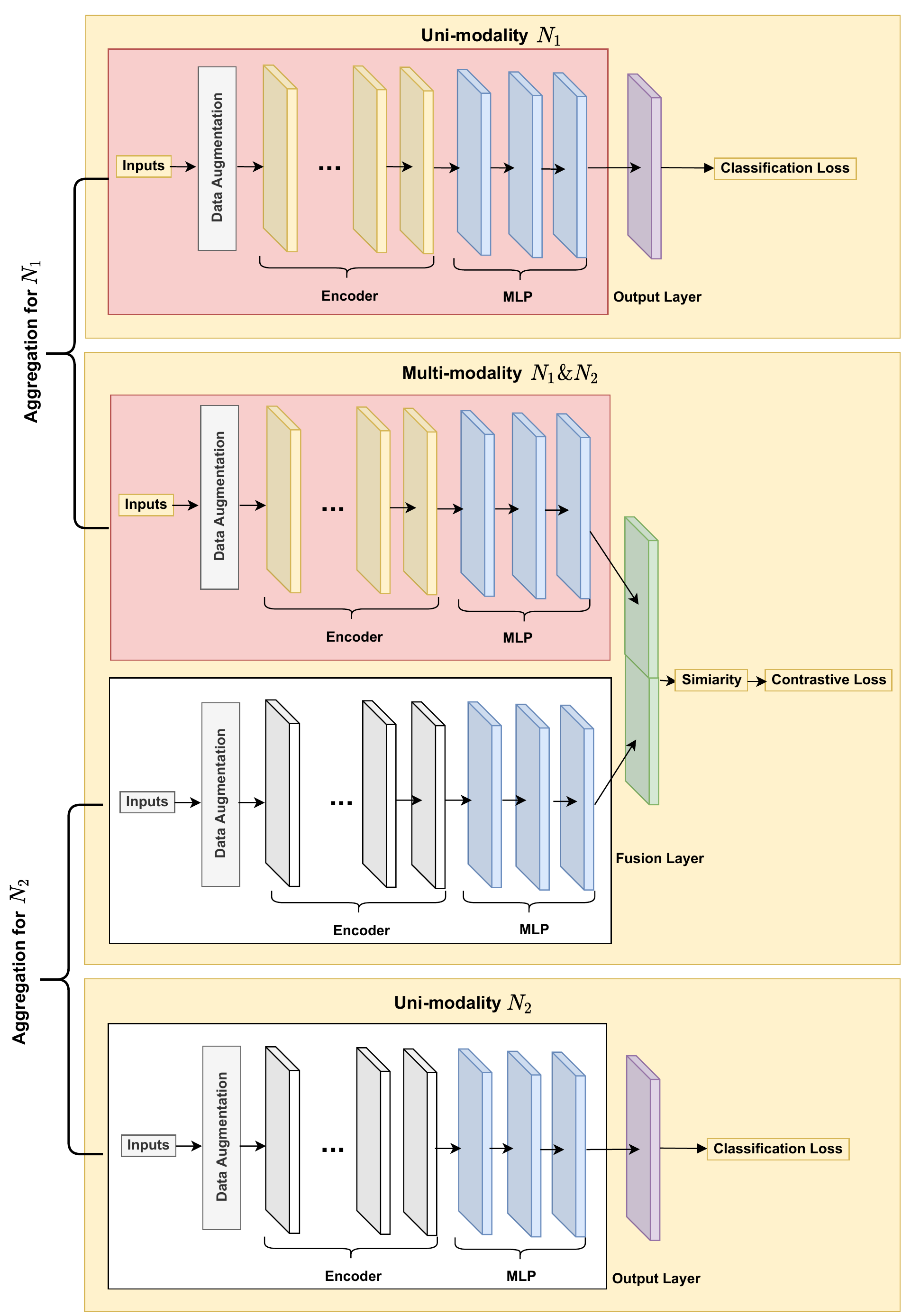}  	  	 	
	\caption{Federated Transfer Learning for multi-modality and uni-modality. $N_1$ and $N_2$ are examples for two types of modalities. Red rectangle and white rectangle represent sub-networks for $N_1$ and $N_2$. The aggregation is applied to grouped federation. This aggregation is to average the same part of sub-networks between uni-modality and multi-modality. The grouped federation (aggregation) exists in the same type modalities in a group. The fusion layer represents concatenation of two sub-networks.}
	% VGG16+MLP is used for image classification. MFCC+DMLP is used for audio classification. In multi-modality, VGG16+MLP and MFCC+DMLP work as sub-networks and perform Contrastive Learning.
	\label{fig:figure_4_6}
\end{figure} %FTL_workflow-group

%缺一张图表述
\subsection{Baseline Models}
There are two baseline models: Federated Learning for unimodal data and Federated Learning for multimodal data. Federated Learning for unimodal data is similar to the first component of our framework. However, Federated Learning for multimodal data is different from our framework. Main difference is in the centralized model. Our framework uses Contrastive Learning, while the baseline model use only fate fusion without similarity calculation, which is directly concatenate the sub-models from multi-modality. \\

\begin{figure}[H]
	\centering
	\includegraphics[width=0.85\textwidth]{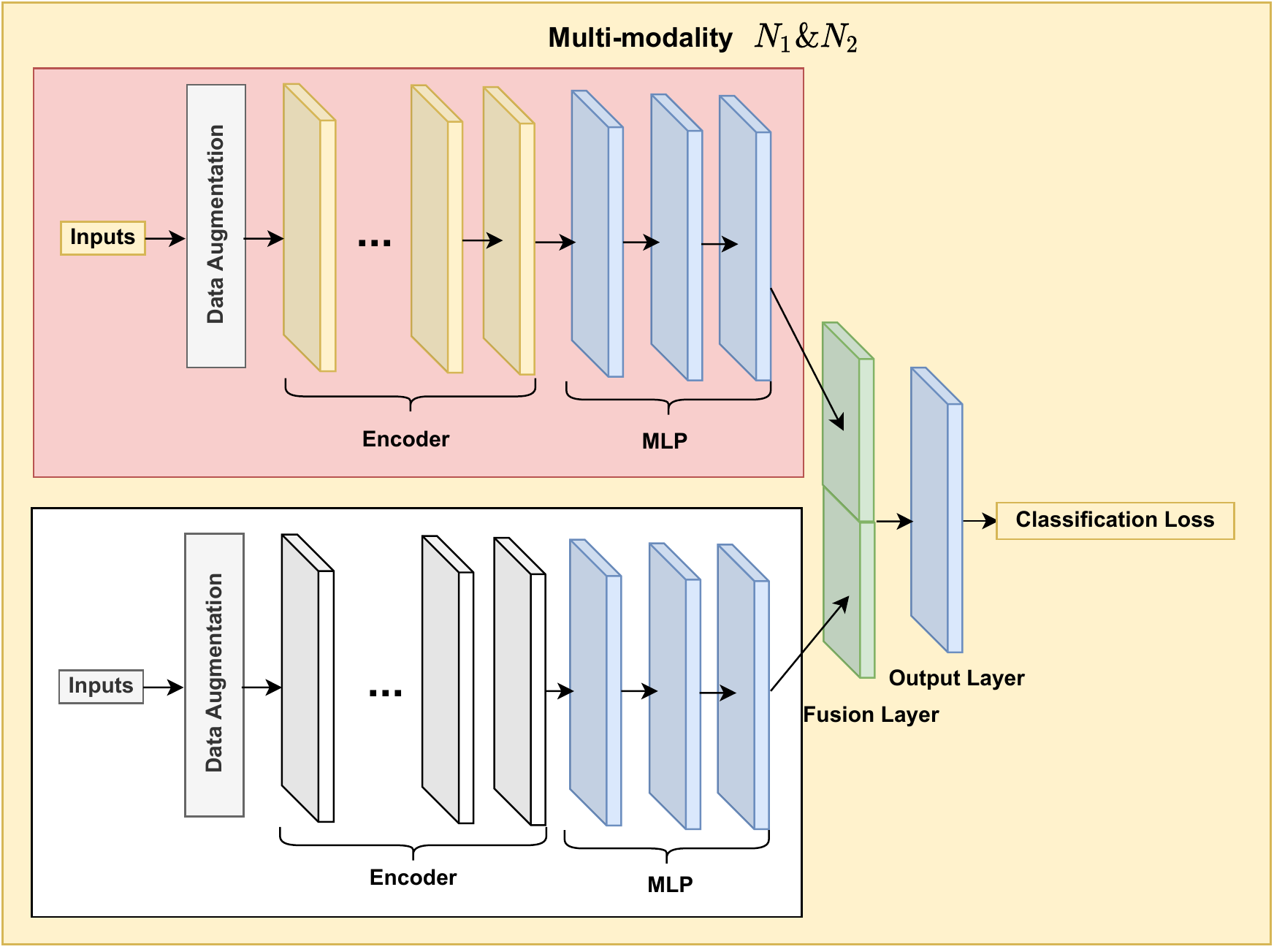}  	  	 	
	\caption{Late fusion. $N_1$ and $N_2$ are examples for two types of modalities. Red rectangle and white rectangle represent sub-networks for $N_1$ and $N_2$. The fusion layer represents concatenation of two sub-networks.} % need to clear explain
	\label{fig:figure_4_7}
\end{figure}

Figure \ref{fig:figure_4_7} shows the late fusion of two modalities.
% Red rectangle is used for $N_1$, and white rectangle is used for $N_2$.

\section{Summary}
The design is based on the combination of Federated Learning and Transfer Learning, while considering the different distributions of multimodal data, some participants may have only unimodal data, and others have multimodal data. In order to verify the entire design, it is necessary to ensure that the baseline models are reproducible and that there is no significant deviation from the corresponding testing accuracy. We also need to apply Federated Learning in baseline models to verify the basic effectiveness. Furthermore, considering the non-IID setting may affect the performance of Federated Transfer Learning, experiments are required. Finally, it is crucial to improve performance while protecting privacy. 

%*****************************************
\chapter{Implementation}
\label{ch:implementation}
%*****************************************
%\hint{This chapter should describe the details of the implementation addressing the following questions: \\ \\
%1. What are the design decisions made? \\
%2. What is the environment the approach is developed in? \\
%3. How are components mapped to classes of the source code? \\
%4. How do the components interact with each other?  \\
%5. What are limitations of the implementation?}
This chapter is to describe the details of implementation: (1) We will give details about the selected dataset and preprocessing of the dataset. (2) The workflow of design is applied into the selected dataset, to verify the efficiency of the design. (3) The algorithms and interactions of Federated Transfer Learning in our framework are introduced.

\section{Design Decisions}
In order to show that our framework is efficient, a multimodal dataset, the corresponding unimodal and multimodal baseline models must be selected. The multimodal dataset as well as baselines models should satisfy following the requirements: (1) each sample of the multimodal dataset has two or more overlapping features, and the modalities do not affect each other; (2) both unimodal and multimodal models must be used for the same classification tasks; (3) multimodal data can be co-trained with two or more unimodal data.

In addition, the data distribution of participants is also an issue that needs to be explored. For example, the data distribution is usually IID. However, in the real world, it is non-IID, e.g. participants hold different amounts of data or only hold data for one category. Thus, it needs to explore how the non-IID setting affects Federated Transfer Learning framework performance.

%It is necessary that a selected dataset satisfies all the requirements. The selected dataset has two modalities, i.e. image and audio modalities. Each sample of the dataset is pairwise connected, which means the two modalities have more or less overlapping features. The modality of 2d images provide visual cues and the 1d audio give us acoustic features. 
%\subsection{Dataset and Preprocessing}
\begin{figure}[H]
	\centering
	\includegraphics[width=1.0\textwidth]{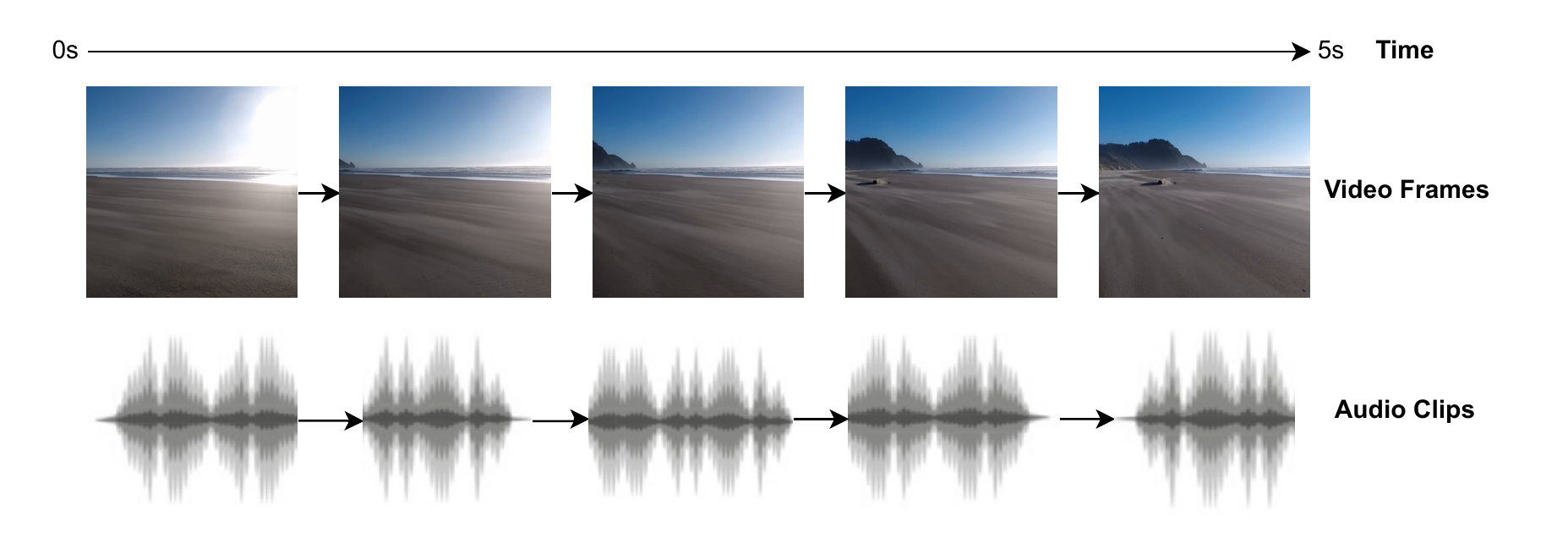}  	  	 	
	\caption{Example of images and audio for scene classification dataset \cite{bird2020look}.}
	\label{fig:figure_5_1}
\end{figure}
% make a table to describe the whole dataset
\textbf{Dataset.}
The dataset consists of videos from 9 different environments. In total, 45 video sources have been processed, and they are at 29.97 frames per second and then reduced to 2000 seconds. Each frame (at 0.5s, 1.5s, 2.5s ... and so on) of a video is extracted as a image, at the same time, each MFCC \cite{muda2010voice} audio statistic is also extracted from corresponding second of video. These 9 environmental classes are in the table \ref{tab:table_5_1}. All the videos are captured in the nature environment. In order to make better distinction between different classes and obtain better recognition results, it is necessary to crop the image at each second to obtain the initial data objects. As it is shown in Figure \ref{fig:figure_5_1}, each frame of video and each piece of audio are extracted in pairs. Thus, 32,000 recognizable objects and 17,252 RGB images pairwise with 17,252 seconds of audio \cite{bird2020look}. 

\begin{table}[H]
	\centering
	\begin{tabular}{|l|l|}
		\hline 
%		\hline 
		\textbf{Labels for Scene Classification} & \textbf{Description} \\
%		\hline 
		\hline 
		Beach & 2080 seconds, from 4 sources \\		
		\hline
		City & 2432 seconds, from 5 sources  \\
		\hline
		Classroom & 2753 seconds, from 6 sources \\
		\hline
		Forest & 2000 seconds, from 3 sources  \\
		\hline		
		Football Match & 2300 seconds, from 4 sources \\
		\hline
		Grocery Store & 2079 seconds, from 4 sources \\
		\hline
		Jungle & 2000 seconds, from 3 sources \\
		\hline
		Restaurant & 2300 seconds, from 8 sources \\
		\hline
		River & 2500 seconds, from 8 sources \\
		\hline 
	\end{tabular}	
	\caption{Description of labels for scene classification \cite{bird2020look}.}
	\label{tab:table_5_1}
\end{table}

The dataset is split into training, validation, and test dataset in a ratio of 8:1:1. Training dataset has 13800 samples, validation dataset has 1726 samples and testing dataset has 1726 samples. Besides, each dataset keeps the same distribution.\\

%\begin{figure}[H]
%	\centering
%	\includegraphics[width=1.0\textwidth]{images/dataset_distribution.png}  	  	 	
%	\caption{Distribution of the dataset with 9 categories. The horizontal axis are class names, and the vertical axis is number of samples.} 	  	 	
%	%	\caption{Example of Images and Audios for Scene Classification Dataset\cite{bird2020look}}
%	%	\label{fig:example1}
%\end{figure}
%dataset_distribution
\begin{figure}[H]
	\centering
	\begin{minipage}[H]{0.49\textwidth}
		\includegraphics[width=1.0\textwidth]{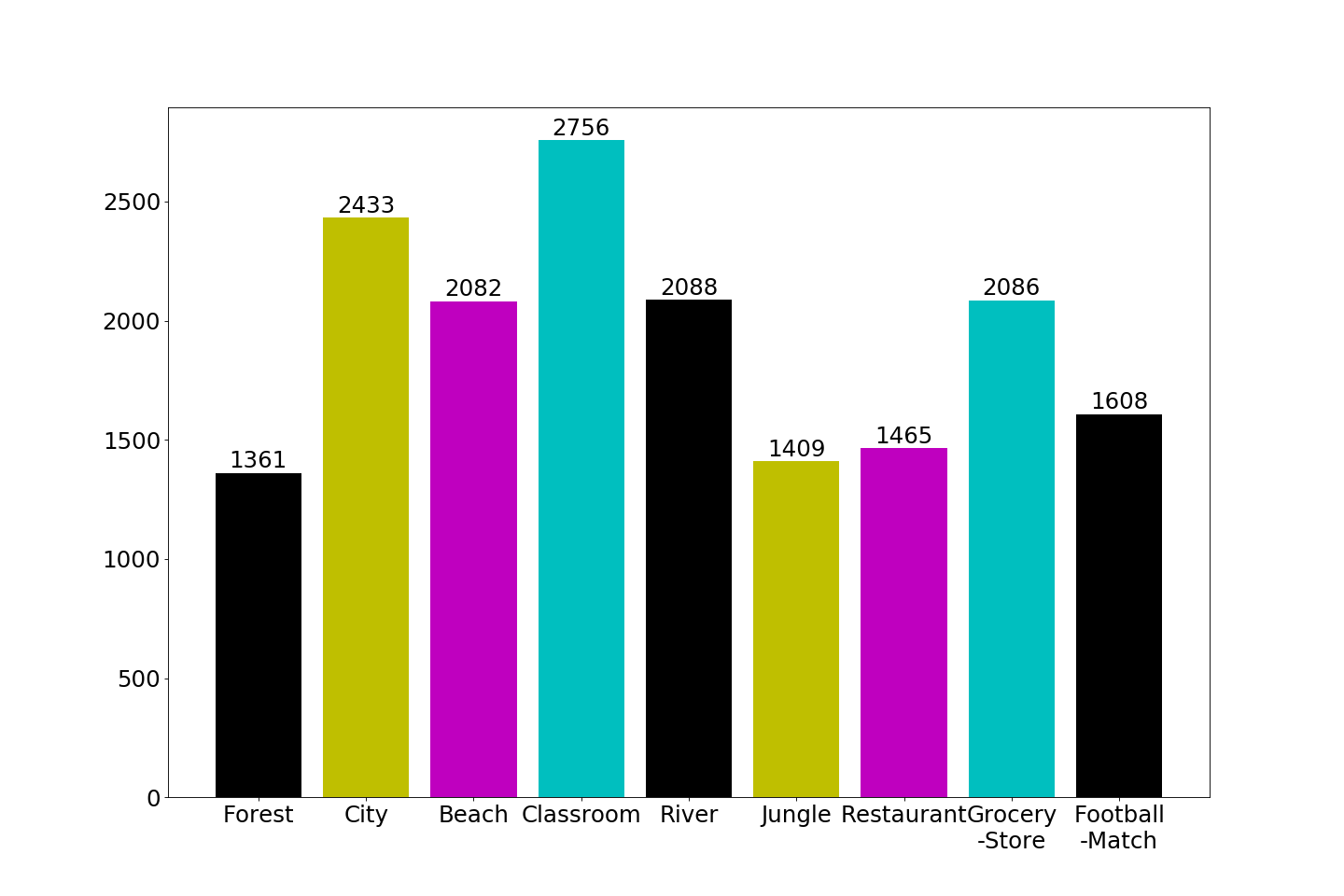}  
		\subcaption{Distribution of the whole dataset.}	  	 
	\end{minipage}
	\begin{minipage}[H]{0.49\textwidth}
		\includegraphics[width=1.0\textwidth]{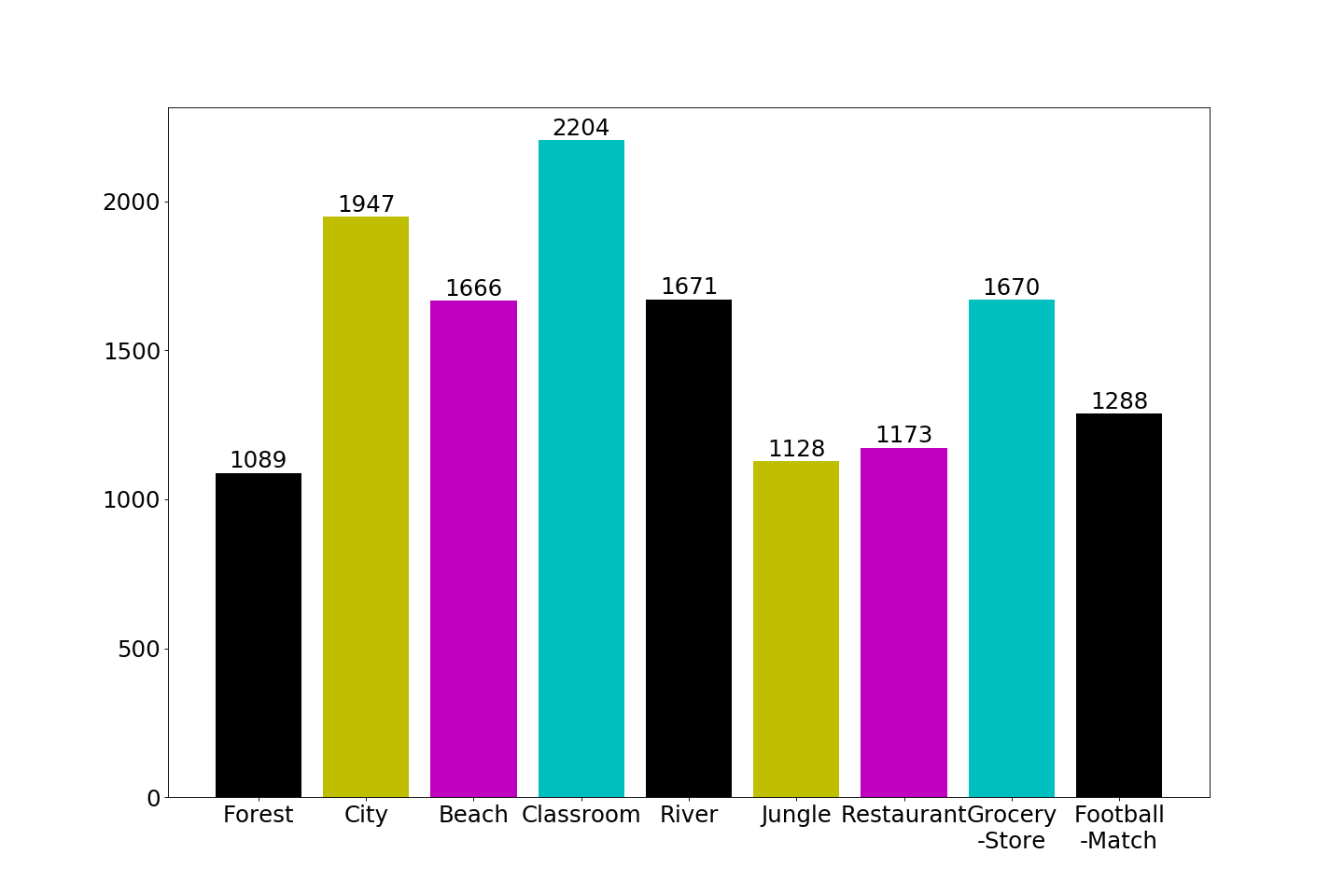}
		\subcaption{Distribution of training dataset.}	    	
	\end{minipage}
	\begin{minipage}[H]{0.49\textwidth}
		\includegraphics[width=1.0\textwidth]{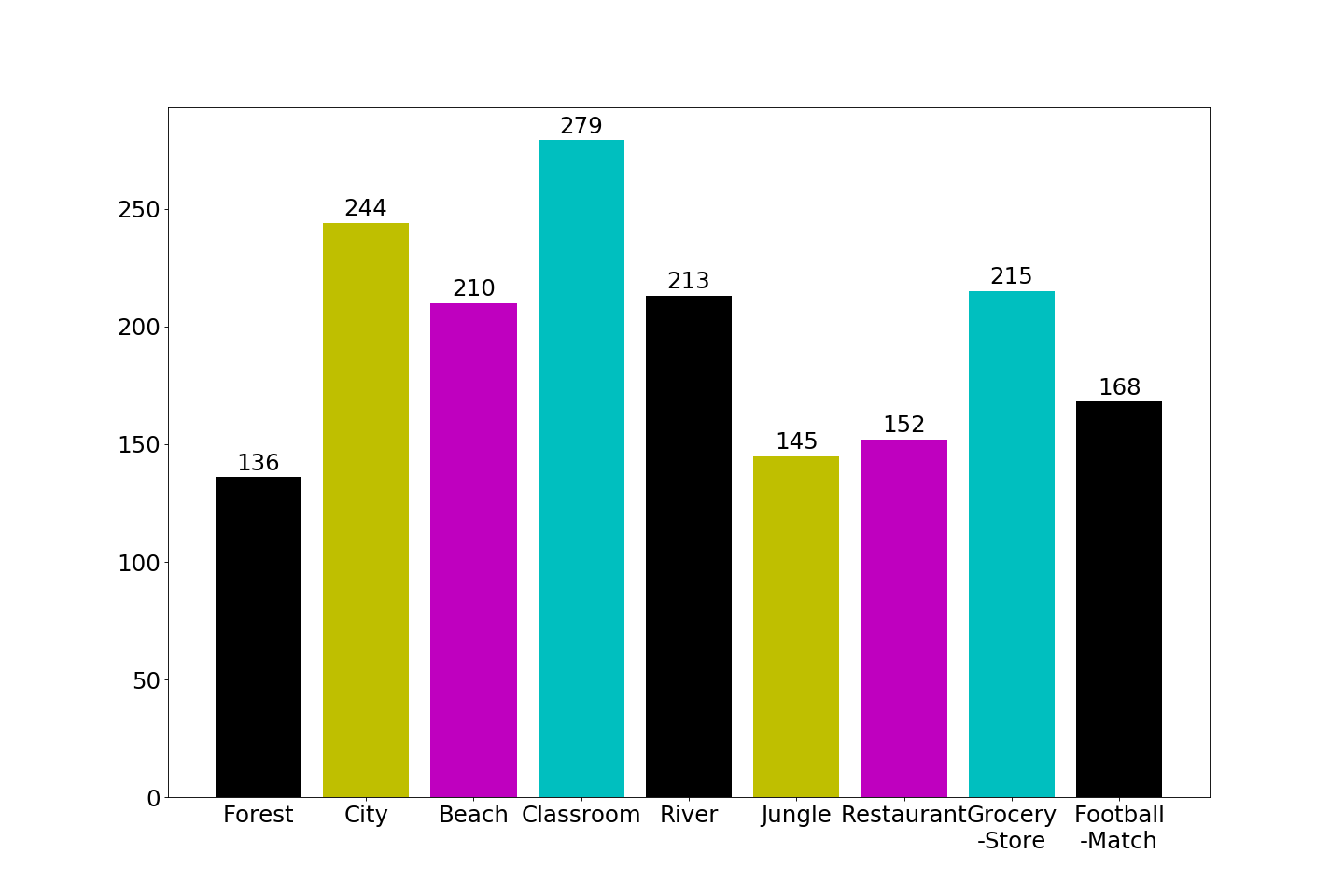}
		\subcaption{Distribution of validation dataset.}  	  	 
	\end{minipage}
	\begin{minipage}[H]{0.49\textwidth}
		\includegraphics[width=1.0\textwidth]{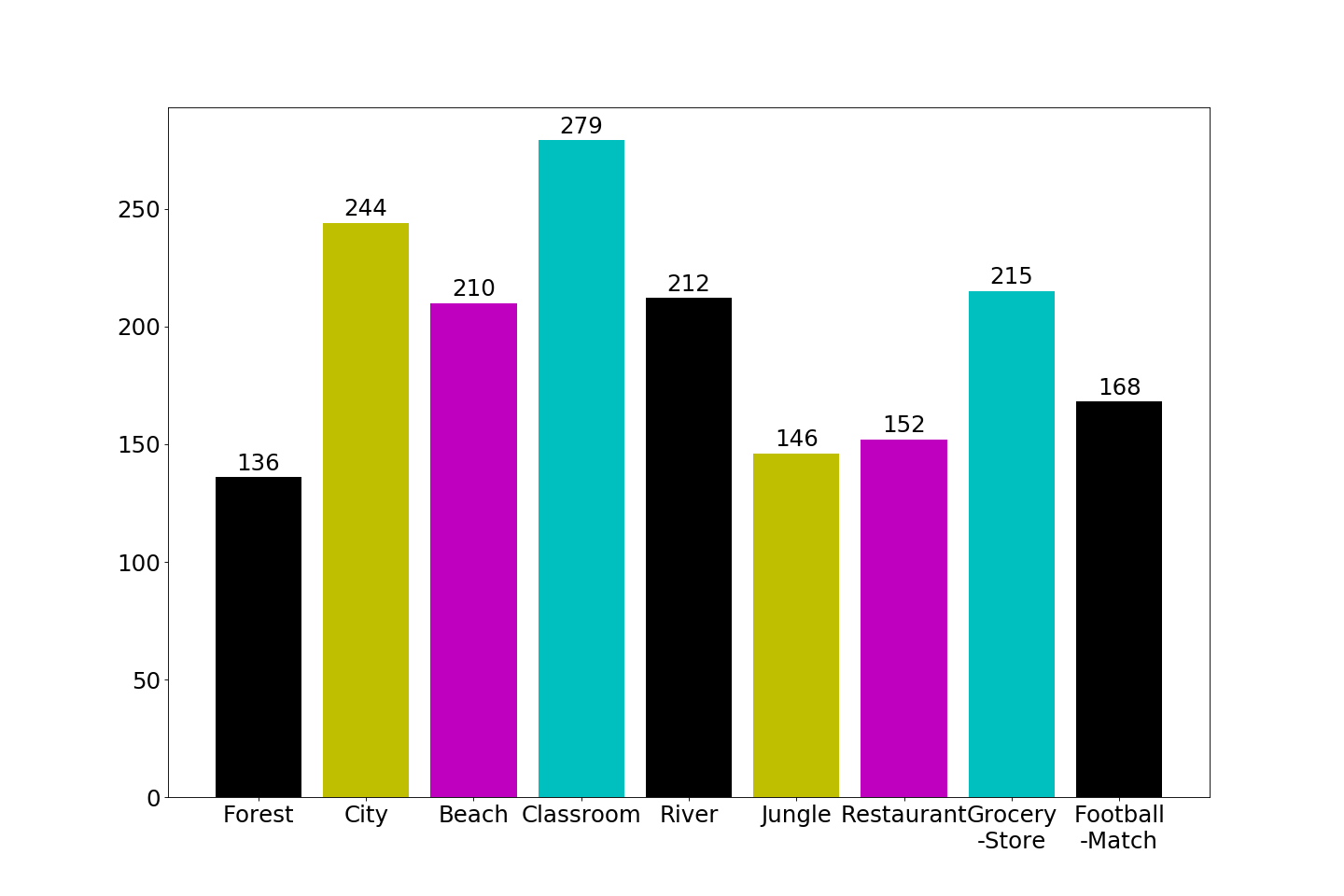}  
		\subcaption{Distribution of testing dataset.} 	
	\end{minipage}
	\caption{\textbf{Audio modality.} Histogram of the whole dataset (top-left), histogram of the training data (top-right), histogram of the validation data (bottom-left), histogram of testing data (bottom-right)}
	\label{fig:figure_5_2}
\end{figure}

%need some description of MFCC
\textbf{Preprocessing of Dataset.} A video is sample of the dataset. Each video has two modalities image and audio modalities. For image modality, each second frame of a video is cropped as a 128x128x3 RGB image. For audio modality, each audio should be extracted as MFCC features.

\begin{figure}[H]
	\centering
	\includegraphics[width=0.9\textwidth]{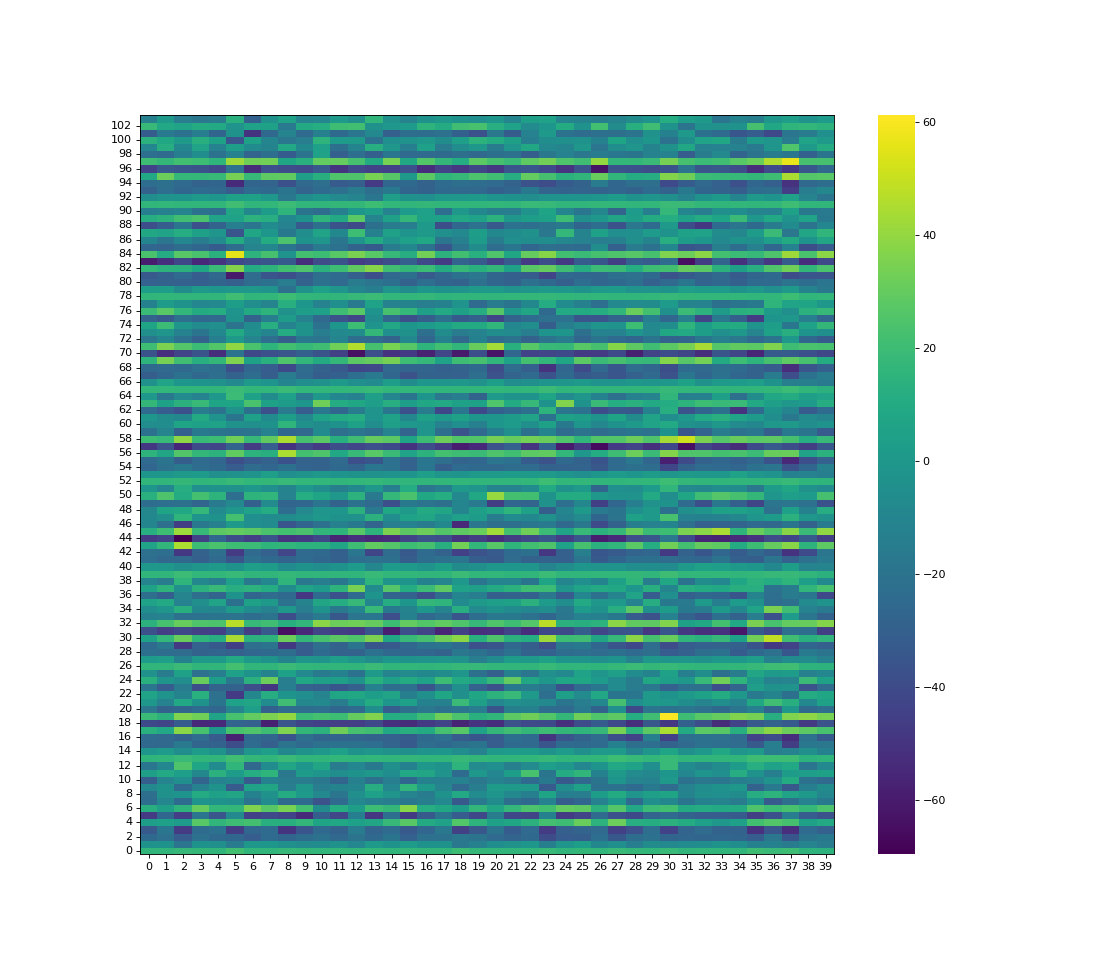}  	  	 	
	\caption{MFCC feature example of forest from 0s-39s. X-axis represents time step, and y-axis represents vectors of extracted MFCC.} 	
	\label{fig:figure_5_3}  	 	
\end{figure}

%\subsection{Central models}

\section{Architecture}

We need to implement two parts: baseline models and the designed framework. \\

\begin{table}[H]
	\centering
	\begin{tabular}{|l|l|}
		\hline 
		%		\hline 
		\textbf{Names} & \textbf{Explanations} \\
		%		\hline 
		\hline 
		VGG16 & Visual Geometry Group \cite{simonyan2014very},a deep neural network model for image classification \\
		\hline
		MLP &  \makecell[l]{Multilayer Perceptron \cite{hastie2009elements}, stacked several layers based on fully connected layers,\\ can be used after VGG16.} \\
		\hline
		DMLP & Deep Multilayer Perceptron, a deep neural network model for audio classification. \\
		\hline 
	\end{tabular}
	\caption{Explanation of models to be used in different modalities in baseline models \cite{bird2020look}.}
	\label{tab:table_5_2}
\end{table}

\textbf{Baseline models.} There are three baseline models: Federated Learning for image, Federated Learning for audio, Federated Learning for image-audio. The centralized models of baseline models: VGG16+MLP for image, MFCC+DMLP for audio, and late fused VGG16+MLP and MFCC+DMLP for image-audio.\\

\textbf{Designed Framework.} The centralized models of designed framework: VGG16+MLP for image, MFCC+DMLP for audio, and Contrastive Learning based model for image-audio. \\

Following figures give us details about all the centralized models, including centralized baseline models and centralized models for new designed framework.\\

\begin{figure}[htb]
	\centering
	\includegraphics[width=1.0\textwidth]{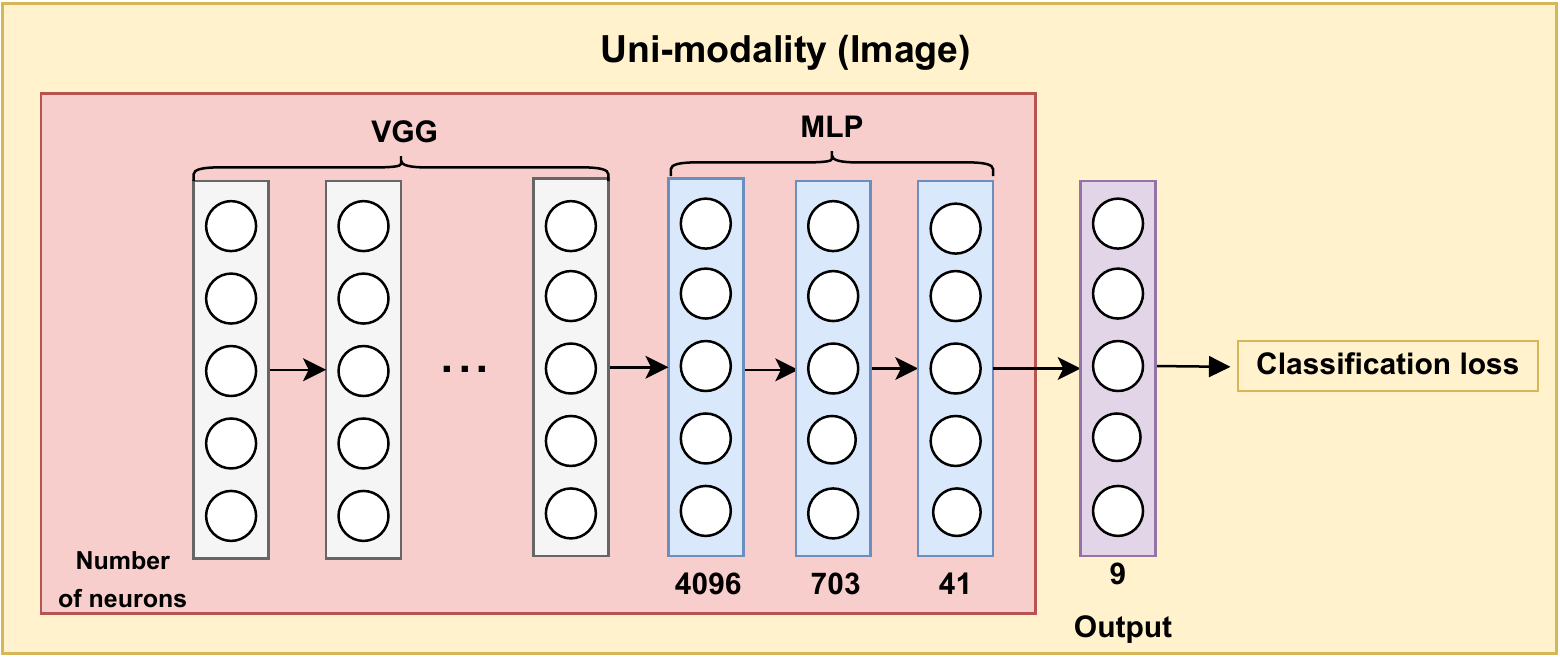}  	  	 	
	\caption{Neural network structure of VGG16+MLP. The red rectangle is the sub-network for image modality. VGG is the encoder. MLP works as projector. The final output layer is calculate the final supervised prediction. This process is a detailed description of $N_1$ modality in Figure \ref{fig:figure_4_6}.} 	  	 	
	%	\caption{Example of Images and Audios for Scene Classification Dataset\cite{bird2020look}}
	\label{fig:figure_5_4}
\end{figure}

Figure \ref{fig:figure_5_4} gives us details about VGG16+MLP. VGG16+MLP keeps the original VGG16 model, the input neurons of last layer is 4096, changes the last output neurons as 703, and follows 41 neurons. In MLP, we use Leaky ReLU as activation.\\

\begin{figure}[htb]
	\centering
	\includegraphics[width=1.0\textwidth]{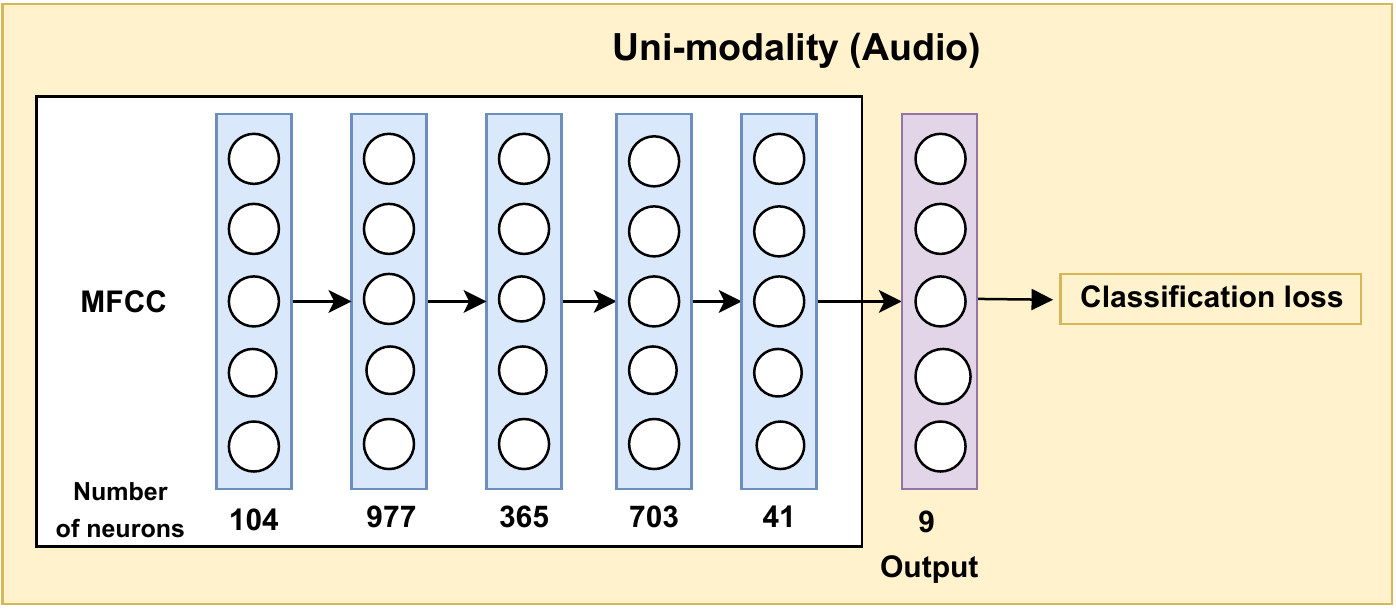}  	  	 	
	\caption{Neural network structure of DMLP. The white rectangle is the sub-network for audio modality. The first three layer is the encoder. The last two layers work as projector. The final output layer is calculate the final supervised prediction. This process is a detailed description of $N_2$ modality in Figure \ref{fig:figure_4_6}.} 	  	 	
	%	\caption{Example of Images and Audios for Scene Classification Dataset\cite{bird2020look}}
	\label{fig:figure_5_5}
\end{figure}

Figure \ref{fig:figure_5_5} gives us details about DMLP. MFCC features are inputs. DMLP with hidden layers is ordered by 104, 977, 365, 703 and 41 neurons. Both VGG16+MLP and DMLP are introduced in \cite{bird2020look}. We add the relu activation function between the hidden layers of DMLP. \\

\textbf{Implementation of Baseline Models of Uni-modality.} Following gives us details about baseline model of single modality, image or audio. 

\begin{enumerate}
\item\textbf{Data Preprocessing.} In our framework, we consider a video sample is $x$, a RGB image frame and a audio sequence from this video are two different modalities (\textit{views}): a RGB image frame is transformed by flipping, denoted as $x_i$, a audio sequence is extract by MFCC, denoted as $x_a$. 

\item\textbf{Feature Representations $M_i$ and $M_a$.} Deep neural networks are applied for feature representations. one sub-network VGG+MLP ($M_i$) is applied for visual feature representation, and the other sub-network DMLP ($M_a$) is applied for auditory features. Here, the aim of feature representations are to enhance the performance of encoders, and align the represents to do late fusion. The outputs of feature representations are $r_{i}=M(x_{i})$ for $x_{i}$ and $r_{a}=M(x_{a})$ for $x_{a}$.

% need to describe the training process.
\item\textbf{Output Layer $f_f$.} We apply a fully connected layer $f_f$ as the final output layer after concatenation. The predicted output is formed as $\widetilde y_{i} = f_f(r_{i})$ for image, and $\widetilde y_{a} = f_f(r_{a})$ for audio.

\item\textbf{Entropy Loss Function.} The entropy loss function \cite{bishop2006pattern} is make the late fusion two sub-networks perform Supervised Learning. Here, the loss function is to measure the distance the prediction and its labels. 

The output of $\widetilde y_{LF}$ is 9, because the dataset has 9 classes. We consider a dataset with augmented samples $\widetilde X$ and labels $y\in Y$. Each sample $x$ has a positive pair $\widetilde x_{i}$ and $\widetilde x_{a}$. The entropy loss function is as follows.

\begin{align}
\operatorname*{arg min}_{M_i} \frac{1}{|\widetilde X|} \sum \limits_{(\widetilde x_i,y)\in \widetilde X} L(M(x_i), \widetilde{y}),
\end{align}
\begin{align}
\operatorname*{arg min}_{M_a} \frac{1}{|\widetilde X|} \sum \limits_{(\widetilde x_a,y)\in \widetilde X} L(M(x_a), \widetilde{y}),
\end{align}
\begin{align}
L_{CE}(y, \widetilde  y_{i}) = \sum\limits_{k=1}^9 {y^k} \text{\xspace log \xspace} \widetilde y_{i}^k,
\end{align}
\begin{align}
L_{CE}(y, \widetilde  y_{a}) = \sum\limits_{k=1}^9 {y^k} \text{\xspace log \xspace} \widetilde y_{a}^k,
\end{align}
where ${y^k}$ equals to 1 if the sample belongs to class $k$, otherwise 0, and $y_{i}^k$ or $y_{a}^k$ are the $k$-th value of in $y_{i}$ or $y_{a}$.\\ 
\end{enumerate}

\textbf{Implementation of Baseline Model of Multi-modality.} Following gives us details about Contrastive Learning model for image-audio, which is similar in SimCLR \cite{chen2020simple}.

\begin{figure}[H]
	\centering
	\includegraphics[width=1.0\textwidth]{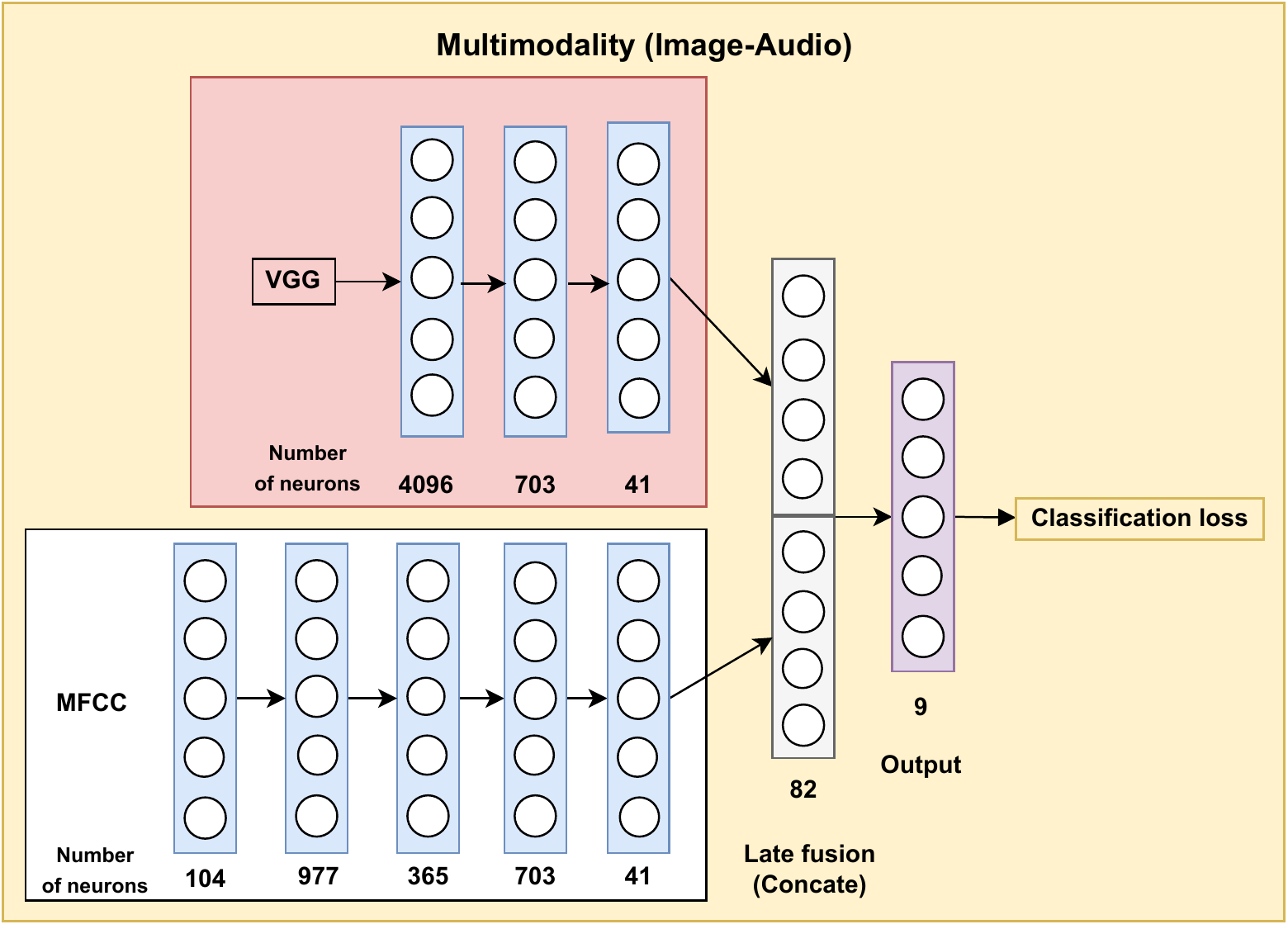}  	  	 	
	\caption{Neural network structure of late fusion. This process is a detailed description of the Figure \ref{fig:figure_4_7}.}
	\label{fig:figure_5_6}
\end{figure}

Figure \ref{fig:figure_5_6} gives us details about late fusion model for image-audio. The late fused model only concatenate two sub-networks together. Then, the concatenated outputs are fed into the last fully connect layer to get final outputs. Finally, the final outputs are used for classification.

\begin{enumerate}
	%\centering
	\item\textbf{Data Preprocessing.} In Our framework, we consider a video sample is $x$, a RGB image frame and a audio sequence from this video are two different modalities (\textit{views}): a RGB image frame is transformed by flipping, denoted as $x_i$, a audio sequence is extract by MFCC, denoted as $x_a$.
	
	\item\textbf{Feature Representations $M_i$ and $M_a$.} Deep neural networks are applied for feature representations. one sub-network VGG+MLP ($M_i$) is applied for visual feature representation, and the other sub-network DMLP ($M_a$) is applied for auditory features. Here, the aim of feature representations are to enhance the performance of encoders, and align the represents to do late fusion. The outputs of project head are $r_{i}=M(x_{i})$ for $x_{i}$ and $r_{a}=M(x_{a})$ for $x_{a}$.
	
	% need to describe the training process.
	\item\textbf{Late Fusion $c$ and Output Layer $f_f$.} We concatenate the outputs from $M_i$ and $M_a$. The late fusion result is $c_{LF}=c(r_i, r_a)$. Then, we apply a fully connected layer $f_f$ as the final output layer after concatenation. The predicted output is formed as $\widetilde y_{LF} = f_f(c_{LF})$
	
	\item\textbf{Entropy Loss Function.} The entropy loss function is make the late fusion two sub-networks perform Supervised Learning. Here, the loss function is to measure the distance the prediction and its labels. The output of $\widetilde y_{LF}$ is 9, because the dataset has 9 classes. We consider a dataset with augmented samples $\widetilde X$ and labels/classes $y\in Y$, the entropy loss function is as follows.
	\begin{align}
	\operatorname*{arg min}_{M_i, M_a} \frac{1}{|\widetilde X|} \sum \limits_{(\widetilde x,y)\in \widetilde X} L(M(x), \widetilde{y}),
	\end{align}
	\begin{align}
	L_{CE}(y, \widetilde y_{LF}) = \sum\limits_{k=1}^9 {y^k} \text{\xspace log \xspace} \widetilde y_{LF}^k,
	\end{align}
	where ${y^k}$ equals to 1 if the sample belongs to class $k$, otherwise 0, and $\widetilde y_{LF}^k$ is the $k$-th value of in $y_{LF}$.
	%There is dataset with augmented samples $\widetilde x\in \widetilde X$ and labels/classes $y\in Y$. Each sample $x$ has a positive pair $\widetilde x_{i}$ and $\widetilde x_{a}$.
\end{enumerate}

\textbf{Implementation of Contrastive Learning.} Following gives us details about centralized model of our framework for multi-modality, which is applied from \cite{bird2020look}. Baseline model with multi-modality has similar neural architecture. \\

\begin{figure}[!h]
	\centering
	\includegraphics[width=1.0\textwidth]{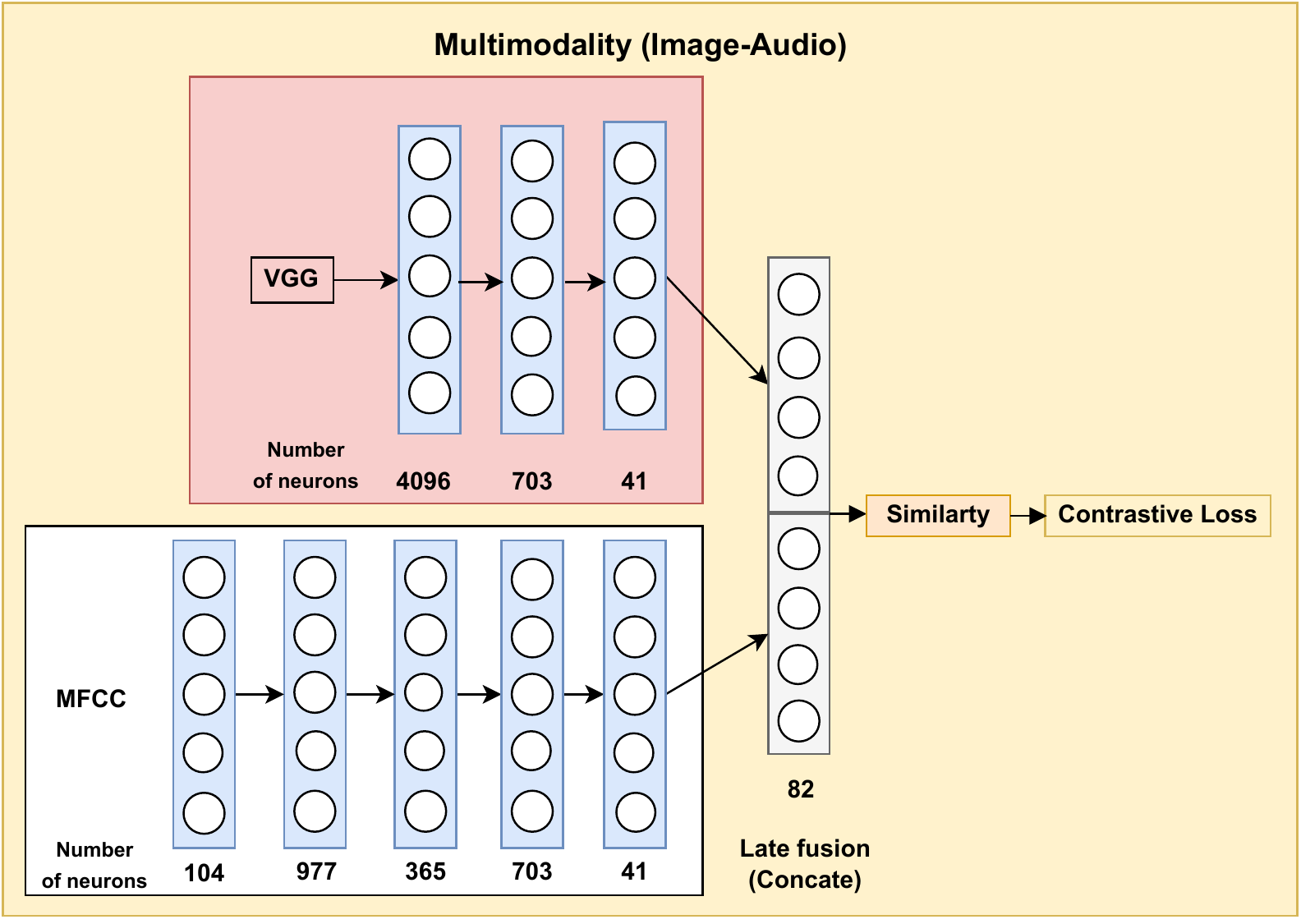}  	  	 	
	\caption{Neural network structure of Contrastive Learning. This process is a detailed description of the Figure \ref{fig:figure_4_6}.}
	\label{fig:figure_5_7}
\end{figure}

Figure \ref{fig:figure_5_7} gives us details about Contrastive Learning for image-audio. The late fusion model only concatenate two sub-networks together. Then, the concatenated outputs are used to compute similarity. Finally, the final outputs are used for calculating contrastive loss.

\begin{enumerate}
	\item\textbf{Data Preprocessing.} In Our framework, we consider a video sample is $x$, a RGB image frame and a audio sequence from this video are two different modalities (\textit{views}): a RGB image frame is transformed by flipping, denoted as $x_i$, a audio sequence is extract by MFCC, denoted as $x_a$.
	
	% need to describe encoders.
	\item\textbf{Encoders $f_i$ and $f_a$.} Encoder (stacked convolutional layers) is the main components in Contrastive Learning. Encoders can be various neural networks. Here, we apply VGG16 for visual modality and the first three layers of DMLP for auditory modality. The representation are $h_{i}=f_i(\widetilde x_{i})$ for $\widetilde x_{i}$ and $h_{a}=f_a(\widetilde x_{a})$ for $\widetilde x_{a}$.
	%In a video sample with multi-modality, we use VGG16 \cite{simonyan2014very} as a encoder for image inputs, and 3 stacked fully connected layers for audio inputs. We obtain the representations from image modality $h_i = f(x_i)$ and audio modality $$. \\
	
	\item\textbf{Project Head $p$.} Projection head $p$ is a shallow neural network, we apply a projection function with several two fully connected layers, projects the representations of two modalities from encoders to hidden space. The aim of project head is to enhance the performance of encoders, and align the represents with a same hidden space shape. The outputs of project head are $z_{i}=p(h_{i})$ for $h_{i}$ and $z_{a}=p(h_{a})$ for $h_{a}$.
	% need to describe the training process.
	
	\item\textbf{Contrastive Loss Function.} The contrastive loss function \cite{gutmann2010noise} is make the encoders to learn the feature representations by themselves. There is dataset with augmented samples $\widetilde X=\{\widetilde x_1, \widetilde x_2, ..., \widetilde x_k\}$. Each sample $x$ has a positive pair $\widetilde x_{i}$ and $\widetilde x_{a}$. The contrastive loss is to maximize the similarity between $\widetilde x_{i}$ and $\widetilde x_{a}$, and minimize the positive pair ($\widetilde x_{i}$, $\widetilde x_{a}$) and other samples. With a batch of $B$ samples, we have $2B$ augmented samples. The contrastive loss function is as follows.
	\begin{align}
	\text{L}_{(i,a)} = -\text{log}\frac{\text{exp}(\frac{\text{cos\_sim}(z_{i},z_{a})}{\tau})}{\sum \limits_{k=1,k\neq i}^{2B} \text{exp}(\frac{\text{cos\_sim}(z_{i},z_{a})}{\tau})},
	\end{align}
	we use cosine similarity \cite{bishop2006pattern}%	With a batch of $B$ samples, we have $mB$ augmented samples.

	\begin{align}
	\text{cos\_sim}(\widetilde x_{i}, \widetilde x_{a}) = \frac{\widetilde x_{i} \cdot \widetilde x_{a}}{\lVert \widetilde x_{i}\lVert \cdot \lVert \widetilde x_{a}\lVert}.
	\end{align}
	At last, the final loss is calculated over all the positive pairs. In a batch with $B$ samples, $\text{L}_{CL}$ is computed as:
	\begin{align}
	\text{L}_{CL} =\frac{1}{2B}\sum \limits_{k=1}^{2B}[L(2k-1, 2k)+L(2k, 2k-1)],
	\end{align}
	where $k$ is the index of samples, $(2k-1, 2k)$ and $(2k, 2k-1)$ are the indices of each positive pair.
\end{enumerate}

We build five comparable experiments. In all experiments, we set hyper-parameters: 100 epochs, 0.001 as learning rate, 10 as batch size, Stochastic Gradient Descent (SGD) as optimizer. Besides, hyperparameters of all federated Learning: 30 participants, 10 as local epochs, 10 as local batch size. The data distribution of each participant and the whole dataset have the same distribution, and sampling follows random sampling without replacement.

% draw a picture for model description
% layer by layer to describe
We use one single GPU (on Colab) with python (version 3.7.3) and PyTorch \cite{pytorch} (version 1.7.1). The training process of Federated Learning with only audio needs 10 minutes. The training process of Federated Learning with only images needs about 2 hours. However, our framework and late fusion model need about 7 hours. \\

Code sources are as follows:
\begin{enumerate}
	\item It is an implement example of Federated Learning without socket communication with PyTorch.
	
	\textit{https://github.com/AshwinRJ/Federated-Learning-PyTorch}\\
	
	\item It is an tutorial which uses Contrastive Learning to learn contrastive representations for the downstream task of music classification. The details are similar in SimCLR \cite{chen2020simple}. 
	
	\textit{https://music-classification.github.io/tutorial/part5\_beyond/self-supervised-learning.html}
\end{enumerate}

% aggregation code description for ...
Because the weight of each sub-network is loaded as the data structure form of dictionary. So we aggregate the weights by keys in dictionaries. Following algorithm (see Algorithm \ref{algorithm1}) shows the aggregation process.
\begin{algorithm}[htbp]
	\caption{Aggregation for model with uni-modality and multi-modality. This algorithm is a detailed description of the Figure \ref{fig:figure_4_6}.}
	\label{algorithm1}
	\begin{algorithmic}[1]
		\Require{lists of each weight for image, audio and image-audio $W_{i}$, $W_{a}$ and $W_{m}$} 
		\Ensure{$w_{i_f}$, $w_{a_f}$, $w_{m_f}$ (aggregated weights for image, audio and image-audio weights)}
		\Statex
		%		Function{Loop}{$A[\;]$}
		\Function{Agg}{$W_u$}  \Comment{Aggregation of weights, $u$ can be $i$, $a$ or $m$}
		\State $w_{u}$ the initialized weight in $W_u$
		\For{\texttt{key in keys of $w_u$}}
		\For{\texttt{$k \gets 1$ to $N$}}  \Comment{$N$ is the length of $W_u$}
		\State \texttt{$w_{u0}$[key] += $W_u$[k][key]}
		\EndFor
		\State \texttt{$w_{u}$[key] = sum($w_{u}$[key])/$N$}
		\EndFor
		\State \Return {$w_{u}$}
		\EndFunction
		\\
		\Function{AGG\_M2U}{$w_u$, $w_m$}  \Comment{Aggregation of a unimodal and a multimodal weight, $u$ represent unimodal, $m$ represent multimodal, $u$ can be $i$ or $a$}
		\State $w_{u_{copy}}$ copy weight in $w_u$
		\State $w_{m_{copy}}$ copy weight in $w_m$
		\State $w_{u_{f}}$ the initialized weight in $w_{u_{f}}$%	With a batch of $B$ samples, we have $mB$ augmented samples.
		
		\For{\texttt{key in keys of $w_{u_{f}}$}}
		\If{key in intersection keys of $w_{u_{copy}}$ and $w_{m_{copy}}$}
		\State \texttt{$w_{u_{f}}$[key] = $w_{u_{copy}}$[key] + $w_{m_{copy}}$[k][key]}
		\State \texttt{$w_{u_{f}}$[key] = sum($w_{u_{f}}$[key])/2}
		\Else
		\State \texttt{$w_{u_{f}}$[key] = $w_{m_{copy}}$[key]}
		\EndIf
		\EndFor
		\State \Return {$w_{u_{f}}$} \Comment{$w_{u_{f}}$ is the aggregated unimodal weight}
		\EndFunction
		\\
		\Function{AGG\_U2M}{$w_i$, $w_a$, $w_m$}  \Comment{Aggregation for a unimodal and a multimodal weight, $i$ represent image, $a$ represent audio, $m$ represent image-audio}
		\State $w_{i_{copy}}$ copy weight in $w_i$
		\State $w_{a_{copy}}$ copy weight in $w_a$
		\State $w_{m_{copy}}$ copy weight in $w_m$
		\State $w_{m_{f}}$ the initialized weight in $w_m$
		\For{\texttt{key in keys of $w_m$}}
		\If{key in intersection keys of $w_{i_{copy}}$ and $w_{m_{copy}}$}
		\State \texttt{$w_{m_{f}}$[key] = $w_{a_{copy}}$[key] + $w_{m_{copy}}$[k][key]}
		\State \texttt{$w_{m_{f}}$[key] = sum($w_{u}$[key])/2}
		\Else
		\State \texttt{$w_{m_{f}}$[key] = $w_{m_{copy}}$[key]}
		\EndIf
		\EndFor
		\\
		\For{\texttt{key in keys of $w_m$}}
		\If{key in intersection keys of $w_{a_{copy}}$ and $w_{m_{copy}}$}
		\State \texttt{$w_{m_{f}}$[key] = $w_{a_{copy}}$[key] + $w_{m_{copy}}$[k][key]}
		\State \texttt{$w_{m_{f}}$[key] = sum($w_{m_{f}}$[key])/2}
		\Else
		\State \texttt{$w_{m_{f}}$[key] = $w_{m_{copy}}$[key]}
		\EndIf
		\EndFor
		\State \Return {$w_{m_{f}}$}
		\EndFunction
		\end{algorithmic}
		\end{algorithm}
		\clearpage
		\begin{algorithm}                 
		\begin{algorithmic} [1]	
		\Function{agg avg}{$W_{i}$, $W_{a}$, $W_{m}$} \Comment{Aggregation for all the weight lists}
		\State $w_{i}$ = AGG($W_i$)
		\State $w_{a}$ = AGG($W_a$)
		\State $w_{m}$ = AGG($W_m$)
		\State $w_{i_f}$ = AGG\_M2U($w_i$, $w_m$)
		\State $w_{a_f}$ = AGG\_M2U($w_a$, $w_m$)
		\State $w_{m_f}$ = AGG\_U2M($w_i$, $w_a$, $w_m$)
		%		avg_i = get_avg_two(w_avg_i, w_avg_m)
		%		avg_a = get_avg_two(w_avg_a, w_avg_m)
		%		w_avg = get_avg_three4multi(w_avg_i, w_avg_a, w_avg_m)
		\State \Return {$w_{i_{f}}$, $w_{a_{f}}$, $w_{m_{f}}$}
		\EndFunction
	\end{algorithmic}
\end{algorithm}

\section{Interaction of Components}
This section aims to explain how Federated Learning, Supervised Learning, Contrastive Learning interact with each other. In other words, we will explain how the whole framework works. The components here indicate all the part of framework. Following figure (Figure \ref{fig:figure_5_8}) shows the interaction and each step in the framework.
\begin{figure}[htbp]
	\centering
	\includegraphics[width=0.88\textwidth]{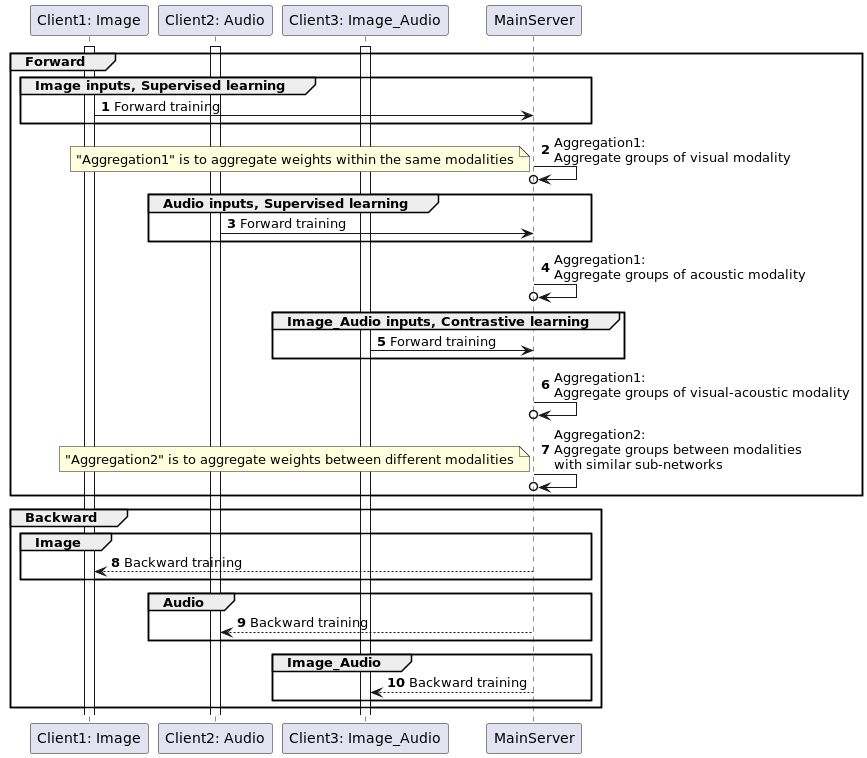}  	  	 	
	\caption{Interaction of components. This is an overview of the workflow of our framework.}
	\label{fig:figure_5_8}
\end{figure}
\section{Summary}
The implementation section shows the models of our framework  and baseline models in detail. We describe the architecture of each model and the number of neurons in each layer. Besides, the two aggregation approaches are described in the algorithm (Algorithm \ref{algorithm1}), which aggregates models within the same modalities or between different modalities. 
%*****************************************
\chapter{Evaluation}
\label{ch:evaluation}
%*****************************************
%\hint{This chapter should describe how the evaluation of the implemented mechanism was done. \\ \\
%1. Which evaluation method is used and why? Simulations, prototype? \\
%2. What is the goal of the evaluation? Comparison? Proof of concept? \\
%3. Which metrics are used for characterizing the performance, costs, fairness, and efficiency of the system?\\
%4. What are the parameter settings used in the evaluation and why? If possible always justify why a certain threshold has been chosen for a particular parameter.  \\
%5. What is the outcome of the evaluation? }

In this chapter, we evaluate the performance of Federated Transfer Learning with visual-audio multimodal data \cite{bird2020look}. In order to better analyze the performance, the implemented baseline models as comparison are required. Accuracy score and confusion matrix are used to evaluate the performance. 

\section{Goal and Methodology}
\label{eval_goal}
% evaluation tool to explain and implement
The goal of the evaluation is to check the effectiveness of our framework compared with baseline models \cite{bird2020look}.
The authors in \cite{bird2020look} used accuracy score and confusion matrix as main evaluation methods. Accuracy score is used to measure the model performance in terms of measuring the ratio of sum of true positives and true negatives out of all the predictions made \cite{powers2020evaluation}. A confusion matrix is a method to summarize the performance of a classification algorithm. Classification accuracy score can not represent the performance of each class \cite{visa2011confusion}. We use both \textit{accuracy score} and \textit{confusion matrix} to evaluate our framework and compare with baseline models. 

%The following is an example of computing accuracy score and confusion matrix.
%The definition and evaluation details are as follows.

\begin{figure}[H]
	\centering
	\includegraphics[width=0.5\textwidth]{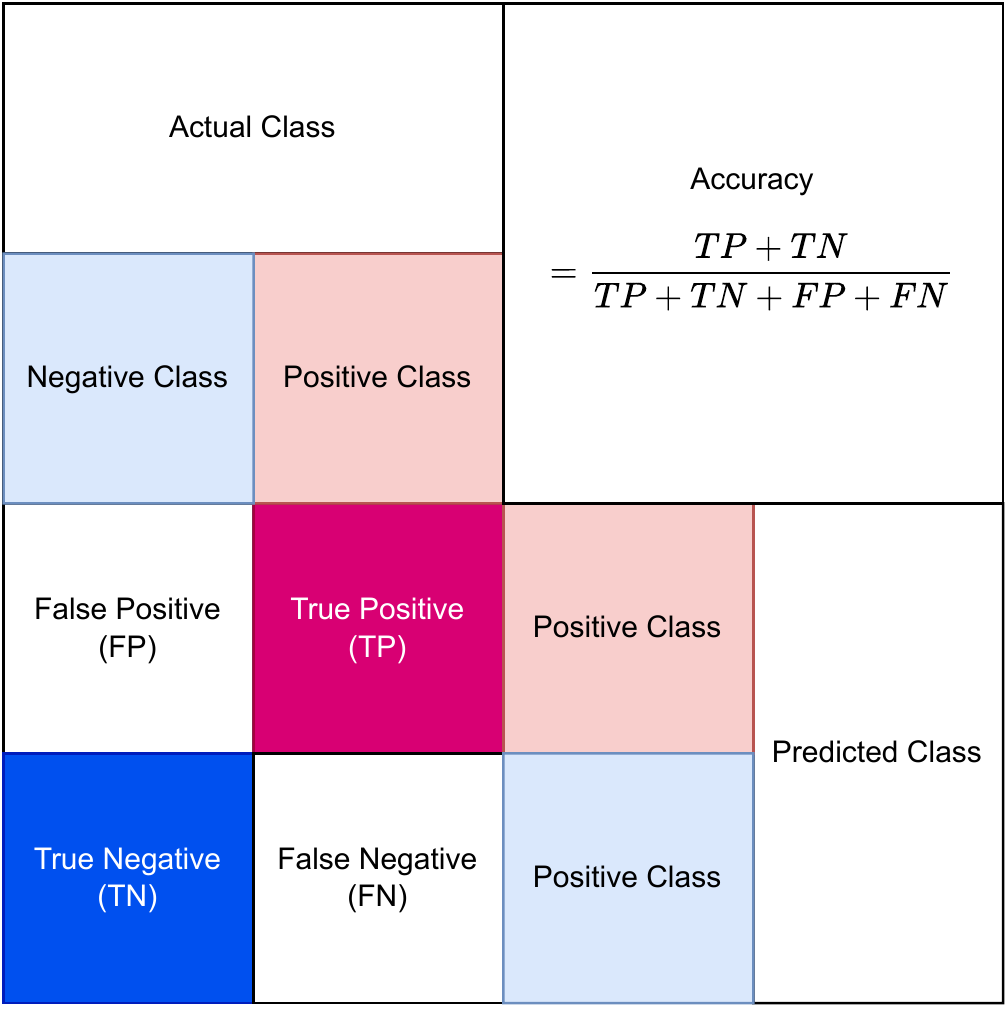}  	  	 	
	\caption{A binary example of confusion matrix. Actual class represents the labels from each sample, and predicted class represents predicted results.} 	  	 	
	%	\caption{Example of Images and Audios for Scene Classification Dataset\cite{bird2020look}}
	\label{fig:figure_6_1}
\end{figure}

\section{Evaluation Setup} 
%Following is the summary of each centralized model's layer settings. Our framework has the similar architecture as the multimodal baseline model, so they are comparable.

We build five experiments to evaluate our framework. We use the metrics above in section \ref{eval_goal}, namely accuracy score and confusion matrix. The first two experiments are for the baseline, and these two experiments have no Transfer Learning. The first experiment is to confirm the effectiveness of centralized baseline models. We add Federated Learning to the centralized model in the second experiment.
This modified model is our major baseline.
\begin{enumerate}
	\item\textbf{Baseline Models without Federated Learning.}
	We evaluate VGG16+MLP for image, MFCC +DMLP for audio, and fused VGG16+MLP and MFCC+DMLP for image-audio without Federated Learning (Section \ref{subsubno_FL}).
	\item\textbf{Baseline Models with Federated Learning.} 
	We apply Federated Learning to baselines model described in previous experiment. Besides, we distribute the training data to 30 participants and each participant with 13800/30=460 samples (Section \ref{subsubonly_FL}).
%	We evaluate VGG16+MLP (Model 1) for image modality, MFCC+DMLP (Model 2) for audio, and fused VGG16+MLP and MFCC+DMLP (Model 3) for image-audio with Federated Learning. We train Model 1, Model 2 and Model 3 separately. Besides, we distribute each Model with 30 participants and each participant with 460 (13800/30) samples. (Section \ref{subsubonly_FL})
\end{enumerate}

The following three experiments are to verify the effectiveness and robustness of our new framework. More specifically, we would like to show the advantage of Federated Transfer Learning, especially when dealing with non-IID data.
The third experiment has the same data distribution of the second experiment to show the basic effectiveness. 
On the other hand, as data in the real world is usually non-IID \cite{sattler2019robust}, we carry out the forth and fifth experiments to verify the robustness of our framework. Intuitively, data in the forth experiment are partially non-IID among the users and data in the fifth experiment are completely non-IID among the users.
The results of all the three experiments below will be compared with the performance of \textbf{Baseline Models with Federated Learning}. 

%All the hyperparameter settings are the same. The difference is the number of samples for each participant. We build different sampling distributions to verify the robustness of our framework in the non-IID setting, where participants have different amounts of samples.
\begin{enumerate}
	\item\textbf{New Design with Balanced Distribution.} The data distribution setting here is the same to that in \textbf{Baseline Models with Federated Learning}.
	To compare with the major baseline, we evaluate VGG16+MLP for image, MFCC+DMLP for audio, and Contrastive Learning based VGG16+MLP and MFCC+DMLP for image-audio with Federated Learning. We divide the participants into three groups. In group one, each participant has data of image uni-modality, and in group two, each participant has data of audio uni-modality. In group three, each participant has data of image-audio multi-modality (see Section \ref{sec:system_overview}). Besides, in each group, there are 30 participants and each participant has 13800/30=460 samples (Section \ref{subsubSSL_all_same}).
	\item\textbf{New Design with Unbalanced Distribution \rom{1}.} In this experiment, we still have three groups and each group has 30 participants. We then assign an index to each participant from 1 to 30 in every group. For participants with the same index, we give them the same number of samples, with the constraint that the total number of samples owned by each group is always 13800. For example, after the initialization, if a participant with index 1 in group 1 has 470 sample, then the participant with index 1 in group 2 must have 470 samples, and participant with index 1 in group 3 must also have 470 samples (Section \ref{subsubSSL_same_diff}).
	 
%	We evaluate VGG16+MLP (Model 1) for image, MFCC+DMLP (Model 2) for audio, and fused VGG16+MLP and MFCC+DMLP (Model 3) for image-audio with Federated Learning. We train Model 1, Model 2 and Model 3 at the same time. Besides, we distribute each Model with 30 participants. If we give each an identity number, we have the same number of samples if the identity (ID) number is the same. For example, Model 1 with ID 1 has 476 samples, Model 2 with ID 1 has 476 samples, Model 3 with ID 1 has 476 samples, and so on. (Section \ref{subsubSSL_same_diff})
	\item\textbf{New Design with Unbalanced Distribution \rom{2}.} In this experiment, we also have three groups and each group has 30 participants. We now allow each participant in each group to have a random number of samples but the total number of samples of each group is always 13800 (Section \ref{subsubSSL_diff_diff}).
%	 We evaluate VGG16+MLP (Model 1) for image, MFCC+DMLP (Model 2) for audio, and fused VGG16+MLP and MFCC+DMLP (Model 3) for image-audio with Federated Learning. We train Model 1, Model 2 and Model 3 at the same time. Besides, we distribute each Model with 30 participants. If we give each an identity number, we have different number of samples even if the identity number is the same. For example, Model 1 with ID 1 has 470 samples, Model 2 with ID 1 has 476 samples, Model 3 with ID 1 has 446 samples, and so on. (Section \ref{subsubSSL_diff_diff})
\end{enumerate}

We carry out the three experiments above to confirm that participants with multimodal data helps participants with unimodal data.
Thus, in the last three experiments, we mainly evaluate the accuracy score and confusion matrix in two uni-modalities.

\section{Evaluation Results}
The Following sections give us details of each experiment. 
%The whole evaluation has three parts: training, validation and testing. We focus on the loss decreasing and accuracy increasing process in the training and validation parts. The following figures show these two processes. Loss is a value that represents the summation of errors. The decreasing process of loss indicates how well our framework is doing. The increasing accuracy process suggests how good our model predictions are by comparing the percentage of the model predictions to the actual value. Besides, The validation process is performed after each training, and validation accuracy is used for early stopping. Early stopping is a regularization method used to avoid over-fitting when a machine learning model is trained in iterative. This method updates the machine learning model to fit the training data at each iteration better. Finally, the accuracy and confusion matrix indicates how well our final prediction is in the testing part. \\

\input{no_FL.tex}
\input{only_FL.tex}
\input{SSL_all_same.tex}
\input{SSL_same_diff.tex}
\input{SSL_diff_diff.tex}

%% need the training and validation figures to describe the loss and accuracy

\section{Analysis of Results}
Our new Federated Transfer Learning framework is applied to multimodal data, where some participants hold a part of data with multi-modality, and the others hold a part of data with only uni-modality. The goal of our framework is that participants have a part of data with multi-modality help the others have a part of data with uni-modality, and it can protect all participants' privacy. 
%The first two experiments above show that our frame achieves the goal, and verify that our framework is effective. 
Because data in the real world are usually non-IID, and non-IID brings often low performance. To confirm our framework in the real world, we apply our framework to different non-IID settings. 
%The last three experiments verify that our framework is robust. 
More specifically, our framework has better performance than the implemented baseline models with Federated Learning. Besides, our framework prevails even when the participants hold a part of data with non-IID distribution, i.e., performs better than Federated Learning with only unimodal data. In addition, when all the participants hold the same portion of data, the testing accuracy of each uni-modality will be close to each other. If there are many participants in one particular scene \cite{bird2020look}, for example, restaurant and grocery-store often achieve low correctness. The experiments confirms that in our framework participants with multimodal data help participants with only unimodal data, which means that participants with multimodal data transfer knowledge to participants with unimodal data.
%*****************************************
\chapter{Conclusions}
\label{ch:closure}
%*****************************************
%\hint{This chapter should self-critically summarize the thesis and describe the main contributions of the thesis. Subsequently, it should describe possible future work in the context of the thesis. What are limitations of the developed solutions? Which things can be improved?}
This chapter is to conclude the thesis and summarize the contributions. Then, we will give the future work beyond the thesis. 
%Besides, we will analyze the limitations of our framework and give the possible idea to improve the framework. % we have solve the problem of the thesis.
\section{Summary and Contribution}
% problems statement
% Data some participants only have unimodal data and others have multimodal data
Data are collected from different devices (participants) using different types (modalities) of sensors. These participants hold a part of data either with multi-modality or only uni-modality. Besides, these participants can not connect directly with each other because of privacy issues. Thus, the problems is how the participants with multimodal data can transfer knowledge to others with only unimodal data.
In this thesis, our new Federated Transfer Learning framework, which combines Federated Learning and Transfer Learning, can solve the above problem. Federated Learning is used to train models from different participants locally with privacy protection. We use the contrastive Learning as the Transfer Learning strategy, which learns transferable representations from participants with multimodal to help participants with unimodal data.
%The highlight in the thesis is that we use Contrastive Learning as a transferring strategy to train multimodal data with similarity. In addition, Contrastive Learning transfers knowledge from participants with multimodal to help participants with unimodal data by average aggregation. We believe that this framework can solve the data islands problem and protect privacy through federated aggregation, even if each participant holds only a small amount of data. In addition, Contrastive Learning needs no labels. So, we can make full use of plenty of unlabeled multimodal data.

The contributions are as follows.
\begin{enumerate}
	\item A new Federated Transfer Learning framework is presented, whose inputs are from input data with unimodal or multimodal data.
	\item Our core transfer learning technique analyzes alignment in precise modalities and uses self-supervision in data with different modalities but data corresponding to (nearly) the same objects.
	\item Experiments over the scene classification dataset (i.e. audio-visual dataset) \cite{bird2020look} show that our method achieves effectiveness and robustness in the sense of transferring knowledge from multi-modality to uni-modality.  
\end{enumerate}

\textbf{Limitations.} There are still limitations in our framework: (1) The participants with unimodal data must perform supervised learning. They still need labeled data. (2) The participants with multimodal data perform contrastive learning, which needs a downstream task to test its performance. However, it is hard to test a downstream task in a participant, because a participant may have only a part of of data. (3) The knowledge transferring in our framework is one directional: Multi-modality helps uni-modality.
\section{Future Work}
%Apply our framework into multimodal data, which have more than two modalities.
We can use more datasets with more than two modalities to verify the effectiveness of our framework. Furthermore, we can try other non-IID settings to verify the its robustness, e.g. participants have only one data category. Finally, we can do further research in the future. 
%For example, scenes 'restaurant' and 'grocery-store' with many people often achieve low correctness to focus the scene classification with more noise.
\begin{enumerate}
	\item We can investigate a new setting of our framework, when all the participants hold unimodal or multimodal data with no labels. All the centralized models are trained with Unsupervised Learning.
	\item In order to achieve transferable features between different modalities, we can try other distance matrices (not cosine similarity).
	\item We can investigate the possibility of other Federated Transfer Learning framework, in which participants with unimodal data help those with multimodal data.
\end{enumerate}
%\section{Final Remarks}

%% file: no_FL.tex
% baseline model reimplementation, only single modal and without federated learning
\subsection{Baseline Models without Federated Learning.}\label{subsubno_FL}
Figure 6.2 to Figure 6.7 show the performance of the implementation of baseline models without Federated Learning. We train all the baseline models \cite{bird2020look} using the whole training dataset with 100 epochs. We train baseline models without Federated Learning three times, and obtain the average accuracy and confusion matrix. 

We implement baseline models in \cite{bird2020look}, VGG16+MLP for image, and MFCC+DMLP for audio and VGG16+MLP and MFCC+DMLP for image-audio are trained from scratch. 
The accuracy is about 7\% higher than the baseline in image modality. The implementation of auditory brings about 1\% lower accuracy than the baseline. The accuracy of image-audio multi-modality is close to the baseline (Table \ref{tab:table_6_1}).

\begin{figure}[!h]
	\centering
	\begin{minipage}[H]{0.49\textwidth}
		\includegraphics[width=1.0\textwidth]{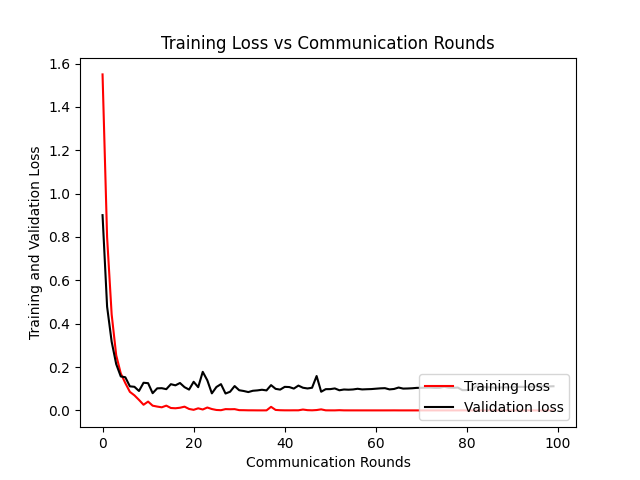}  	  	 
	\end{minipage}
	\begin{minipage}[H]{0.49\textwidth}
		\includegraphics[width=1.0\textwidth]{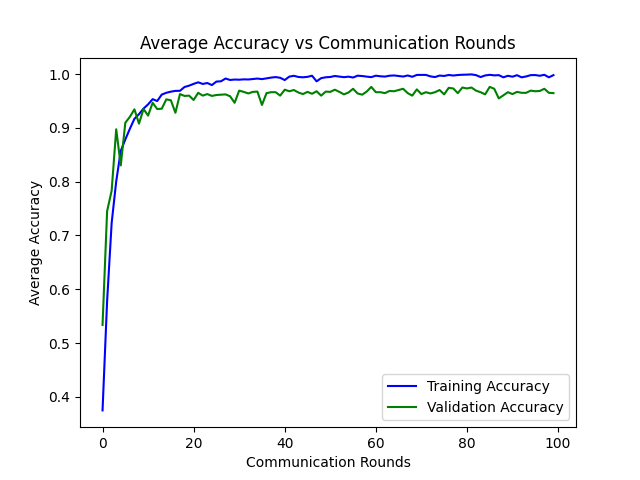}
	\end{minipage}
	\caption{Loss and accuracy for \textbf{image modality}. The Loss of training and validation decreases when the number of epochs of communications increases (left). The Accuracy of training and validation increases when the number of epochs of communications decreases (right).}
	\label{fig:figure_6_2}
\end{figure}

\begin{figure}[!h]
	\centering
	\begin{minipage}[H]{0.49\textwidth}
		\includegraphics[width=1.0\textwidth]{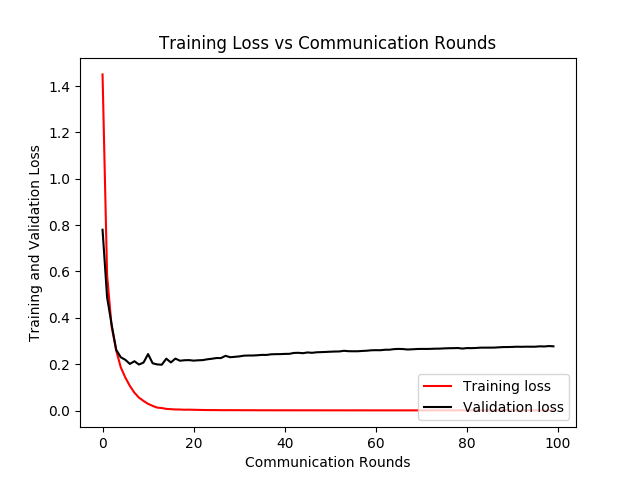}  	  	 
	\end{minipage}
	\begin{minipage}[H]{0.49\textwidth}
		\includegraphics[width=1.0\textwidth]{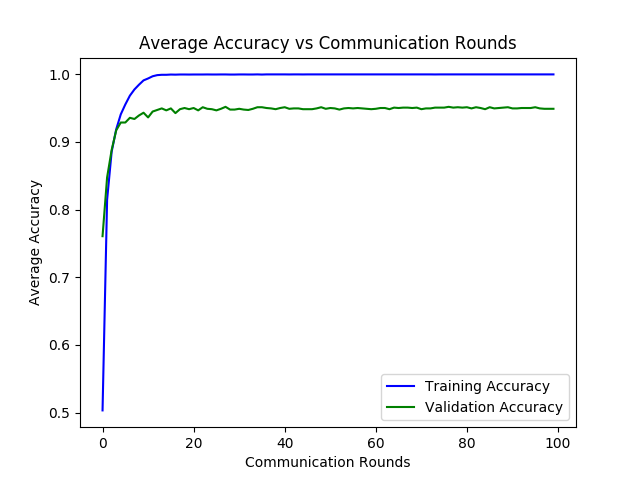}
	\end{minipage}
	\caption{Loss and accuracy for \textbf{audio modality}. The Loss of training and validation decreases when the number of epochs of communications increases (left). The Accuracy of training and validation increases when the number of epochs of communications decreases (right).}
	\label{fig:figure_6_3}
\end{figure}

\begin{figure}[!h]
	\centering
	\begin{minipage}[H]{0.49\textwidth}
		\includegraphics[width=1.0\textwidth]{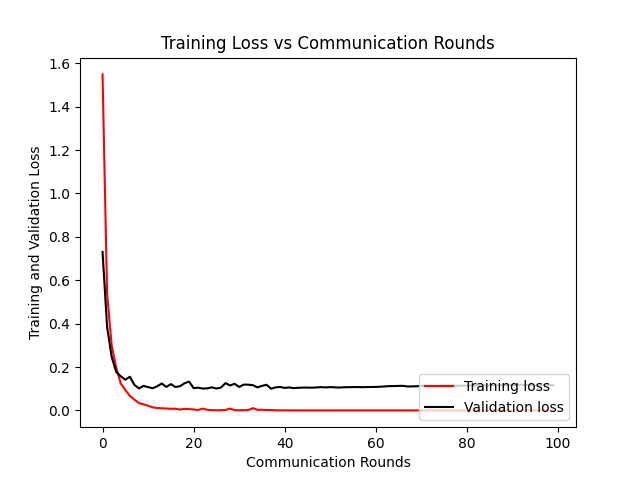}  	  	 
	\end{minipage}
	\begin{minipage}[H]{0.49\textwidth}
		\includegraphics[width=1.0\textwidth]{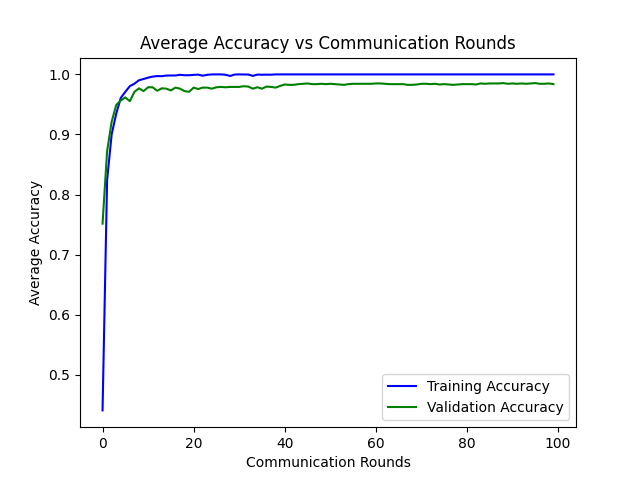}  
	\end{minipage}
	\caption{Loss and accuracy for \textbf{image-audio multi-modality}. The Loss of training and validation decreases when the number of epochs of communications increases (left). The Accuracy of training and validation increases when the number of epochs of communications decreases (right).}
	\label{fig:figure_6_4}
\end{figure}

The training losses of implemented and baseline models dramatically decrease in 10 global communication epochs, and the losses remain at 0.05 after ten global epochs. However, all validation losses decrease with jitters, and validation loss of auditory modality increases slowly after 20 global epochs. Besides, all training and validation accuracies increase dramatically in 10 global epochs and remain stable after 20 global epochs except for image modality. The validation accuracy of image modality remains unchanged with jitter (See Figures \ref{fig:figure_6_2}, \ref{fig:figure_6_3} and \ref{fig:figure_6_4}).

\begin{figure}[!h]
	\centering
	\begin{minipage}[H]{0.49\textwidth}
		\includegraphics[width=1.0\textwidth]{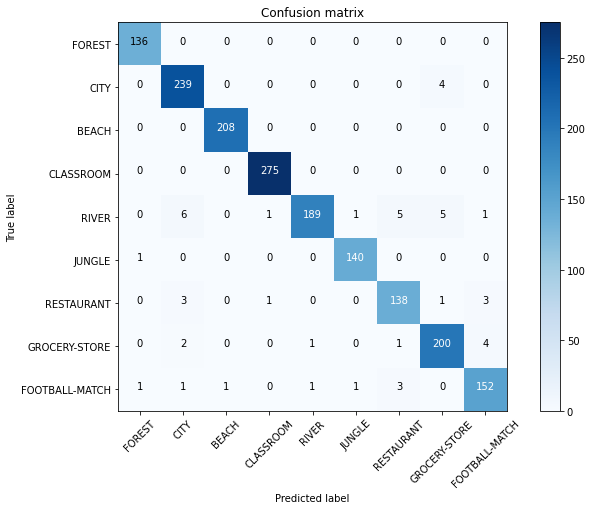}  	  	 
	\end{minipage}
	\begin{minipage}[H]{0.49\textwidth}
		\includegraphics[width=1.0\textwidth]{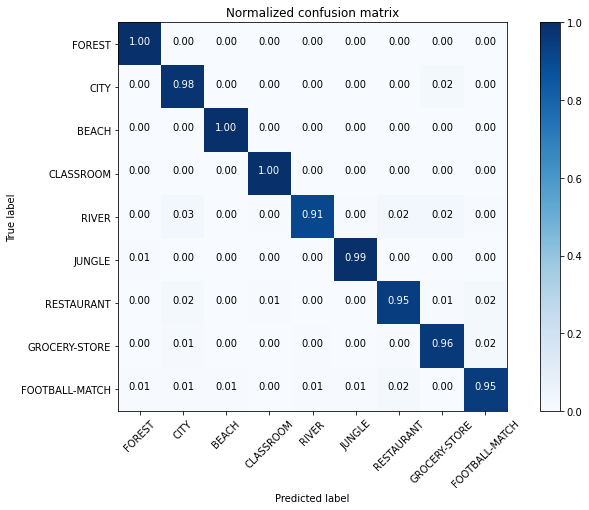}  	
	\end{minipage}
	\caption{Confusion matrices for \textbf{image modality}. The confusion matrix on testing dataset (left), and the normalized confusion matrix of testing dataset (right).}
	\label{fig:figure_6_5}
\end{figure}

\begin{figure}[!h]
	\centering
	\begin{minipage}[H]{0.49\textwidth}
		\includegraphics[width=1.0\textwidth]{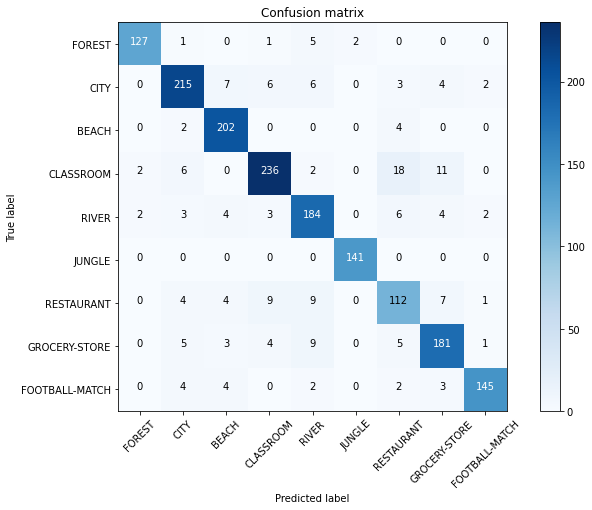}  	  	 
	\end{minipage}
	\begin{minipage}[H]{0.49\textwidth}
		\includegraphics[width=1.0\textwidth]{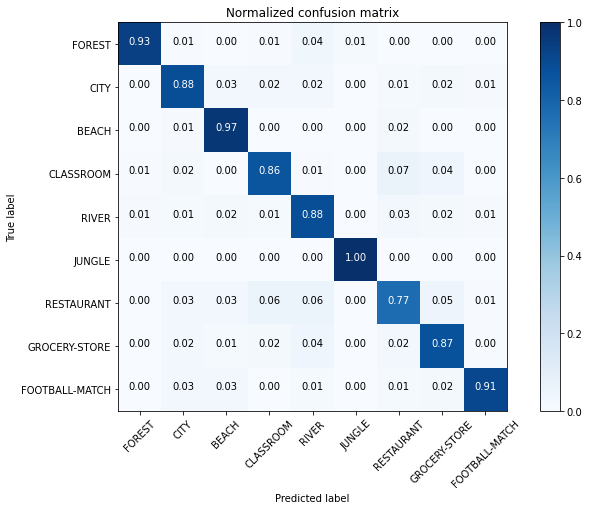}  	
	\end{minipage}
	\caption{Confusion matrices for \textbf{audio modality}. The confusion matrix on testing dataset (left), and the normalized confusion matrix of testing dataset (right).}
	\label{fig:figure_6_6}
\end{figure}

\begin{figure}[!h]
	\centering
	\begin{minipage}[H]{0.49\textwidth}
		\includegraphics[width=1.0\textwidth]{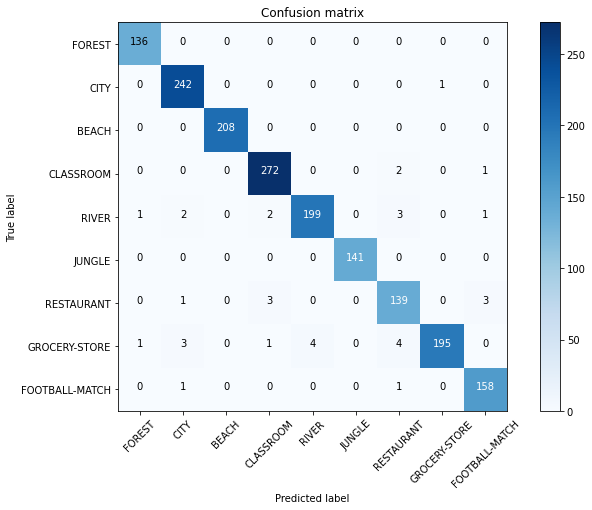}  	  	 
	\end{minipage}
	\begin{minipage}[H]{0.49\textwidth}
		\includegraphics[width=1.0\textwidth]{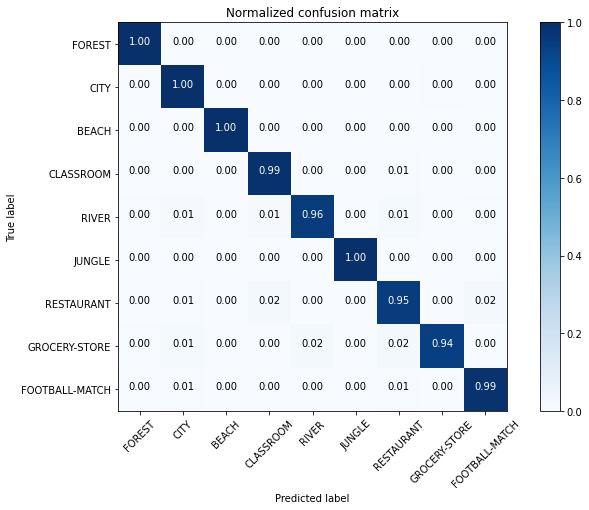}  	
	\end{minipage}
	\caption{Confusion matrices for \textbf{image-audio modality}. The confusion matrix on testing dataset (left), and the normalized confusion matrix of testing dataset (right).}
	\label{fig:figure_6_7}
\end{figure}

The confusion matrices for two uni-modalities (image using VGG16+MLP and audio using MFCC+DMLP) show that each category is correctly classified. All the categories can be over 90\% when the confusion matrices are normalized. (see Figures \ref{fig:figure_6_5}, \ref{fig:figure_6_6}) However, the predicted labels and true labels ratios of categories \textit{city}, \textit{classroom}, \textit{restaurant} and \textit{grocery-store} are lower than 90\% showed in the normalized confusion matrix in multi-modality (see Figure \ref{fig:figure_6_7}).
% describe the figures of 
\begin{table}[htbp]
	\centering
	\begin{tabular}{|l|l|l|}
		\hline 
%		\hline 
		\textbf{Modality} & \textbf{Average Testing Accuracy} &\makecell{\textbf{Testing Accuracy} \\ \textbf{(Baselines \cite{bird2020look})}}\\
		\hline 
%		\hline 
		\makecell[l]{\textbf{Image (Visual)} \\ VGG16+MLP} & 96.31\% & 89.27\%\\
		\hline
		\makecell[l]{\textbf{Audio (Auditory)}\\ MFCC+DMLP}& 92.99\% & 93.72\%\\
		\hline
		\makecell[l]{\textbf{Image-Audio (Multimodality)}\\ VGG16+MLP and MFCC+DMLP} & 97.55\% & 96.81\% \\
		\hline 
	\end{tabular}
	\caption{Average performance is about uni-modality and multi-modality without federated learning. The training process begins from scratch.}
	\label{tab:table_6_1}
\end{table}

%Only for image without federated learning 
%Only for audio without federated learning 
%Only for image-audio without federated learning 

%CM Only for image without federated learning 

%CM Only for audio without federated learning 

%Only for image-audio without federated learning 

%% file: only_FL.tex
\subsection{Baseline Models with Federated Learning.}\label{subsubonly_FL}
Figure 6.8 to 6.13 show the performance over image, audio and image-audio using baseline models as centralized models with Federated Learning when the participants have the same size of samples. We set up 30 participants, each holding 13800/30=460 samples. All the participants have similar distribution with the whole training dataset. In total, we set 100 global epochs for global models. We train baseline models with Federated Learning three times, and achieve the average accuracy and confusion matrix.

The accuracies of all the implemented baseline models with Federated Learning are lower than those without Federated Learning.
More specifically, the accuracies of baseline models in Federated Learning are about 3.5\% lower than those without Federated Learning in two uni-modalities. In addition, the accuracy of the implemented baseline model of multi-modality with Federated Learning is about 1\% lower than the same central model with no Federated Learning (Table \ref{tab:table_6_2}).

\begin{figure}[h]
	\centering
	\begin{minipage}[H]{0.49\textwidth}
		\includegraphics[width=1.0\textwidth]{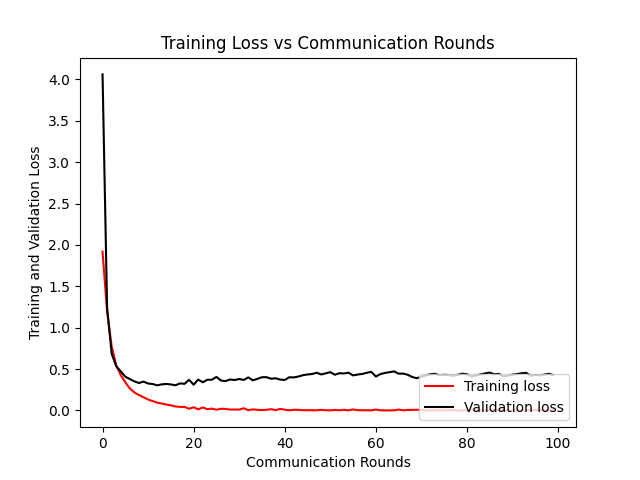}  	  	 
	\end{minipage}
	\begin{minipage}[H]{0.49\textwidth}
		\includegraphics[width=1.0\textwidth]{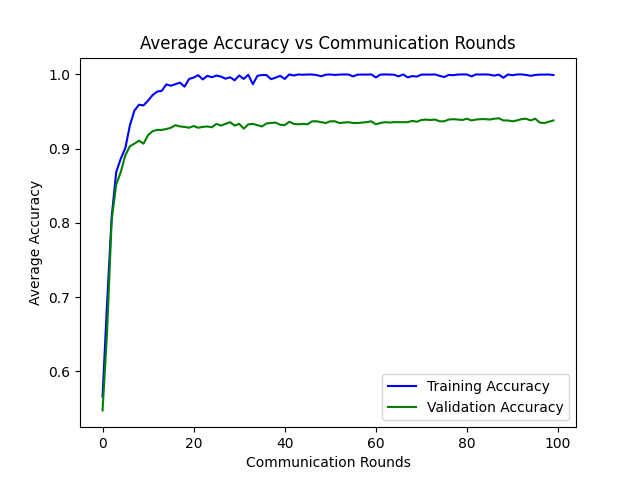}  
	\end{minipage}
	\caption{Loss and accuracy for \textbf{image modality}. The Loss of training and validation decreases when the number of epochs of communications increases (left). The Accuracy of training and validation increases when the number of epochs of communications decreases (right).}
	\label{fig:figure_6_8}
\end{figure}

\begin{figure}[h]
	\centering
	\begin{minipage}[H]{0.49\textwidth}
		\includegraphics[width=1.0\textwidth]{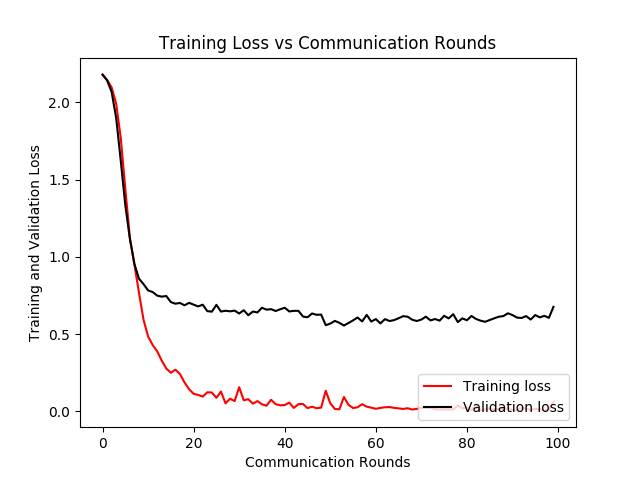}  	  	 
	\end{minipage}
	\begin{minipage}[H]{0.49\textwidth}
		\includegraphics[width=1.0\textwidth]{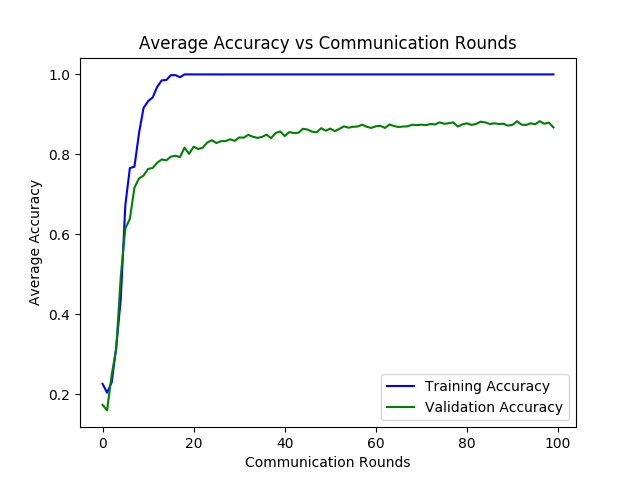}  
	\end{minipage}
	\caption{Loss and accuracy for \textbf{audio modality}. The Loss of training and validation decreases when the number of epochs of communications increases (left). The Accuracy of training and validation increases when the number of epochs of communications decreases (right).}
	\label{fig:figure_6_9}
\end{figure}

\begin{figure}[!h]
	\centering
	\begin{minipage}[H]{0.49\textwidth}
		\includegraphics[width=1.0\textwidth]{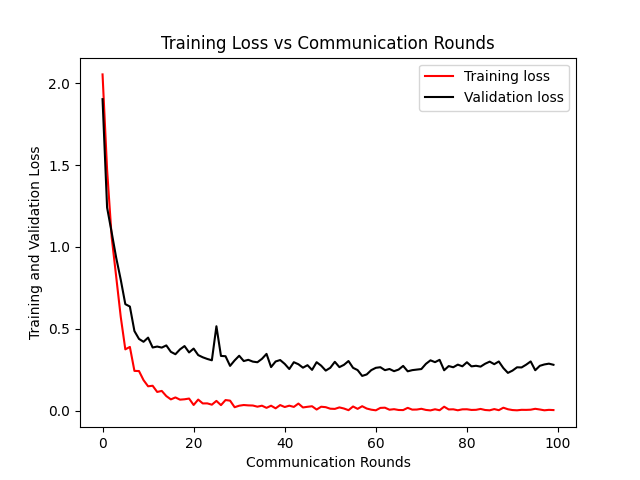}  	  	 
	\end{minipage}
	\begin{minipage}[H]{0.49\textwidth}
		\includegraphics[width=1.0\textwidth]{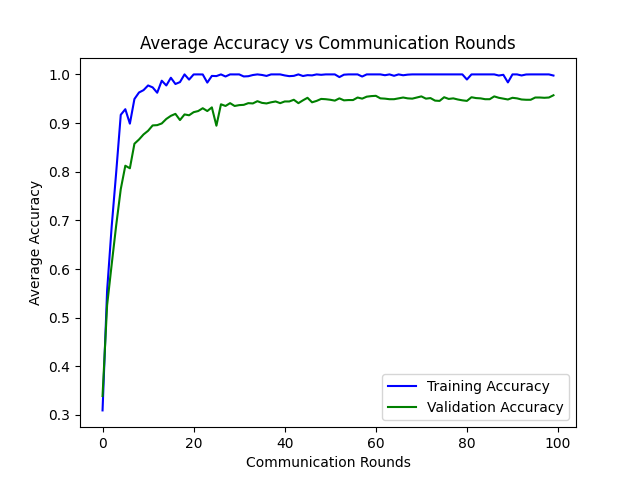}  
	\end{minipage}
	\caption{Loss and accuracy for \textbf{image-audio multi-modality}. The Loss of training and validation decreases when the number of epochs of communications increases (left). The Accuracy of training and validation increases when the number of epochs of communications decreases (right).}
	\label{fig:figure_6_10}
\end{figure}

The training and validation losses of implemented baseline models with Federated Learning dramatically decrease in 10 global epochs, and they remain unchanged after 20 global epochs. 
The validation losses of image and image-audio modalities remain unchanged with jitter at 0.25 after 20 global epochs, while it stays at 0.75 for audio modality.
Besides, all training and validation accuracies increase dramatically in 10 global epochs and remain stable with slight jitter after 20 global epochs (Figures \ref{fig:figure_6_8}, \ref{fig:figure_6_9} and \ref{fig:figure_6_10}). 

\begin{figure}[!h]
	\centering
	\begin{minipage}[H]{0.49\textwidth}
		\includegraphics[width=1.0\textwidth]{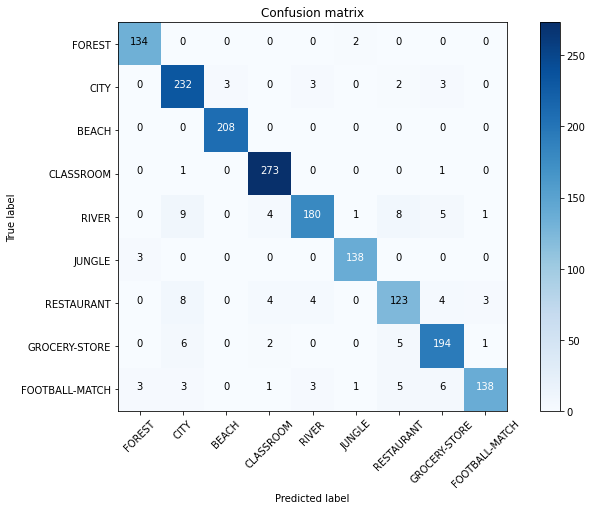}  	  	 
	\end{minipage}
	\begin{minipage}[H]{0.49\textwidth}
		\includegraphics[width=1.0\textwidth]{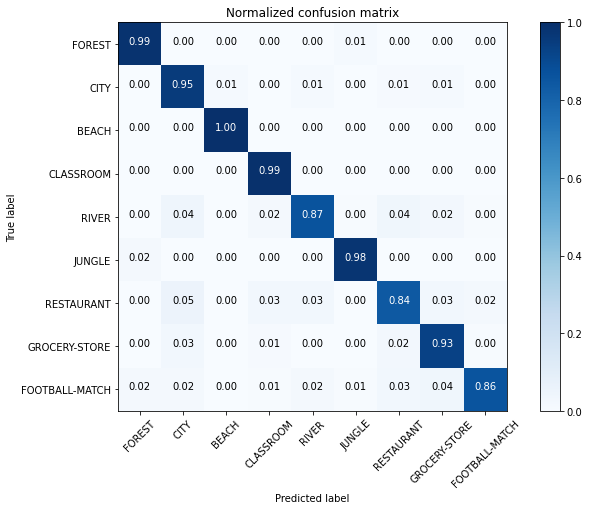}  
	\end{minipage}
	\caption{Confusion matrices for \textbf{image modality}. The confusion matrix on testing dataset (left), and the normalized confusion matrix of testing dataset (right).}
	\label{fig:figure_6_11}
\end{figure}

\begin{figure}[!h]
	\centering
	\begin{minipage}[H]{0.49\textwidth}
		\includegraphics[width=1.0\textwidth]{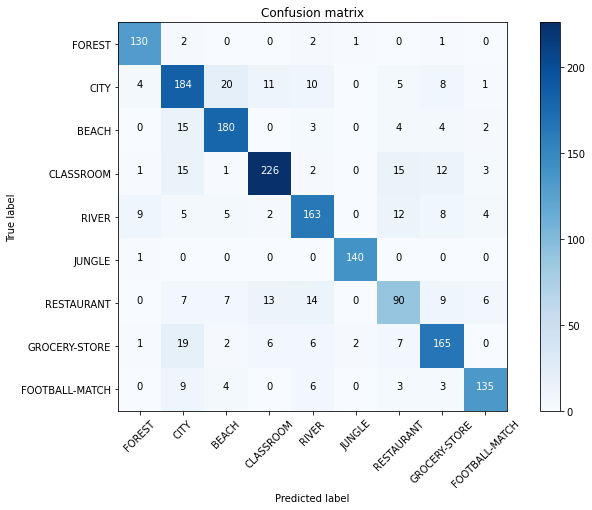}  	  	 
	\end{minipage}
	\begin{minipage}[H]{0.49\textwidth}
		\includegraphics[width=1.0\textwidth]{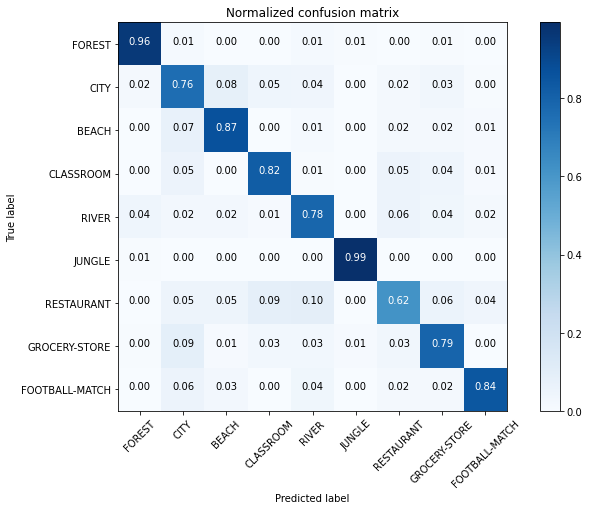}  
	\end{minipage}
	\caption{Confusion matrices for \textbf{audio modality}. The confusion matrix on testing dataset (left), and the normalized confusion matrix of testing dataset (right).}
	\label{fig:figure_6_12}
\end{figure}

\begin{figure}[!h]
	\centering
	\begin{minipage}[H]{0.49\textwidth}
		\includegraphics[width=1.0\textwidth]{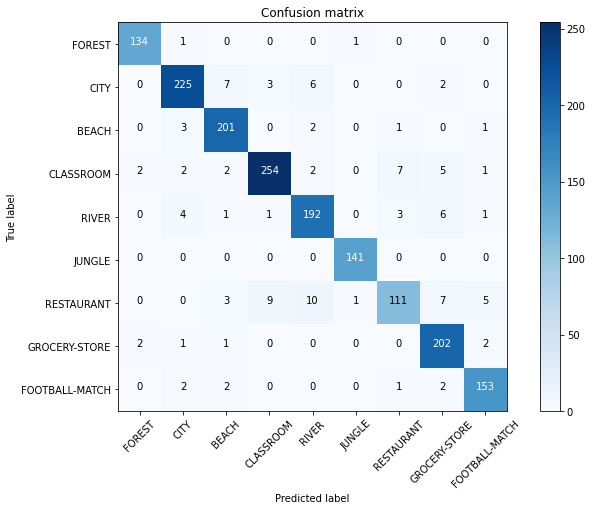}  	  	 
	\end{minipage}
	\begin{minipage}[H]{0.49\textwidth}
		\includegraphics[width=1.0\textwidth]{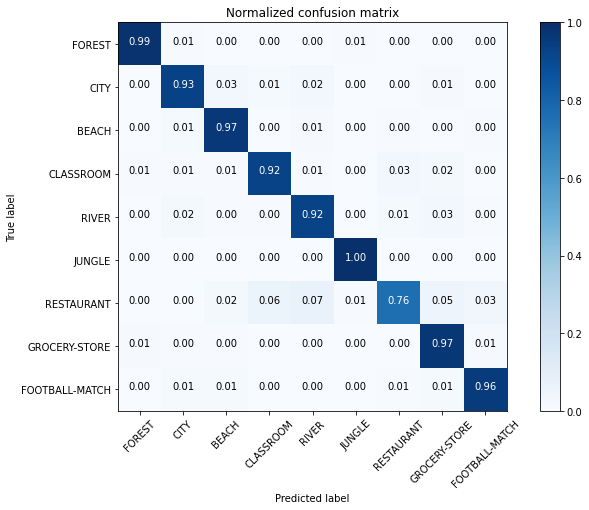}  
	\end{minipage}
	\caption{Confusion matrices for \textbf{image-audio multi-modality}. The confusion matrix on testing dataset (left), and the normalized confusion matrix of testing dataset (right).}
	\label{fig:figure_6_13}
\end{figure}

All confusion matrices show how well each category is classified. On the one hand, for image modality, \textit{river} and \textit{restaurant} can be about 85\%, and the other categories can be over 90\% when the confusion matrices are normalized. On the other hand, for audio modality, the predicted labels and true labels ratios of categories \textit{city}, \textit{river}, \textit{restaurant}, \textit{grocery-store} are lower than 80\%.
However, most categories are lower than 90\% in the normalized confusion matrix in multi-modality except for \textit{restaurant} (Figures \ref{fig:figure_6_11}, \ref{fig:figure_6_12} and \ref{fig:figure_6_13}).

\begin{table}[htbp]
	\centering
	\begin{tabular}{|l|l|l|}
		\hline 
		\textbf{Modality} &\makecell{\textbf{Average Testing Accuracy} \\ (Implemented Baselines \\ \textbf{with} Federated Learning)} &\makecell{\textbf{ Average Testing Accuracy} \\ (Implementd Baselines \\ \textbf{without} Federated Learning)}\\
		\hline 
		\makecell{\textbf{Image (Visual)} \\ VGG16+MLP} & 93.68\% & 96.31\%\\
		\hline
		\makecell{\textbf{Audio (Auditory)} \\ MFCC+DMLP}& 88.16\% & 92.99\%\\
		\hline
		\makecell{\textbf{Image-Audio (Multimodality)} \\ VGG16+MLP and MFCC+DMLP} & 96.41\% & 97.55\% \\
		\hline 
	\end{tabular}
	\caption{Average performance of un-modality and multi-modality with Federated Learning.}
	\label{tab:table_6_2}
\end{table}

% FL for image

% FL for audio

% FL for image-audio

%CM FL for image

%CM FL for audio

%CM FL for image-audio

%% file: SSL_all_same.tex
% loss, multiloss, acc for self-supervision when the pariticipants have different number 460v460
\subsection{New Design with Balanced Distribution.}\label{subsubSSL_all_same}
\begin{figure}[!h]
	\centering
	\begin{minipage}[H]{0.49\textwidth}
		\includegraphics[width=1.0\textwidth]{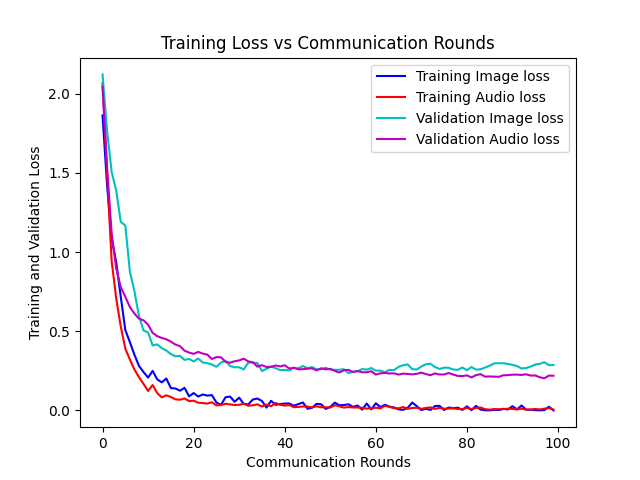} 	  
	\end{minipage}
	\begin{minipage}[H]{0.49\textwidth}
		\includegraphics[width=1.0\textwidth]{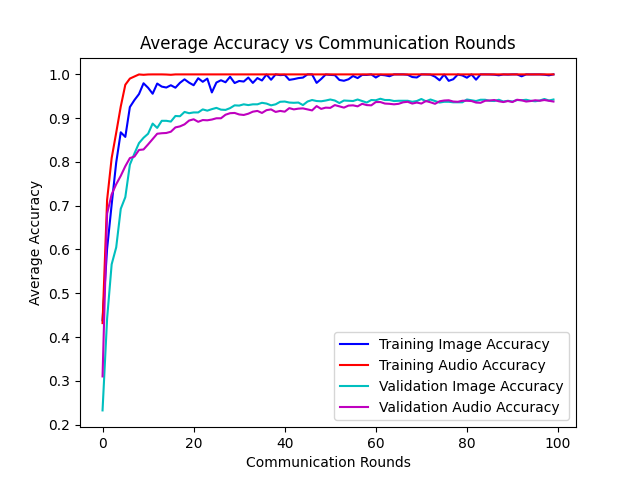}  
	\end{minipage}
	\begin{minipage}[H]{0.49\textwidth}
		\includegraphics[width=1.0\textwidth]{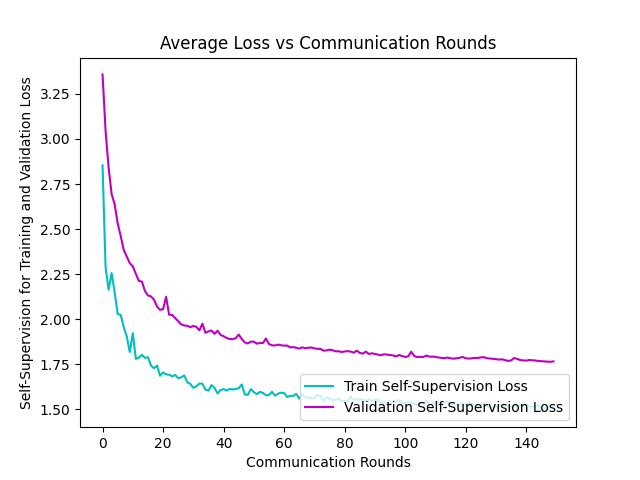}  
	\end{minipage}
	\caption{Loss and accuracy for \textbf{our framework}. The loss of training and validation of unimodal supervision decreases when the number of epochs of communications increases (left). The accuracy of training and validation increases when the number of epochs of communications decreases (right). The loss of training and validation of multi-modal self-supervision decreases when the number of epochs of communications increases (bottom).}
	\label{fig:figure_6_14}
\end{figure}

\begin{figure}[!h]
	\centering
	\begin{minipage}[H]{0.49\textwidth}
		\includegraphics[width=1.0\textwidth]{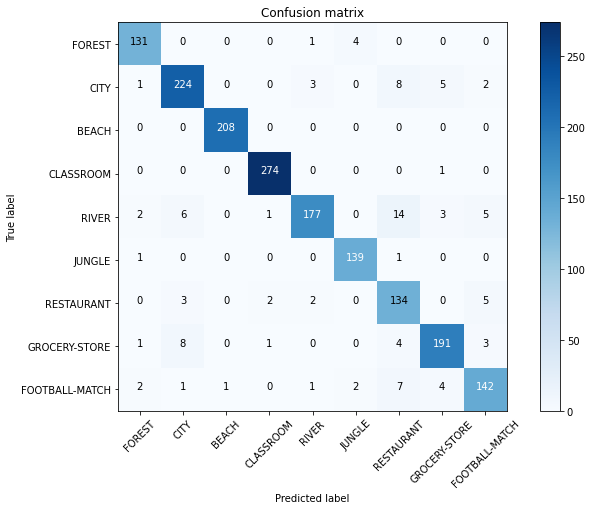}
	\end{minipage}
	\begin{minipage}[H]{0.49\textwidth}
		\includegraphics[width=1.0\textwidth]{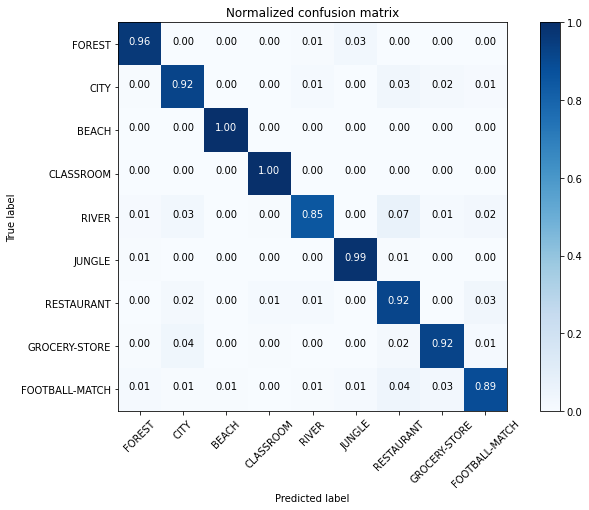}  
	\end{minipage}
	\caption{Confusion matrices for \textbf{image modality}. The confusion matrix on testing dataset (left), and the normalized confusion matrix of testing dataset (right).}
	\label{fig:figure_6_15}
\end{figure}

\begin{figure}[!h]
	\centering
	\begin{minipage}[H]{0.49\textwidth}
		\includegraphics[width=1.0\textwidth]{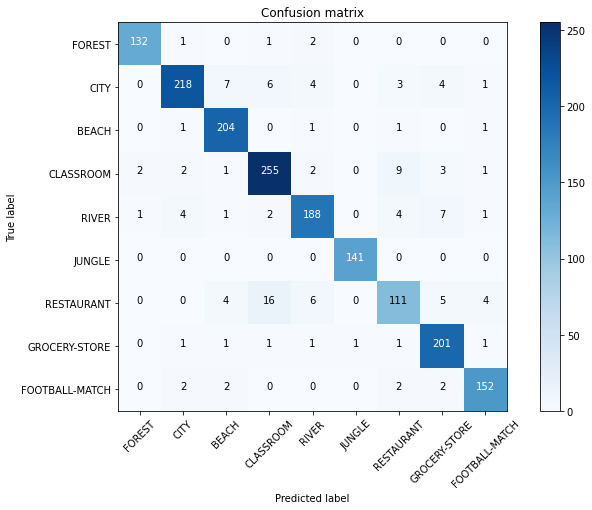}
	\end{minipage}
	\begin{minipage}[H]{0.49\textwidth}
		\includegraphics[width=1.0\textwidth]{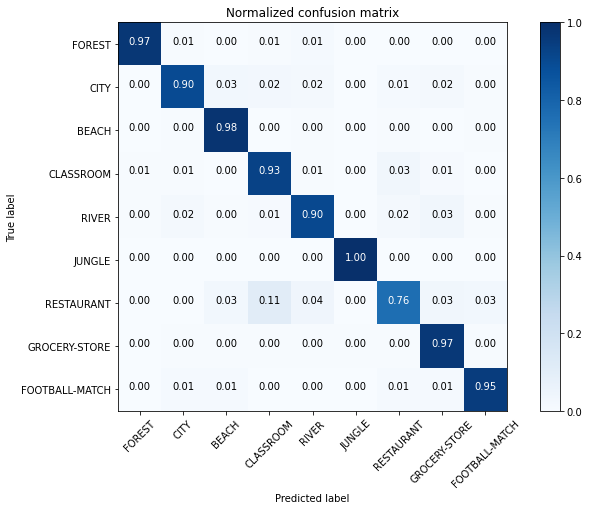}  
	\end{minipage}
	\caption{Confusion matrices for \textbf{audio modality}. The confusion matrix on testing dataset (left), and the normalized confusion matrix of testing dataset (right).}
	\label{fig:figure_6_16}
\end{figure}

Figures \ref{fig:figure_6_14}, \ref{fig:figure_6_15} and \ref{fig:figure_6_16} show the average performance of our framework. The values are computed from averaging the results of three trained models with randomized initialization. We divide 90 participants into 3 groups, 30 for image modality as group 1, 30 for audio modality as group 2, and 30 for image-audio modality as group 3, each participant of each modality has 13800/30=460 samples. The data samples in all participants are randomly assigned with 460 sample, and the total number of samples owned by a group is always 13800. This assignment is an independent sampling without replacement. In total, we set 100 global epochs for global models.

The accuracies of our framework (Federated Transfer Learning) are higher than the implemented baseline models with Federated Learning.
More specifically, the accuracy of image modality is about 0.5\% higher, and that of audio modality is about 5\% improved (Table \ref{tab:table_6_3}).

Our framework's training and validation losses for two uni-modalities dramatically decrease in 10 global epochs, and they remain stable after 20 global epochs. 
The training losses of image and audio modalities remain unchanged with jitter at 0.25 after 20 global epochs, while the validation of those remains stable with jitter at 0.3. Besides, after 60 global epochs, the validation loss of image modality increases negligibly.
For image-audio multi-modality (Contrastive Learning), its training and validation losses dramatically decrease in 40 global epochs and remain unchanged after 60 global epochs.
Moreover, all training and validation accuracies increase dramatically in 10 global epochs and remain stable with slight jitter after 40 global epochs. From 80 to 100 global epochs, the validation accuracies of image and audio overlap (Figures \ref{fig:figure_6_14}).

Most categories are correctly predicted, and true label ratios are higher than 90\% in the normalized confusion matrix for image modality, although \textit{football-match} has only 89\%. Compared with the image, the correctness of \textit{restaurant} is only 76\%, respectively, and the rest categories are higher than 90\% (Figures \ref{fig:figure_6_15} and \ref{fig:figure_6_16}).

\begin{table}[htbp]
	\centering
	\begin{tabular}{|l|l|l|}
		\hline  
		\textbf{Modality} & \makecell{\textbf{Testing Accuracy}\\ Our Framework} & \makecell[l]{\textbf{Testing Accuracy} \\ (Implemented Baselines Models \\ \textbf{with} Federated Learning)} \\
		\hline 
		\makecell[l]{\textbf{Image (Visual)}} & 93.91\% & 93.68\%\\
		\hline
		\makecell[l]{\textbf{Audio (Auditory)}} & 92.87\% & 88.16\%\\
		\hline 
	\end{tabular}
	\caption{Average accuracy of our Framework.}
	\label{tab:table_6_3}
\end{table}

% CM for self-supervision when the pariticipants have different number 10v10; 20v20

% CM for self-supervision when the pariticipants have different number 10v20; 10v15

%% file: SSL_same_diff.tex
\subsection{New Design with Unbalanced Distribution \rom{1}.}\label{subsubSSL_same_diff}
\begin{figure}[!h]
	\centering
	\begin{minipage}[H]{0.49\textwidth}
		\includegraphics[width=1.0\textwidth]{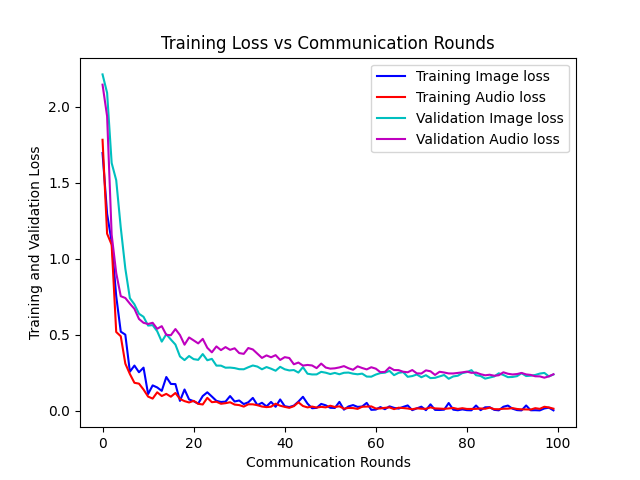}  	  	 
	\end{minipage}
	\begin{minipage}[H]{0.49\textwidth}
		\includegraphics[width=1.0\textwidth]{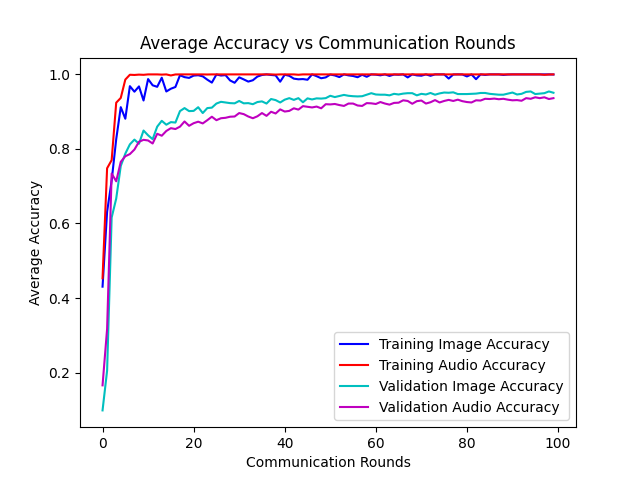}  
	\end{minipage}
	\begin{minipage}[H]{0.49\textwidth}
		\includegraphics[width=1.0\textwidth]{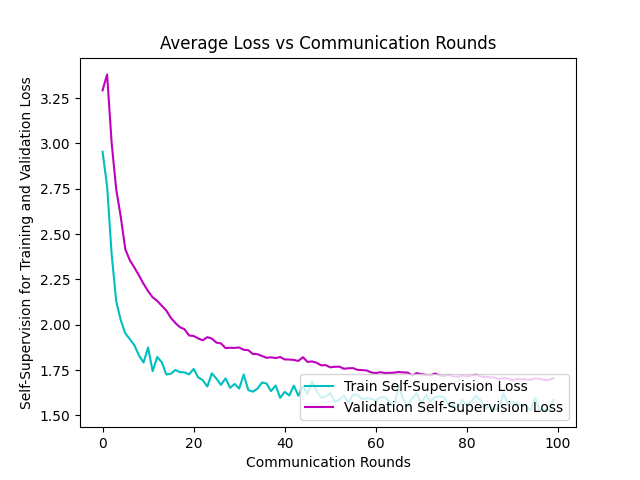}  
	\end{minipage}
	\caption{Loss and accuracy for \textbf{our framework}. The loss of training and validation of unimodal supervision decreases when the number of epochs of communications increases (left). The accuracy of training and validation increases when the number of epochs of communications decreases (right). The loss of training and validation of multi-modal self-supervision decreases when the number of epochs of communications increases (bottom). }
	\label{fig:figure_6_17}
\end{figure}

\begin{figure}[!h]
	\centering
	\begin{minipage}[H]{0.49\textwidth}
		\includegraphics[width=1.0\textwidth]{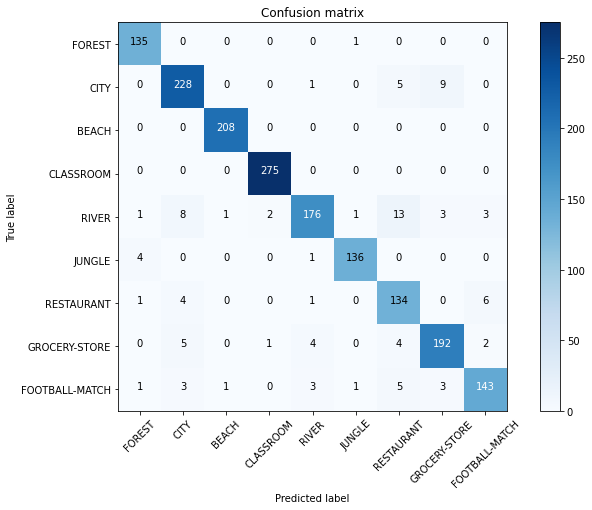}
	\end{minipage}
	\begin{minipage}[H]{0.49\textwidth}
		\includegraphics[width=1.0\textwidth]{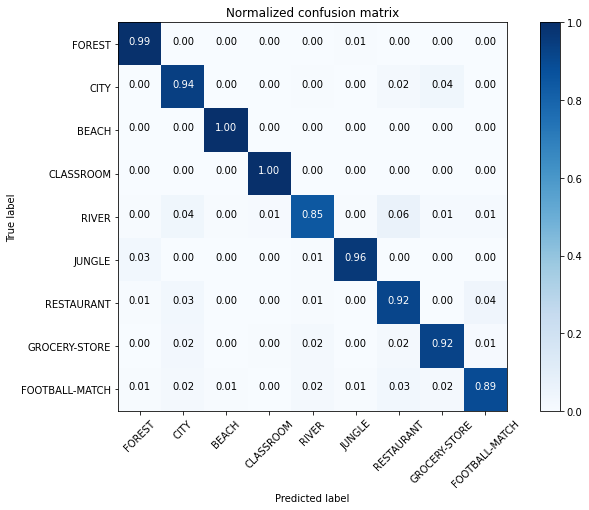}  
	\end{minipage}
	\caption{Confusion matrices for \textbf{image modality}. The confusion matrix on testing dataset (left), and the normalized confusion matrix of testing dataset (right).}
	\label{fig:figure_6_18}
\end{figure}

\begin{figure}[!h]
	\centering
	\begin{minipage}[H]{0.49\textwidth}
		\includegraphics[width=1.0\textwidth]{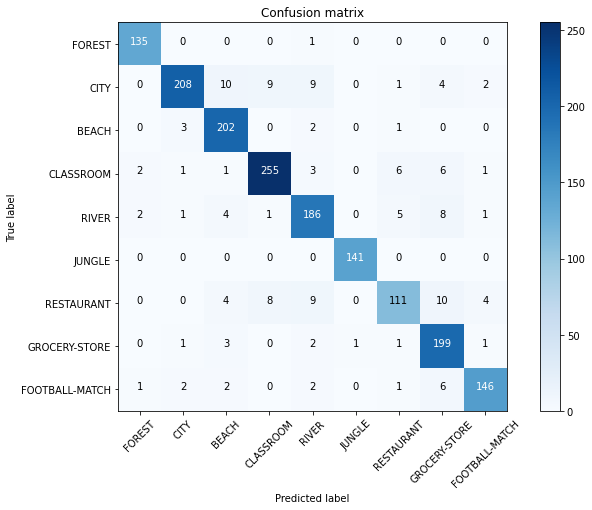}
	\end{minipage}
	\begin{minipage}[H]{0.49\textwidth}
		\includegraphics[width=1.0\textwidth]{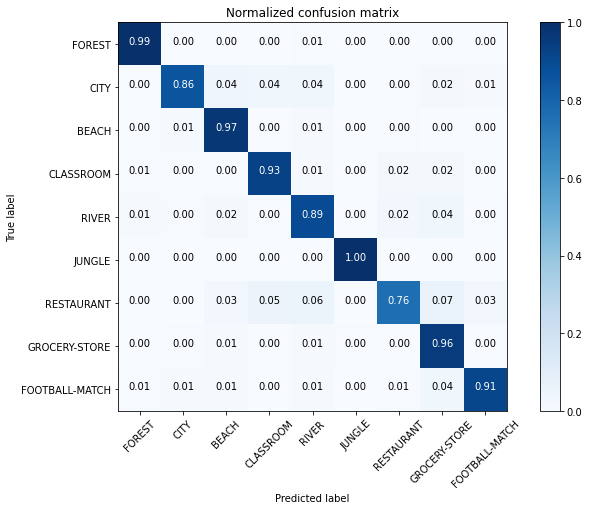}  
	\end{minipage}
	\caption{Confusion matrices for \textbf{audio modality}. The confusion matrix on testing dataset (left), and the normalized confusion matrix of testing dataset (right).}
	\label{fig:figure_6_19}
\end{figure}

Figures \ref{fig:figure_6_17}, \ref{fig:figure_6_18} and \ref{fig:figure_6_19} show the average performance of our framework three times. In this experiment, we still divide 90 participants into 3 groups, 30 for image modality as group 1, 30 for audio modality as group 2, and 30 for image-audio modality as group 3. We then assign an index to each participant from 1 to 30 in each group. For the participants hold the same index, we give the same number of samples, with the limit that the total number of samples owned by each group is always 13800. For example, after we initialize the framework, if a participant with index 1 in group 1 has 470 samples, then the participant with index 1 group 2 must have 470 samples, and participant with index 1 in group 3 must also have 470 samples. 
The data samples in all participants are randomly assigned. This assignment is an independent sampling without replacement. In total, we set 100 global epochs for global models.

The accuracies of our framework (Federated Transfer Learning) are higher than the implemented baseline models in Federated Learning.
More specifically, the accuracy of image modality is about 1\% higher, and that of audio modality is about 4\% improved (Table \ref{tab:table_6_4}).

Our framework's training and validation losses for two uni-modalities dramatically decrease in 10 global epochs, and they remain stable after 20 global epochs. 
The training losses of image and audio modalities remain unchanged with a jitter at 0.15 after 20 global epochs, while the validation of those remains stable with small jitters at 0.4.
For image-audio multi-modality (Contrastive Learning), training loss dramatically decreases in 40 global epochs. Moreover, it remains unchanged with small jitters after 60 global epochs, while the validation is smoother after 40 global epochs.
In addition, all training and validation accuracies increase dramatically in 10 global epochs and remain stable with slight jitter after 40 global epochs. From 20 to 100 global epochs, the validation accuracy of the image is always higher than the audio (Figure \ref{fig:figure_6_17}).

Most categories are correctly predicted, and true label ratios are higher than 90\% in the normalized confusion matrix for image modality, while the correctnesses of \textit{city}, \textit{river}, \textit{restaurant}, and \textit{football-match} are 89\%, 80\%, 81\% and 85\% in sequence. Compared with the image, the correctnesses of \textit{city}, \textit{classroom}, \textit{river}, and \textit{restaurant} are 82\%, 89\%, 87\% and 75\%, respectively, and the rest categories are higher than 90\% (Figures \ref{fig:figure_6_18} and \ref{fig:figure_6_19}).

\begin{table}[htbp]
	\centering
	\begin{tabular}{|l|l|l|}
		\hline 
%		\hline 
		\textbf{Modality} & \makecell{\textbf{Average Testing Accuracy}\\ Our Framework} & \makecell[l]{\textbf{Average Testing Accuracy}\\ (Implemented Baselines Models \\ \textbf{with} Federated Learning)} \\
%		\hline \label{key}
		\hline 
		\makecell[l]{\textbf{Image (Visual)}} & 94.42\% & 93.68\% \\
		\hline
		\makecell[l]{\textbf{Audio (Auditory)}}& 91.77\% & 88.16\%\\
		\hline
	\end{tabular}
	\caption{Average accuracy of our framework.}
	\label{tab:table_6_4}
\end{table}

% loss, multiloss, acc for self-supervision

% CM for self-supervision

%% file: SSL_diff_diff.tex
% different clients have different size of samples
\subsection{New Design with Unbalanced Distribution \rom{2}.}\label{subsubSSL_diff_diff}
\begin{figure}[!h]
	\centering
	\begin{minipage}[H]{0.49\textwidth}
		\includegraphics[width=1.0\textwidth]{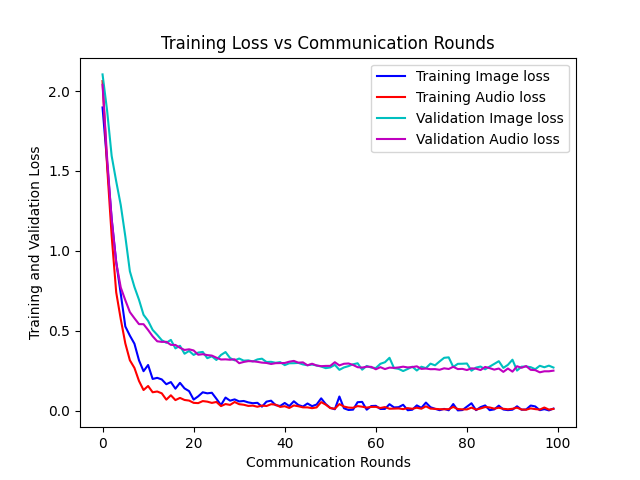}  	  	 
	\end{minipage}
	\begin{minipage}[H]{0.49\textwidth}
		\includegraphics[width=1.0\textwidth]{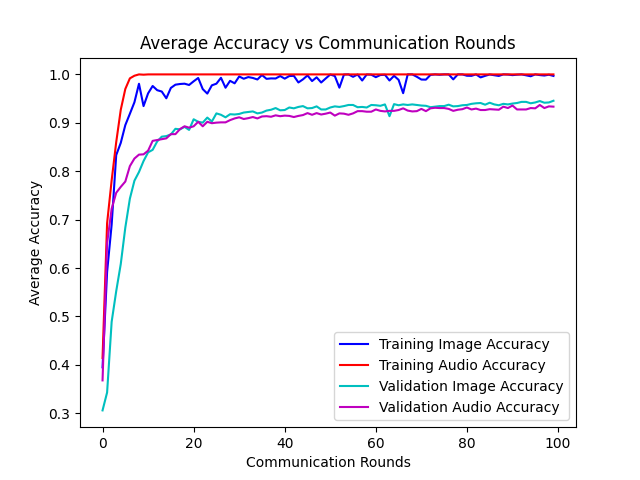}  
	\end{minipage}
	\begin{minipage}[H]{0.49\textwidth}
		\includegraphics[width=1.0\textwidth]{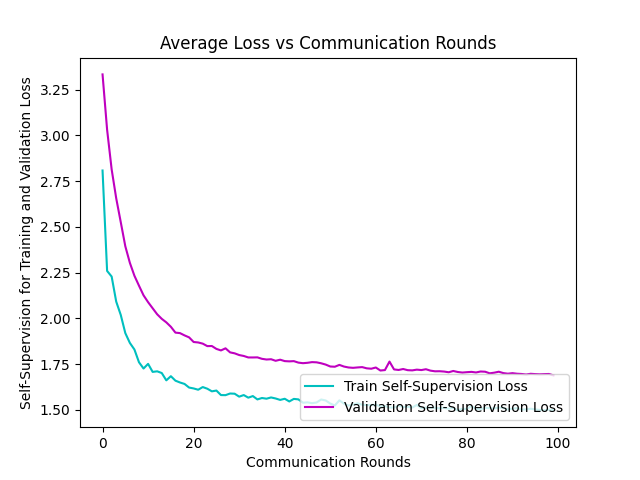}  
	\end{minipage}
	
	\caption{The loss of training and validation of unimodal supervision decreases when the number of epochs of communications increases (left). The accuracy of training and validation increases when the number of epochs of communications decreases (right). The loss of training and validation of multi-modal self-supervision decreases when the number of epochs of communications increases (bottom).}
	\label{fig:figure_6_20}
\end{figure}

\begin{figure}[!h]
	\centering
	\begin{minipage}[H]{0.49\textwidth}
		\includegraphics[width=1.0\textwidth]{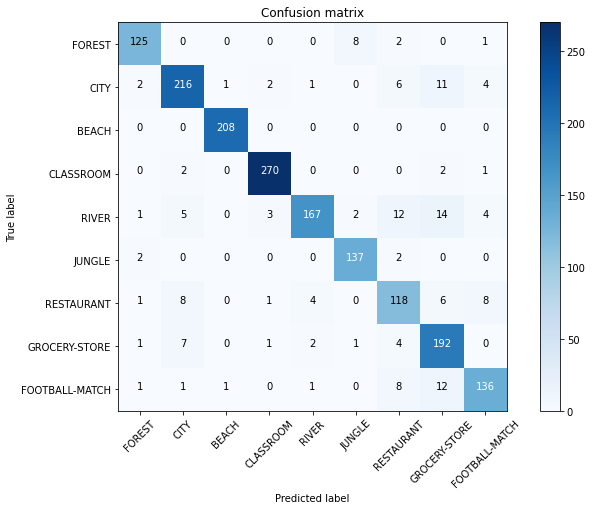}
	\end{minipage}
	\begin{minipage}[H]{0.49\textwidth}
		\includegraphics[width=1.0\textwidth]{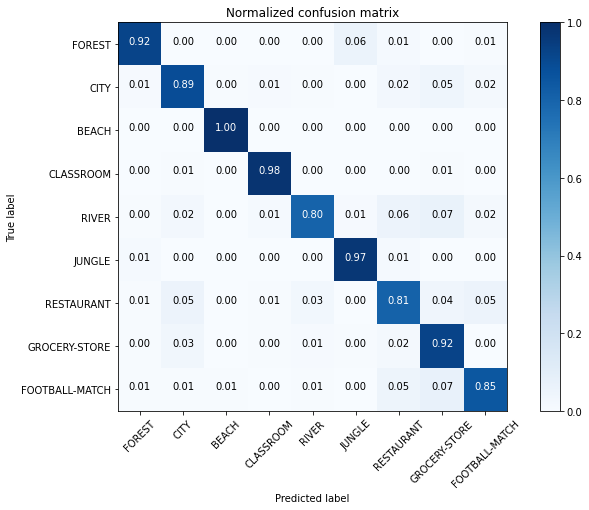}  
	\end{minipage}
	\caption{\textbf{Image modality}:The confusion matrix on testing dataset (left), and the normalized confusion matrix of testing dataset (right).}
	\label{fig:figure_6_21}
\end{figure}

\begin{figure}[!h]
	\centering
	\begin{minipage}[H]{0.49\textwidth}
		\includegraphics[width=1.0\textwidth]{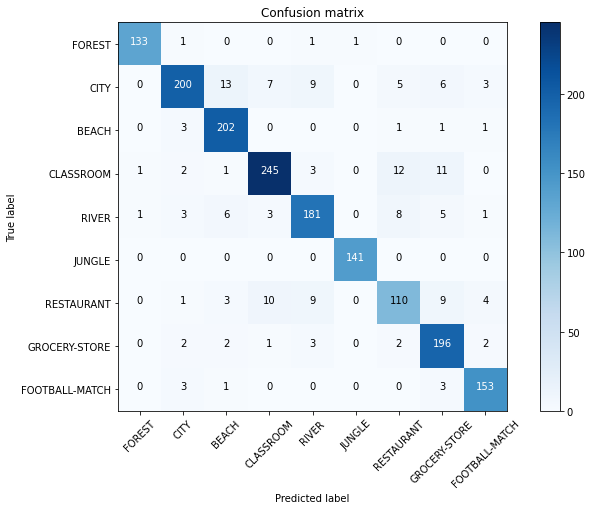}
	\end{minipage}
	\begin{minipage}[H]{0.49\textwidth}
		\includegraphics[width=1.0\textwidth]{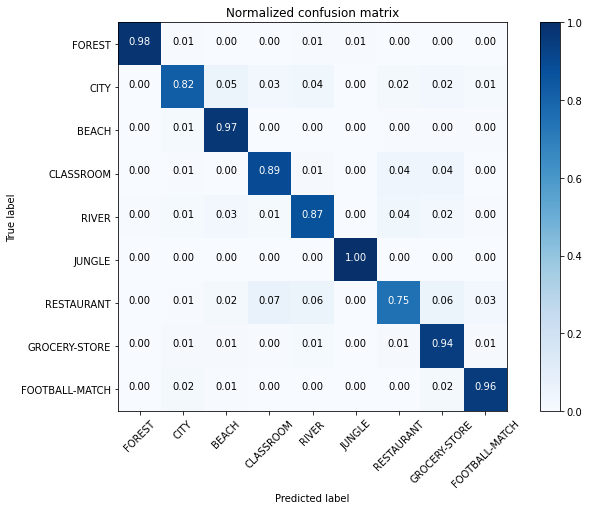}  
	\end{minipage}
	\caption{\textbf{Audio modality}:The confusion matrix on testing dataset (left), and the normalized confusion matrix of testing dataset (right).}
	\label{fig:figure_6_22}
\end{figure}

Figures \ref{fig:figure_6_20}, \ref{fig:figure_6_21} and \ref{fig:figure_6_22} show the average performance of our framework. In this experiment, we also divide 90 participants into three groups, 30 for image modality as group 1, 30 for audio modality as group 2, and 30 for image-audio modality as group 3. All the participant hold a random number of samples. However, the total number of sample owned by each group is still 13800.
%Each modality holds the whole dataset, and each modality divides into different sizes of samples in different ways, e.g. 460 for image, 510 for audio, and 470 for image-audio. All the participants have a similar distribution.
The data samples in all participants are randomly assigned.
This assignment is an independent sampling without replacement.
In total, we set 100 global epochs for global models.

The accuracies of our framework (Federated Transfer Learning) are higher than the implemented baseline models in Federated Learning.
More specifically, the accuracy of image modality is about 0.5\% higher, and that of audio modality is about 4.5\% improved (Table \ref{tab:table_6_5}).

Our framework's training and validation losses for two uni-modalities dramatically decrease in 10 global epochs, and they remain stable after 20 global epochs. 
The training losses of image and audio modalities remain unchanged with jitter at 0.1 after 20 global epochs, while the validation of those remains stable with jitter at 0.3. Besides, after 50 global epochs, the validation loss of image modality increases negligibly. Both uni-modalities of training loss overlap after 35 global epochs.
For image-audio multi-modality (self-supervision), its training and validation losses dramatically and smoothly decrease in 40 global epochs and remain unchanged after 60 global epochs.
In addition, all training and validation accuracies increase dramatically in 10 global epochs and remain stable with slight jitter after 40 global epochs. From 10 to 100 global epochs, the validation accuracy of the image is always higher than the audio (Figure \ref{fig:figure_6_20}).

Most categories are correctly predicted, and true label ratios are higher than 90\% in the normalized confusion matrix for image modality, while \textit{football-match} has only 89\%. Compared with the image, the correctness of \textit{restaurant} is only 76\%, and the rest categories are higher than 90\% (Figures \ref{fig:figure_6_21} and \ref{fig:figure_6_22}). 

\begin{table}[htbp]
	\centering
	\begin{tabular}{|l|l|l|}
		\hline 
		\textbf{Modality} & \makecell{\textbf{Average Testing Accuracy}\\ Our Framework} & \makecell[l]{\textbf{Average Testing Accuracy}\\ (Implemented Baselines Models \\ \textbf{with} Federated Learning).} \\
		\hline 
		\makecell[l]{\textbf{Image (Visual)}} & 93.91\% & 93.68\%\\
		\hline
		\makecell[l]{\textbf{Audio (Auditory)}} & 92.64\% & 88.16\%\\
		\hline 
	\end{tabular}
	\caption{Average accuracy of our framework.}
	\label{tab:table_6_5}
\end{table}
% CM for self-supervision when the pariticipants have different number 10v20; 10v15

% CM for self-supervision when the pariticipants have different number 10v10; 10v15
